%% file: main.tex
\documentclass[Afour,sageh,times]{sagej} 

\usepackage{moreverb,url}
\usepackage[colorlinks,bookmarksopen,bookmarksnumbered,citecolor=red,urlcolor=red]{hyperref}
\usepackage{styles/basic,styles/math,styles/tables,styles/algorithms}
\usepackage{graphicx}
\usepackage{subfigure}
\usepackage{hyperref}

\newcommand\BibTeX{{\rmfamily B\kern-.05em \textsc{i\kern-.025em b}\kern-.08em
T\kern-.1667em\lower.7ex\hbox{E}\kern-.125emX}}

\graphicspath{{figures/}}

\begin{document}

\runninghead{Hou et al.}

\title{Generation of Conservative Dynamical Systems Based on Stiffness Encoding}

\author{Tengyu Hou\affilnum{1}, Hanming Bai\affilnum{1}, Ye Ding\affilnum{1} and Han Ding\affilnum{1,2}}

\affiliation{\affilnum{1}State Key Laboratory of Mechanical System and Vibration, School of Mechanical Engineering, Shanghai Jiao Tong University, Shanghai 200240, China\\
\affilnum{2}State Key Laboratory of Intelligent Manufacturing Equipment and Technology, School of Mechanical Science and Engineering, Huazhong University of Science and Technology, Wuhan, Hubei 430074, China}


\corrauth{Ye Ding, State Key Laboratory of Mechanical System and Vibration, School of Mechanical Engineering, Shanghai Jiao Tong University, Shanghai 200240, China.}

\email{y.ding@sjtu.edu.cn}

\begin{abstract}
    Dynamical systems (DSs) provide a framework for high flexibility, robustness, and control reliability and are widely used in motion planning and physical human-robot interaction. The properties of the DS directly determine the robot's specific motion patterns and the performance of the closed-loop control system. In this paper, we establish a quantitative relationship between stiffness properties and DS. We propose a stiffness encoding framework to modulate DS properties by embedding specific stiffnesses. In particular, from the perspective of the closed-loop control system's passivity, a conservative DS is learned by encoding a conservative stiffness. The generated DS has a symmetric attraction behavior and a variable stiffness profile. The proposed method is applicable to demonstration trajectories belonging to different manifolds and types (e.g., closed and self-intersecting trajectories), and the closed-loop control system is always guaranteed to be passive in different cases. For controllers tracking the general DS, the passivity of the system needs to be guaranteed by the energy tank. We further propose a generic vector field decomposition strategy based on conservative stiffness, which effectively slows down the decay rate of energy in the energy tank and improves the stability margin of the control system. Finally, a series of simulations in various scenarios and experiments on planar and curved motion tasks demonstrate the validity of our theory and methodology.
\end{abstract}


\maketitle

\import{sections/}{1_introduction}
\import{sections/}{2_related}

\import{sections/}{3_background}
\import{sections/}{4_generation_conserv_ds}
\import{sections/}{5_general_method}
\import{sections/}{6_simu_exp}
\import{sections/}{7_conclusion}

\begin{acks}
    Tengyu Hou would like to thank Mr. Pingyun Nie, Mr. Sen Xu, Mr. Yuhang Chen and Mr. Jiexin Zhang for the valuable discussion.
\end{acks}
	
\begin{funding}
	The authors disclosed receipt of the following financial support for the research, authorship, and/or publication of this article:
	This work was partially supported by the National Natural Science Foundation of China (Grant Nos.51822506, 51935010).
\end{funding}

\bibliographystyle{bibliography/SageH}
\bibliography{bibliography/reference}

\appendix
\import{appendices/}{appendix_A}
\import{appendices/}{appendix_B}
\import{appendices/}{appendix_C}
\import{appendices/}{appendix_D}

\end{document}

%% file: sections/1_introduction.tex

\section{Introduction}
\label{sec:introduction}

Using dynamical systems (DSs) for robot motion planning and control problems has become popular in robotics because they can generate highly flexible and robust motion and control strategies. DS has been used in a wide variety of scenarios, ranging from performing point-to-point and periodic motions~\citep{yaoGuidVectorField2021}, such as pick-and-place and imitating specific motion patterns, to handling tasks in dynamic scenarios, such as real-time obstacle avoidance~\citep{koptevReactiveCollisionfreeMotion2024}, catching flying objects~\citep{kimCatchingObjectsFlight2014,salehianDynamicalSystemApproach2016}, and in-hand manipulation~\citep{khadivarAdaptiveFingersCoordination2023}. DS was successfully applied to contact tasks such as assembly and polishing by properly regulating the interaction forces between the robot and the environment~\citep{amanhoudDynamicalSystemApproach2019,amanhoudForceAdaptationContact2020}.

In DS-based motion planning, for a robotic system defined in a state space  ${\boldsymbol{\xi }} \in {\mathbb{R}^n}$, we specify the system's motion evolution given by a suitably designed ordinary differential equation ${\boldsymbol{\dot \xi }} = {\mathbf{f}}\left( {\boldsymbol{\xi }} \right)$, where ${\mathbf{f}}\left( {\boldsymbol{\xi }} \right) \in {\mathbb{R}^n}$ is a continuously differentiable vector field.

\subsection{Learning DS from demonstrations} \label{sec:dsgen}
Learning from Demonstration (LfD) is a powerful method for effectively imitating task patterns by observing demonstration tasks. Many learning methods have been developed to generate DSs with stability guarantees, ensuring stability at the kinematic level.

A natural idea is to obtain stability constraints from Lyapunov's second method. \cite{Khansari2011SEDS} seminally introduced the stable estimator of dynamical systems (SEDS). They introduced constraints in the Gaussian Mixture Regression (GMR) framework based on the quadratic Lyapunov function (QLF), thus ensuring that the generated motion is globally asymptotically stable (GAS). Although using QLF simplifies the parameter estimation process for DS, SEDS can only accurately inscribe motion trajectories with a monotonically decreasing $L_2$-norm distance from the equilibrium point. To overcome this problem, \cite{figueroa2018phyconsis} entirely use the properties of a polytopic linear parameter varying (LPV) system. They propose the DS learning method LPV-DS in combination with parametrized QLF (P-QLF), which exhibits excellent reproduction accuracy for general nonlinear motions. \cite{mohammadkhansari-zadehLearningControlLyapunov2014} further proposed the weighted sum of asymmetric quadratic functions (WSAQF), which achieves a wider range of fitting capabilities while satisfying global validity. Combined with WSAQF, the control corrections are obtained by solving the convex optimization problem with constraints in real time, thus ensuring that the system is GAS. \cite{kronanderIncrementalMotionLearning2015} proposed locally modulated DS (LMDS) based on Lyapunov theory. This method locally shapes the existing DS through a modulation matrix while ensuring that the modulated DS is locally asymptotically stable (LAS). To address the problem of robots failing to track a specific reference trajectory when perturbed, \cite{figueroaLocallyActiveGlobally2022} novelly proposed locally active globally stable DS (LAGS-DS), which causes the DS to exhibit stiffness-like symmetric attraction behaviors.

To ensure stability at the kinematic level, contraction analysis and partial contraction analysis are also effective tools to provide stability constraints~\citep{lohmillerContractionAnalysisNonlinear1998,wangPartialContractionAnalysis2005}. The contraction theory ensures that all DS trajectories converge exponentially to each other and that the trajectories globally exponentially converge towards the equilibrium point. Similar to the real-time correction strategy in~\cite{mohammadkhansari-zadehLearningControlLyapunov2014}, \cite{blocherLearningStableDynamical2017} computed the control inputs based on the contraction theory to online stabilize the original DS with the spurious attractor. \cite{ravichandar2017learning} generated nonlinear trajectories with guaranteed exponential convergence conditions by providing stability constraints through partial contraction analysis. Further, \cite{ravichandarLearningPositionOrientation2019} learned position and orientation dynamics based on contraction theory.

A framework for learning diffeomorphism is proposed to generate more complex trajectories while ensuring stability. Diffeomorphism establishes a one-to-one correspondence between the original space and the transformed space, thus transforming the highly nonlinear and complex learning tasks in the original space into easy-to-learn tasks in the transformed space. Since the diffeomorphism maintains the consistency of DS stability in the original and transformed spaces, the stable and highly nonlinear DS in the original space can be obtained by learning the stable DS in the transformed space. \cite{neumannLearningRobotMotions2015} gave an analytic expression for diffeomorphism, which corresponds the complex Lyapunov function in the original space to the QLF in the transformed space, thus combining SEDS to generate a stable DS in the transformed space. \cite{perrinFastDiffeomorphicMatching2016} use diffeomorphism to correspond complex demonstration trajectories in the original space to simple linear trajectories in the transformed space, where the diffeomorphism is the composition of the locally weighted translations.

Scholars have proposed a series of novel works using neural networks, a powerful framework, in combination with the aforementioned Lyapunov theory, contraction theory, and diffeomorphic transformation approach \citep{dawsonSafeControlLearned2023}. \cite{Lemme2014neuralds} learn DS vector fields based on an extreme learning machine, and Lyapunov theory guarantees DS stability.
\cite{rana2020euclideanizing} construct diffeomorphism using the non-volume preserving (NVP) network architecture, which is a general class of mappings, and its parameter optimization is achieved through back-propagation. In the deep neural network framework, \cite{perez-dattariStableMotionPrimitives2023} used imitation and contrastive learning to give a learning loss that ensures stability, generating first-order and second-order stable DSs. Combining with the contraction theory, \cite{mohammadineural} proposed a neural network-based method to generate the contractive DS. The DS with contractive property is obtained by integrating the negative definite Jacobi matrix generated by a neural network. They further used the variational autoencoder to generate the contractive DS in high dimensional space. In addition, many scholars have used differential geometry, a powerful geometric tool, to generate DSs on manifolds~\citep{duanStructuredPredictionApproach2024,ficheraHybridQuadraticProgramming2023}.

\subsection{Research on impedance control} \label{sec:reimpedancon}
The aforementioned DS theory provides a powerful framework for robot trajectory planning. In realistic scenarios, considering the interaction between the robot and the environment in order for the robot to perform its tasks better, in addition to planning the robot's position and velocity, it is also necessary to compliantly regulate the interaction forces between the robot and the environment. Impedance control provides a solution to coordinate the dynamic relationship between position, velocity, and interaction forces~\citep{HoganImpedanceControl1985}.

When the robot interacts with the environment, an important property of the controller is passivity, which brings along advantageous robustness properties~\citep{ott2008cartesian}. The controller's passivity allows the robot to produce stable behavior when in contact with a passive environment. In this perspective, the classical impedance controller in the regulation case is passive. When considering more complex tasks and environments, researchers often require robots to track continuous trajectories or interact with the environment exhibiting variable impedance characteristics, but these dynamic properties are achieved at the expense of impedance controller passivity. To keep the robot passive while performing tracking tasks, many scholars have advocated using a time-independent velocity field to guide the robot's motion. \cite{liPassiveVelocityField1999} introduced the augmented coordinates to achieve the system's passivity in the framework of velocity field control (VFC) by rationally distributing the kinetic energy of the original system and the corresponding kinetic energy of the augmented coordinates. \cite{duindamPassiveCompensationNonlinear2004} devised a passive control rate to enable the robot to track the desired motion in an artificial potential field. Such methods are complex, and selecting the appropriate controller parameters is difficult. \cite{kishiPassiveImpedanceControl2003} ensured the system's passivity by introducing an energy tank structure to keep the direction of motion constant during trajectory tracking and regulating the velocity magnitude in real time, which degraded the tracking performance at the velocity level. Another area of research focuses on maintaining system passivity while the robot executes variable impedance control. \cite{ferragutiTankbasedApproachImpedance2013} guarantee the passivity of the variable stiffness control process by incorporating an energy tank, but the discontinuous control law degrades the system's dynamic performance when the energy tank is depleted. \cite{kronanderStabilityConsiderationsVariable2016} ensure that the impedance control is passive by imposing constraints on the impedance damping and stiffness matrices, but the method relies on external force measurements with an accurate dynamics model. In addition, many passivity-based control methods in the impedance control framework have been successfully applied in flexible joint robots~\citep{albu-schafferUnifiedPassivitybasedControl2007,ottPassivityBasedImpedanceControl2008,kepplerElasticStructurePreserving2018,spyrakos-papastavridisPassivityPreservationVariable2020}. 

Combining the idea of impedance control in the DS framework, \cite{Kronan2016passinter} give the widely used passive-DS controller based on the damping control law. By rational design, the impedance damping matrix can selectively dissipate energy in task-irrelevant directions, while the stiffness characteristics of the closed-loop control system are encoded in the DS ${\mathbf{f}}\left( {\boldsymbol{\xi }} \right)$. It is worth noting that when the vector field ${\mathbf{f}}\left( {\boldsymbol{\xi }} \right)$ is conservative, the robotic system is passive, thus ensuring stability when interacting with passive environments.
The passive-DS controller ensures the system's passivity when the robot tracks a continuous trajectory and performs variable impedance control by uniformly encoding the trajectory tracking information and the variable stiffness characteristics in the DS  ${\mathbf{f}}\left( {\boldsymbol{\xi }} \right)$. For a general non-conservative DS, \cite{Kronan2016passinter} introduced an energy tank to offset the loss of passivity due to the non-conservative component ${{\mathbf{f}}_{nc}}\left( {\boldsymbol{\xi }} \right)$. However, when the energy in the tank is depleted, the closed-loop system's dynamic performance and trajectory tracking performance are greatly reduced due to the switching control law introduced by the energy tank structure. Therefore, solving the problem of rationally decomposing the general DS into conservative and non-conservative components to slow down the energy tank's consumption rate remains a challenge. Under the framework of the passive-DS controller, \cite{huberPassiveObstacleAwareControl2024} further design the impedance damping matrix reasonably for obstacle avoidance scenarios, effectively reducing the probability of collision. Using the optimal control technique, \cite{ficheraHybridQuadraticProgramming2023} give a torque-based control framework for the second-order DS.

\subsection{Contribution} \label{sec:contribution}
From the above literature review, it is clear that most of the existing studies related to learning DS only consider stability at the kinematic level and do not consider the passivity of the control system (which corresponds to the conservativeness of the DS). A few studies, although learning conservative DS, are limited to the simple case of a 2D linear space and lack a systematic discussion for the generation of conservative DS on more complex trajectories or general manifolds. DS decomposition plays an important role in the performance of energy tank-based controllers. However, there is yet to be a method to propose a generic decomposition strategy. In addition, the current study only discusses the stiffness properties of DS at a qualitative level, and a systematic discussion of the stiffness properties for DS modulation is still needed.

This paper proposes a DS generation framework based on stiffness encoding, as shown in Figure~\ref{fig:StiffEncode}. We establish and summarise the connection between the encoded stiffness properties and the generated DS, and generate DS to meet the requirements of specific scenarios by designing the stiffness properties. We utilize the conservative stiffness matrix to obtain conservative DS, which guarantees the passivity of the controller. We further extend our approach to generate conservative DS on different manifolds and for different trajectory types. Further, we also propose a generic decomposition algorithm for non-conservative DS based on the conservative stiffness matrix, which significantly slows down the depletion rate of the energy tank. As shown above, we provide a systematic analysis and discussion for the conservativeness of DS.

The contribution of this paper can be summarised as follows.
\begin{enumerate}
\item 
We develop a framework for DS generation based on stiffness encoding (see Figure~\ref{fig:StiffEncode}), giving the connection between stiffness properties and their corresponding DS characteristics.
\item 
Based on Gaussian processes, the conservative DS with a symmetric attraction behavior and a variable stiffness profile is learned in linear space. The method is further generalized to SE(3) as well as to closed-loop and self-intersecting trajectories. In all scenarios, we prove the controller's passivity.
\item	
Combined with the controller's energy tank structure, we propose a generic decomposition strategy for non-conservative DS based on the conservative stiffness matrix.	
\item	
A series of simulations and experiments are conducted to validate the proposed theory and methodology.
\end{enumerate}

The rest of the paper is organized as follows. Section~\ref{sec:related} illustrates the contents of this paper in comparison with related work. Preliminaries are given in Section~\ref{sec:background}, where we introduce the framework of stiffness encoding and the necessary basics. In Section~\ref{sec:Conservative DS}, we generate conservative DS with symmetric attraction properties based on Gaussian processes in a 2D linear space. A generalization of the conservative DS generation method is given in Section~\ref{sec:generalization}: we discuss the case on SE(3) as well as closed-loop and self-intersecting trajectories. At the same time, we propose effective DS decomposition strategies. Section~\ref{sec:Simu_Exp} verifies the validity of the proposed theory and method through a series of simulations and experiments. Finally, Section~\ref{sec:Conclusion} concludes the whole paper.

%% file: sections/2_related.tex

\begin{figure*}[t]
	\centering
	\includegraphics[width=1.9\columnwidth]{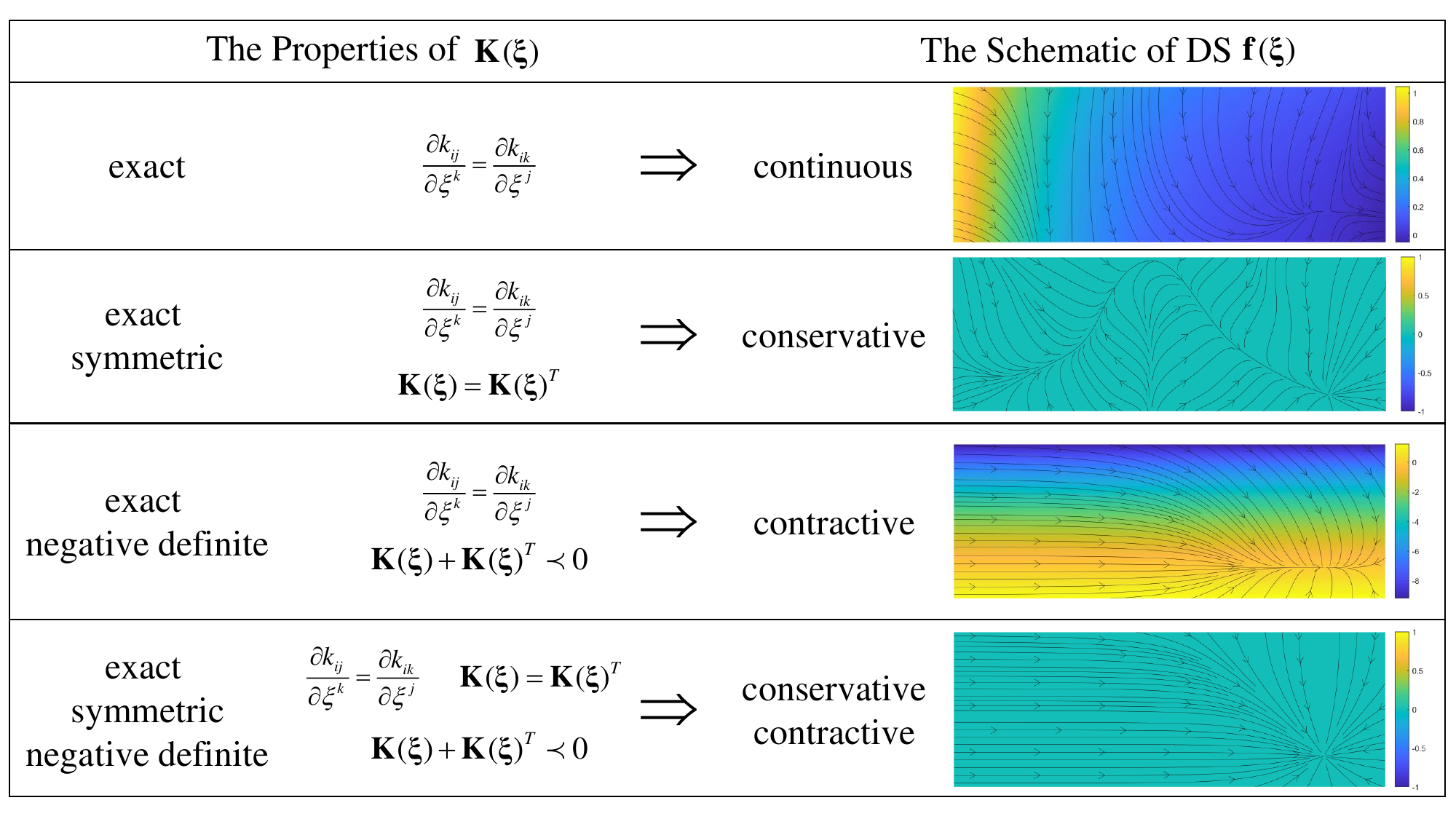}
	\caption{Schematic diagram of the stiffness encoding. The color of each point in the figure indicates the value of the angular velocity of the vector field at the corresponding point. When encoding a specific stiffness, the corresponding DS exhibits particular properties. (1) When the stiffness matrix satisfies the exactness, a well-defined DS is guaranteed, i.e., a continuous vector field in space. (2) The DS is conservative when the stiffness is exact and symmetric. Conservativeness is shown in the graph as a constant angular velocity of zero. In the figure, we show a conservative DS with symmetric attraction behavior, and it is not contractive. (3) For the DS to be contractive, the encoded stiffness must satisfy both exactness and negative definiteness. It can be seen that all trajectories of the DS converge to each other. (4) The DS is conservative and contractive when the stiffness matrix is exact, symmetric, and negative definite.}
	\label{fig:StiffEncode}
\end{figure*}

\section{Related work} \label{sec:related}
In our previous work, \cite{hou2024conserstiff} systematically discussed the connection between the exactness and symmetry of the stiffness matrix on the elastic force vector field it generates. The vector field is continuously distributed in space when the stiffness matrix is exact. The vector field is conservative when the stiffness matrix is exact and symmetric. In this paper, we use these quantitative relations to guide the generation of DS. In particular, on SE(3), we use the stiffness matrix's properties to generate a conservative DS (see Lemma~\ref{lem:consisconserv}). In addition to the symmetry and exactness mentioned above as properties affecting DS, negative definiteness is also important. As mentioned in Section~\ref{sec:dsgen}, many works learned DS based on the contraction theory, which ensures its asymptotic stability. In this paper, we clarify that these works essentially modulate the DS at the stiffness level. The simplest case of the contraction theory is equivalent to constraining the stiffness matrix to be constantly negative definite. We first sort out and summarise the mechanisms by which exactness, symmetry, and negative definiteness affect the DS, giving a framework for modulating the DS from the point of view of stiffness encoding, as shown in Figure~\ref{fig:StiffEncode}. In this framework, contraction theory is naturally included, and by combining different stiffness properties, we can obtain DS of various properties.

\cite{figueroaLocallyActiveGlobally2022} first clarified that stiffness-like symmetric attraction behaviors can enhance the perturbation resistance of a DS, and the robot can be quickly attracted to the demonstration trajectory after a perturbation. By designing DS with variable stiffness properties, \cite{michelPassivityBasedApproachVariable2024} enabled the robot to exhibit different impedance stiffnesses at different stages of motion, and the system was thus able to interact well with environments with different physical properties. Although the above DSs exhibit interesting stiffness properties, they are all non-conservative, and all of them require the introduction of an energy tank to ensure the passivity of the system when control is performed. \cite{khansari-zadehLearningPotentialFunctions2017} imitate the demonstration behavior by generating a conservative DS by learning potential functions. However, to improve the learning accuracy, they need to introduce a dissipative field, which results in the total DS no longer being conservative. In this paper, we consider the above important properties and present the generation of conservative DS with stiffness-like symmetric attraction behaviors and variable stiffness properties. We also present the generation strategy of conservative DS for more complex trajectories and spaces.

\cite{Kronan2016passinter} propose a practical and effective passive-DS controller, and they further point out that non-conservativeness leads to loss of passivity, but the decomposition of DS into conservative and non-conservative parts can alleviate this phenomenon. To address this issue, they propose vector field decomposition methods for DSs generated by specific methods (e.g., SEDS and LMDS). They use the Lyapunov function corresponding to the learning process of the DS to construct the conservative part, but this is not general, and other ways exist to improve the decomposition to a large extent. In this paper, we propose a decomposition strategy for general DS. The decomposition of DS is achieved by optimizing the decomposition index based on the energy tank structure. Simulations and experiments show that the proposed decomposition method can effectively slow down the energy decay of the energy tank.

%% file: sections/3_background.tex

\section{Preliminaries}\label{sec:background}
In this section, we give preliminaries. 
We begin by introducing impedance control in the DS framework, then present the idea of stiffness encoding in this paper, and finally introduce passivity analysis and Gaussian processes (GP).

\subsection{Impedance Control}\label{subsubsec:ImpCon}
The rigid-body dynamics of a robotic system are described by equations based on generalized coordinates $\boldsymbol{\xi} \in \mathbb{R}^n$:
\begin{equation}
	\label{eq:RigidDyn}
	{\mathbf{M}}({\boldsymbol{\xi }}){\boldsymbol{\ddot \xi }} + {\mathbf{C}}({\boldsymbol{\xi }},{\boldsymbol{\dot \xi }}){\boldsymbol{\dot \xi }} + {\mathbf{G}}({\boldsymbol{\xi }}) = {{\boldsymbol{\tau }}_c} + {{\boldsymbol{\tau }}_e},
\end{equation}
where ${\mathbf{M}}({\boldsymbol{\xi }}) \in \mathbb{R}^{n \times n}$, $ {\mathbf{C}}({\boldsymbol{\xi }},{\boldsymbol{\dot \xi }}) \in \mathbb{R}^{n \times n}$, and ${\mathbf{G}}({\boldsymbol{\xi }}) \in \mathbb{R}^n$ denote the mass matrix, the Coriolis matrix, and the gravity vector, respectively.

Based on the DS framework, \cite{Kronan2016passinter} proposed the passive-DS controller:
\begin{equation}
	\label{eq:PassiveDSContr}
	{{\boldsymbol{\tau }}_c} = {\mathbf{G}}({\boldsymbol{\xi }}) - {\mathbf{D}}({\boldsymbol{\xi }})({\boldsymbol{\dot \xi }} - {\mathbf{f}}({\boldsymbol{\xi }})) = {\mathbf{G}}({\boldsymbol{\xi }}) - {\mathbf{D}}({\boldsymbol{\xi }}){\boldsymbol{\dot \xi }} + {\lambda _1}{\mathbf{f}}({\boldsymbol{\xi }}).
\end{equation}
In (\ref{eq:PassiveDSContr}), the control damping matrix ${\mathbf{D}}({\boldsymbol{\xi }})$ is set in the following form:
\begin{equation}
	{\mathbf{D}}({\boldsymbol{\xi }}) = {\mathbf{Q}}({\boldsymbol{\xi }}){\boldsymbol{\Lambda Q}}{({\boldsymbol{\xi }})^T},
\end{equation}
where ${\boldsymbol{\Lambda }} = \text{diag}\left[ {{\lambda _1}, \cdots ,{\lambda _n}} \right]$ $({\lambda _1}, \cdots ,{\lambda _n}>0)$, and ${\mathbf{Q}}\left( {\boldsymbol{\xi }} \right) = \left[ {{{\mathbf{e}}_1}, \ldots ,{{\mathbf{e}}_n}} \right]$ is the orthogonal matrix where ${{\mathbf{e}}_1} = {\mathbf{f}}({\boldsymbol{\xi }})/\left\| {{\mathbf{f}}({\boldsymbol{\xi }})} \right\|$.	

The closed loop dynamics of the robotic system (\ref{eq:RigidDyn}) under controller (\ref{eq:PassiveDSContr}) can be expressed as
\begin{equation}
	\label{eq:ClosedRigidDyn}
	{\mathbf{M}}({\boldsymbol{\xi }}){\boldsymbol{\ddot \xi }} + \left( {{\mathbf{C}}({\boldsymbol{\xi }},{\boldsymbol{\dot \xi }}) + {\mathbf{D}}({\boldsymbol{\xi }})} \right){\boldsymbol{\dot \xi }} - {\lambda _1}{\mathbf{f}}({\boldsymbol{\xi }}) = {{\boldsymbol{\tau }}_e}.
\end{equation}
At this point, $- {\lambda _1}{\mathbf{f}}({\boldsymbol{\xi }})$ can be viewed as a nonlinear elastic force driving the robotic system. In comparison with classical impedance control, it is clear that the apparent stiffness ${{\mathbf{K}}_a}({\boldsymbol{\xi }})$ of the closed-loop system (\ref{eq:ClosedRigidDyn}) can be expressed as~\citep{chen2000conservative,figueroaLocallyActiveGlobally2022}
\begin{equation}
	\label{eq:ApparentStiffness}
	{{\mathbf{K}}_a}({\boldsymbol{\xi }}) = \frac{{\partial \left( { - {\lambda _1}{\mathbf{f}}({\boldsymbol{\xi }})} \right)}}{{\partial {\boldsymbol{\xi }}}} =  - {\lambda _1}\frac{{\partial {\mathbf{f}}({\boldsymbol{\xi }})}}{{\partial {\boldsymbol{\xi }}}}.
\end{equation}
Note that ${\lambda _1}$ is a constant scalar in practical control, and it is clear that the first-order derivative of DS ${\mathbf{f}}({\boldsymbol{\xi }})$ fully inscribes the stiffness properties of the closed-loop control system.

\subsection{Stiffness Encoding}
Section~\ref{subsubsec:ImpCon} shows that the stiffness information ${\mathbf{K}}({\boldsymbol{\xi }}) = \frac{{\partial {\mathbf{f}}({\boldsymbol{\xi }})}}{{\partial {\boldsymbol{\xi }}}}$ portrays important properties of closed-loop control systems and DS. Therefore, it is natural to think that we can obtain structure-specific DS and satisfactory system performance by carefully designing the stiffness ${\mathbf{K}}({\boldsymbol{\xi }})$.

To design the stiffness it is first necessary to study the critical properties of the stiffness matrix. One property worth noting is the conservativeness of the stiffness matrix:

\begin{lemma}
	[Conservative stiffness matrix~\citep{hou2024conserstiff}]
	\label{lem:constiffness}
	The conservativeness of the stiffness matrix ${\mathbf{K}}({\boldsymbol{\xi }})$ in a linear space is equivalent to the simultaneous satisfaction of symmetry and exactness:
	\begin{equation}
		\left\{ {\begin{array}{*{20}{c}}
				{k_{ij}} = {k_{ji}}, \\ 
				\frac{{\partial {k_{ij}}}}{{\partial {\xi^k}}} = \frac{{\partial {k_{ik}}}}{{\partial {\xi^j}}}.
		\end{array}} \right. \Rightarrow \left\{ {\begin{array}{*{20}{c}}
				{\mathbf{f}} = \oint_{\partial \Sigma } {\mathbf{K}}({\boldsymbol{\xi }})  \cdot d{\boldsymbol{\xi}} = \mathbf{0}, \\ 
				{W = \oint_{\partial \Sigma } {\mathbf{f}}  \cdot d{\boldsymbol{\xi}} = 0,} 
		\end{array}} \right.
	\end{equation}
	where ${k_{ij}}$ means the $i$th row and $j$th column of ${\mathbf{K}}$, and ${\partial \Sigma }$ is an arbitrary closed path.
\end{lemma}
\cite{hou2024conserstiff} showed that when the stiffness matrix ${\mathbf{K}}({\boldsymbol{\xi }})$ satisfies only exactness, the corresponding DS ${\mathbf{f}}({\boldsymbol{\xi }}) = \oint_{\partial \Sigma } {\mathbf{K}}({\boldsymbol{\xi }})  \cdot d{\boldsymbol{\xi}}$ is guaranteed to be a continuous vector field.

Contractive behavior is a property one usually expects DS to have, and the following lemma gives its definition:
\begin{lemma}
	[Contraction theory~\citep{lohmillerContractionAnalysisNonlinear1998}]
	\label{lem:ContracTheory}
	Given an autonomous DS ${\boldsymbol{\dot \xi }} = {\mathbf{f}}\left( {\boldsymbol{\xi }} \right)$ and a positive definite matrix ${\mathbf{U}}\left( {\boldsymbol{\xi }} \right)$, the system is said to be globally contracting with respect to ${\boldsymbol{\xi }}$ if it satisfies the following condition:
	\begin{equation}
		\forall {\boldsymbol{\xi }} \in {\mathbb{R}^n}{\text{, }}{\frac{{\partial {\mathbf{f}}({\boldsymbol{\xi }})}}{{\partial {\boldsymbol{\xi }}}}^T}{\mathbf{U}}({\boldsymbol{\xi }}) + {\mathbf{\dot U}}({\boldsymbol{\xi }}) + {\mathbf{U}}({\boldsymbol{\xi }})\frac{{\partial {\mathbf{f}}({\boldsymbol{\xi }})}}{{\partial {\boldsymbol{\xi }}}} \leqslant  - {\gamma _c}{\mathbf{U}}({\boldsymbol{\xi }}),
	\end{equation}
	where ${\gamma _c}$ is a positive constant. At this point, all trajectories of the above contractive system converge to each other exponentially.
\end{lemma}
In this paper, we make it clear that the  contraction theory essentially constrains the DS from the stiffness level:
\begin{equation}
	{{\mathbf{K}}({\boldsymbol{\xi }})}^T{\mathbf{U}}({\boldsymbol{\xi }}) + {\mathbf{\dot U}}({\boldsymbol{\xi }}) + {\mathbf{U}}({\boldsymbol{\xi }}){\mathbf{K}}({\boldsymbol{\xi }}) \leqslant  - {\gamma _c}{\mathbf{U}}({\boldsymbol{\xi }}).
\end{equation}
When we take ${\mathbf{U}}({\boldsymbol{\xi }}) \equiv {\mathbf{I}}$, contraction implies that the symmetric part of ${\mathbf{K}}({\boldsymbol{\xi }})$ is negative definite: ${\mathbf{K}}({\boldsymbol{\xi }}) + {\mathbf{K}}{({\boldsymbol{\xi }})^T} \prec 0$.

Interestingly, we can combine the above exactness, symmetry and negative definiteness on stiffness with each other, so that we can flexibly modulate the properties of DS. The framework of stiffness encoding is shown in Figure~\ref{fig:StiffEncode}.

\subsection{Passivity Analysis}
In scenarios where a robotic system interacts with its environment, passivity is a very important property. If the controller ensures a passive relationship between the external force and the robot velocity, then the robot will behave stably in contact with any passive environment~\citep{Hogan1988control}.
The passive-DS controller (\ref{eq:PassiveDSContr}) has the following property:

\begin{lemma}
	[\citep{Kronan2016passinter}]
	\label{lem:stiffpasscon}  
	When ${\mathbf{f}}({\boldsymbol{\xi }})$ is conservative with the corresponding potential function ${V_{\mathbf{f}}}({\boldsymbol{\xi }})$, the system (\ref{eq:RigidDyn}) controlled by (\ref{eq:PassiveDSContr}) is passive with regard to the input-output pair $( {{\boldsymbol{\dot \xi }},{{\boldsymbol{\tau }}_e}} )$ with the storage function $W({\boldsymbol{\xi }},{\boldsymbol{\dot \xi }}) = \frac{1}{2}{{\boldsymbol{\dot \xi }}^T}{\mathbf{M}}({\boldsymbol{\xi }}){\boldsymbol{\dot \xi }} + {\lambda _1}{V_{\mathbf{f}}}({\boldsymbol{\xi }})$.
\end{lemma}

From the stiffness point of view, combined with Lemma~\ref{lem:stiffpasscon}, we naturally obtain the following proposition.
\begin{proposition} 
	\label{pro:passivityControl}
	The system (\ref{eq:RigidDyn}) is passive under joint control law ${{\boldsymbol{\tau }}_c}({\boldsymbol{\xi }})$ given by
	\begin{equation}
		\label{eq:conserlaw}
		{{\boldsymbol{\tau }}_c}({\boldsymbol{\xi }}) = {\mathbf{G}}({\boldsymbol{\xi }}) - {\mathbf{D}}({\boldsymbol{\xi }}){\boldsymbol{\dot \xi }} + \int_{{{\boldsymbol{\xi }}_0}}^{\boldsymbol{\xi }} {{{\mathbf{K}}_{ctrl} }}({\boldsymbol{\xi }}) d{\boldsymbol{\xi }},
	\end{equation}	
	where ${{\mathbf{K}}_{ctrl} }$ is symmetric and exact, i.e.,
	\begin{equation}
		{\left[ {{{\mathbf{K}}_{ctrl} }} \right]_{ij}} = {\left[ {{{\mathbf{K}}_{ctrl} }} \right]_{ji}},\frac{{\partial {{\left[ {{{\mathbf{K}}_{ctrl} }} \right]}_{ki}}}}{{\partial {\xi ^j}}} = \frac{{\partial {{\left[ {{{\mathbf{K}}_{ctrl} }} \right]}_{kj}}}}{{\partial {\xi ^i}}}.
	\end{equation}
\end{proposition}

The passivity of the controllers~(\ref{eq:PassiveDSContr}) and~(\ref{eq:conserlaw}) imposes corresponding requirements on the property of ${\mathbf{f}}({\boldsymbol{\xi }})$ and ${\mathbf{K}}({\boldsymbol{\xi }})$. A conservative DS ${\mathbf{f}}({\boldsymbol{\xi }})$ would be a better choice which would improve the performance of the controller.
This paper focuses on guaranteeing the controller's passivity; hence, it concentrates on utilizing a conservative stiffness matrix to modulate the DS in the framework of stiffness encoding.

When ${\mathbf{f}}({\boldsymbol{\xi }})$ is a non-conservative DS, one often introduces an energy tank to correct the controller~\citep{Kronan2016passinter}:

\begin{subequations}
	\label{eq:EnergyTank}
	\begin{align}
		{{\boldsymbol{\tau }}_c} &= {\mathbf{G}}({\boldsymbol{\xi }}) - {\mathbf{D}}({\boldsymbol{\xi }}){\boldsymbol{\dot \xi }} + \beta({\boldsymbol{\xi }, s}) {\lambda _1}{\mathbf{f}}({\boldsymbol{\xi }}),\label{eq:EnergyTankA} \\
		\dot s &= \alpha(s) {{\boldsymbol{\dot \xi }}^T}{\mathbf{D}}({\boldsymbol{\xi }}){\boldsymbol{\dot \xi }} - \beta({\boldsymbol{\xi }, s}) {\lambda _1}{{\boldsymbol{\dot \xi }}^T}{\mathbf{f}}({\boldsymbol{\xi }}), \label{eq:EnergyTankB}
	\end{align}
\end{subequations}	
where $\alpha(s)$ and $\beta({\boldsymbol{\xi }, s}) $ are the energy tank switching functions, and $s \in \left[ {{s^l},{s^u}} \right]$ is the energy of the energy tank.

When the energy tank is depleted, the performance of the robotic system is affected (see Figure~\ref{fig:EnergyA}). To slow down the rate of energy depletion, the DS can be decomposed into the conservative part ${{\mathbf{f}}_c}({\boldsymbol{\xi }})$ and the non-conservative part ${{\mathbf{f}}_{nc}}({\boldsymbol{\xi }})$:
\begin{equation}
	{\mathbf{f}}({\boldsymbol{\xi }}) = {{\mathbf{f}}_c}({\boldsymbol{\xi }}) + {{\mathbf{f}}_{nc}}({\boldsymbol{\xi }}).
\end{equation}
At this point the controller is shown below:
\begin{subequations}
	\label{eq:EnergyTank1}
	\begin{align}
		{{\boldsymbol{\tau }}_c} &= {\mathbf{G}}({\boldsymbol{\xi }}) - {\mathbf{D}}({\boldsymbol{\xi }}){\boldsymbol{\dot \xi }} + {\lambda _1}{{\mathbf{f}}_c}({\boldsymbol{\xi }}) + \beta({\boldsymbol{\xi }, s}) {\lambda _1}{{\mathbf{f}}_{nc}}({\boldsymbol{\xi }}), \label{eq:EnergyTank1A} \\
		\dot s &= \alpha(s) {{\boldsymbol{\dot \xi }}^T}{\mathbf{D}}({\boldsymbol{\xi }}){\boldsymbol{\dot \xi }} - \beta({\boldsymbol{\xi }, s}) {\lambda _1}{{\boldsymbol{\dot \xi }}^T}{{\mathbf{f}}_{nc}}({\boldsymbol{\xi }}). \label{eq:EnergyTank1B}
	\end{align}
\end{subequations}
As can be seen from~(\ref{eq:EnergyTank1}), only the non-conservative part ${{\mathbf{f}}_{nc}}({\boldsymbol{\xi }})$ is consuming energy. Therefore, when DS is non-conservative, a reasonable DS decomposition plays an important role in the controller performance.

\subsection{Gaussian Processes}
The Gaussian process (GP) used to describe the distribution of the function is completely determined by its mean function and covariance function 
$k^{\boldsymbol{\theta }}({{\boldsymbol{\xi }},{{\boldsymbol{\xi }}^{'}}})$, where ${\boldsymbol{\theta }}$ is the hyperparameters of GP. In this paper, we take the mean function as zero and choose the commonly used squared exponential covariance function.

For the $N$ training points
$\boldsymbol{X = }[{{\boldsymbol{\xi }}_1}{\text{ }}{{\boldsymbol{\xi }}_2}{\text{ }} \cdots {\text{ }}{{\boldsymbol{\xi }}_N}]$ and the observed target values ${\mathbf{y}} = {[{y_{1{\text{ }}}}{y_2} \cdots {\text{ }}{y_N}]^T}$, the prediction output ${y_{pre}}$ corresponding to the test point ${\boldsymbol{\xi }}$ for GP regression can be described as

\begin{equation}
	{y_{pre}}\mid {\mathbf{X}},{\mathbf{y}},{\boldsymbol{\xi }}
	\sim \mathcal{N}\left( {\overline {{y_{pre}}} ,\operatorname{cov} \left( {{y_{pre}}} \right)} \right),
\end{equation}
where
\begin{equation}
	\begin{aligned}
		\overline {{y_{pre}}} 
		&= gp\left( {{\boldsymbol{\theta }},{\boldsymbol{X}},{\mathbf{y}},{\boldsymbol{\xi }}} \right) \\
		&= {K^{\boldsymbol{\theta }}}\left( {{\boldsymbol{\xi }},{\boldsymbol{X}}} \right){\left[ {{K^{\boldsymbol{\theta }}}({\boldsymbol{X}},{\boldsymbol{X}}) + \sigma _n^2{\mathbf{I}}} \right]^{ - 1}}{\mathbf{y}},
	\end{aligned} 
\end{equation}
\begin{equation}
	\begin{aligned}
	&\operatorname{cov} \left( {{y_{pre}}} \right)\\
	&= {K^{\boldsymbol{\theta }}}\left( {{\boldsymbol{\xi }},{\boldsymbol{\xi }}} \right) - {K^{\boldsymbol{\theta }}}\left( {{\boldsymbol{\xi }},{\boldsymbol{X}}} \right){\left[ {{K^{\boldsymbol{\theta }}}({\boldsymbol{X}},{\boldsymbol{X}}) + \sigma _n^2{\mathbf{I}}} \right]^{ - 1}}{K^{\boldsymbol{\theta }}}\left( {{\boldsymbol{X}},{\boldsymbol{\xi }}} \right).
	\end{aligned} 
\end{equation}
$\overline {{y_{pre}}}$ and $\operatorname{cov} \left( {{y_{pre}}} \right)$ are the posterior mean and covariance matrix of ${y_{pre}}$. $\mathbf{I}$ is the $N \times N$ identity matrix. ${K^{\boldsymbol{\theta }}}({\boldsymbol{X}},{\boldsymbol{X}})$ is the $N \times N$ covariance matrix which can be evaluated elementwise based on the training points with the covariance function $k^{\boldsymbol{\theta }}({{\boldsymbol{\xi }},{{\boldsymbol{\xi }}^{'}}})$,  and similarly for ${K^{\boldsymbol{\theta }}}\left( {{\boldsymbol{\xi }},{\boldsymbol{\xi }}} \right)$, ${K^{\boldsymbol{\theta }}}\left( {{\boldsymbol{\xi }},{\boldsymbol{X}}} \right)$.
${\sigma _n^2}$ is the variance of the noise, contained in ${\boldsymbol{\theta }}$.
The hyperparameters ${\boldsymbol{\theta }}$ can be optimized by minimizing the negative logarithmic marginal likelihood (NLML) $-{\text{log}}\text{ }p\left( {\mathbf{y}}{\left| {\boldsymbol{X},\boldsymbol{\theta }} \right.} \right)$ \citep{williams2006gaussian}.

%% file: sections/4_generation_conserv_ds.tex

\begin{figure*}[!ht]
	\begin{center}
		\subfigure[\label{fig:V0L}]
		{\includegraphics[width=0.5\columnwidth]{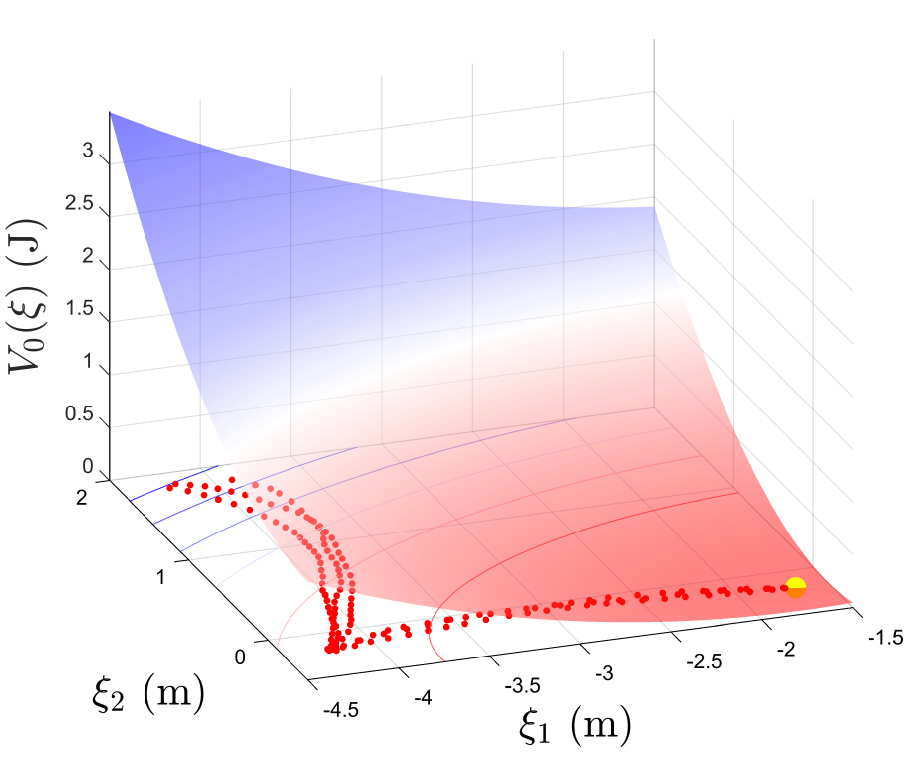}}%
		\hspace{0.00cm}
		\subfigure[\label{fig:VtankL}]
		{\includegraphics[width=0.5\columnwidth]{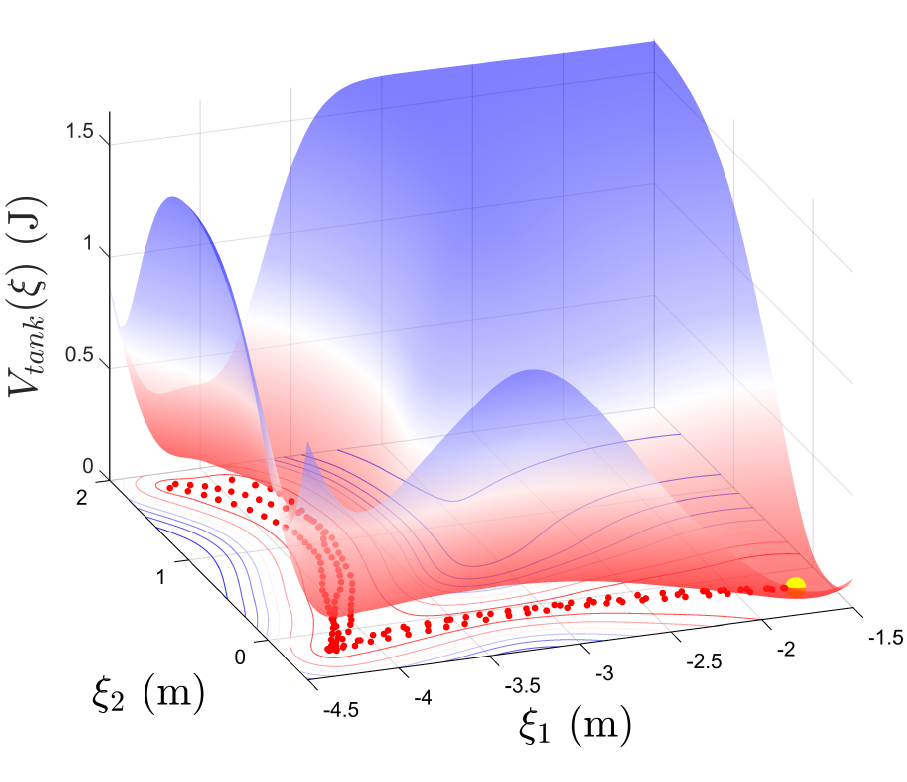}}
		\vspace{0.00cm}		
		\subfigure[\label{fig:V1L}]
		{\includegraphics[width=0.5\columnwidth]{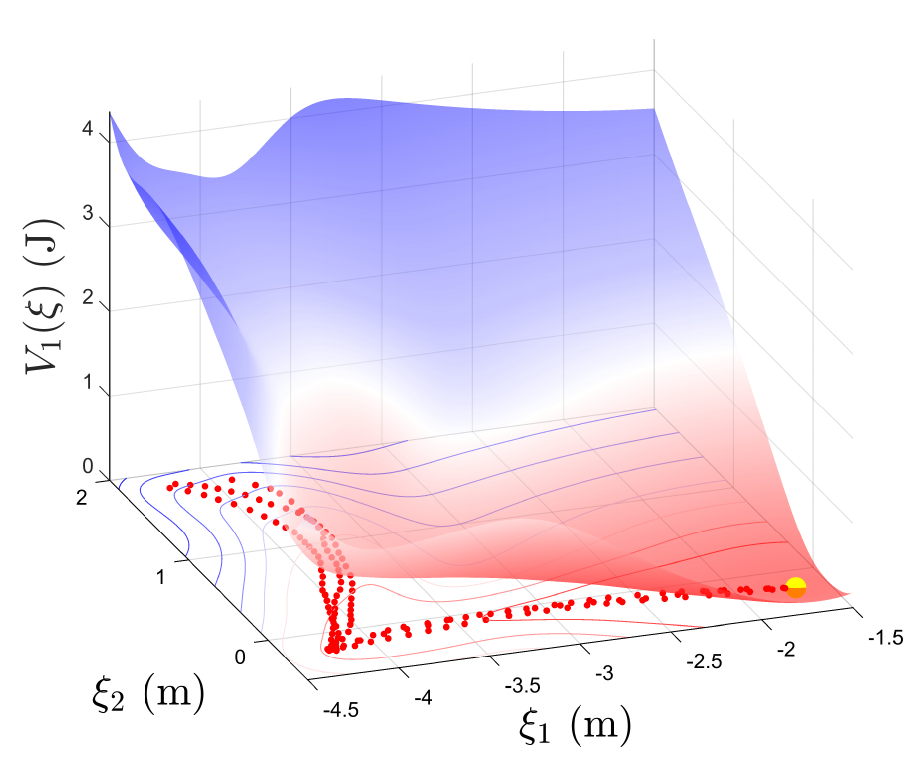}}%
		\hspace{0.00cm}
		\subfigure[\label{fig:IntPathL}]
		{\includegraphics[width=0.5\columnwidth]{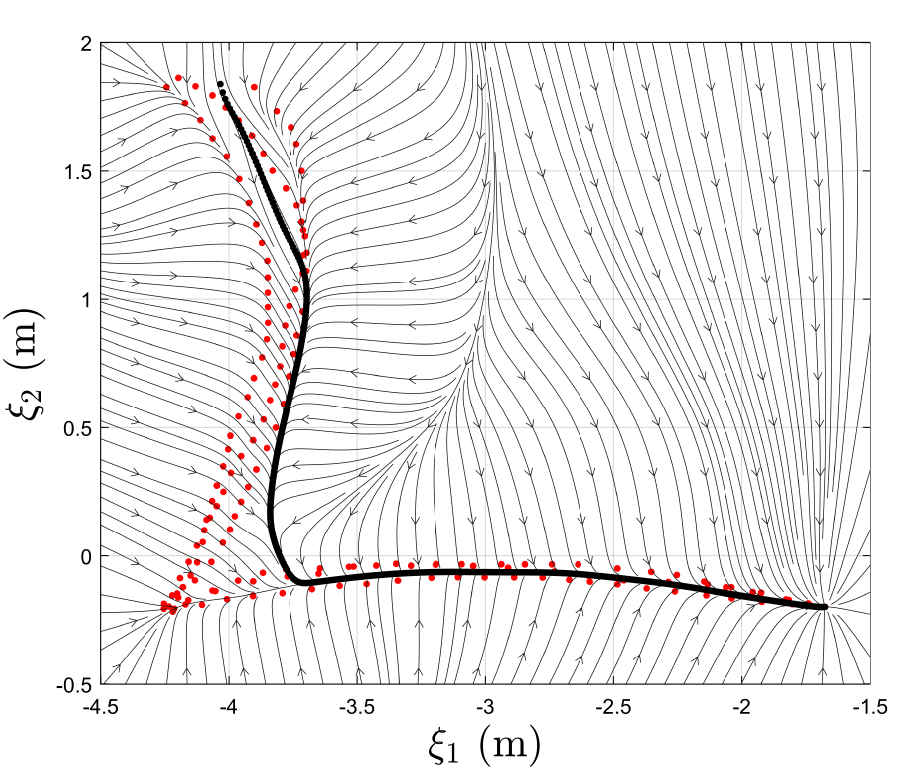}}
		\vspace{0.00cm}	
	\end{center}
	\caption{\label{fig:GenIntPathL} {Schematic of integral path generation in Section~\ref{subsubsec:Step1}. (a)-(c) show the construction process of ${V_1}\left( {\boldsymbol{\xi }} \right)$. The red dots represent demonstration trajectories. (d) shows ${{\mathbf{f}}_1}({\boldsymbol{\xi }}) 
	=  - \nabla {V_1}({\boldsymbol{\xi }})$, where the black curve is the integration path from the average start point of the demonstration trajectories.
	(a) The initial function ${V_0}\left( {\boldsymbol{\xi }} \right)$. (b) The notch function ${V_{\text{tank}}}\left( {\boldsymbol{\xi }} \right)$. (c) The composite function ${V_1}\left( {\boldsymbol{\xi }} \right)$. (d) The vector field ${{\mathbf{f}}_1}({\boldsymbol{\xi }})$ corresponding to ${V_1}\left( {\boldsymbol{\xi }} \right)$.}}
\end{figure*}

\section{Generation of Conservative DS}\label{sec:Conservative DS}
In this section, we present the conservative DS generation method with symmetric attraction property, while the stiffness is variably tunable.
As can be seen from Figure~\ref{fig:StiffEncode}, the DS's symmetric attraction behavior sometimes contradicts its contraction. Therefore, in this study, we focus on the symmetry and exactness of the stiffness matrix.
When the stiffness ${\mathbf{K}}$ is symmetric and exact, the elastic force ${\mathbf{f}}\left( {\boldsymbol{\xi }} \right)$ and the corresponding potential function $V\left( {{\boldsymbol{\xi }}} \right)$ can be expressed as (see Appendix~\ref{apd:EfPf} for details)
\begin{subequations}
	\label{eq:ConFV}
	\begin{align}
		{\mathbf{f}}\left( {\boldsymbol{\xi }} \right)&= -\nabla V\left( {\boldsymbol{\xi }} \right) =\nabla g\left( {{{\boldsymbol{\xi }}_0}} \right) - \nabla g\left( {\boldsymbol{\xi }} \right),\\
		V\left( {{\boldsymbol{\xi }}} \right) 
		&= g\left( {\boldsymbol{\xi }} \right) - g\left( {{{\boldsymbol{\xi }}_0}} \right) - \nabla g\left( {{{\boldsymbol{\xi }}_0}} \right) \cdot \left( {{\boldsymbol{\xi }} - {{\boldsymbol{\xi }}_0}} \right),
	\end{align}
\end{subequations}
where $g\left( {\boldsymbol{\xi }} \right)$ is a differentiable function that will be constructed by GP, and ${{\boldsymbol{\xi }}_0}$ is the equilibrium point of the DS.

From the previous description, ${\boldsymbol{\dot \xi }} = {\mathbf{f}}\left( {\boldsymbol{\xi }} \right)$ is considered as the designed DS and $V\left( {{\boldsymbol{\xi }}} \right)$ can be regarded as the Lyapunov function. According to the Lyapunov stability theory, the global asymptotic stability of the DS at point ${{{\boldsymbol{\xi }}_0}}$ imposes the following requirements on $V\left( {{\boldsymbol{\xi }}} \right)$:
\begin{equation}
	\label{eq:StabilityCons}
	\left\{ {\begin{array}{*{20}{c}}
			{{\text{(a)}}} \\ 
			{{\text{(b)}}} \\ 
			{{\text{(c)}}} 
		\end{array}\begin{array}{*{20}{c}}
			{V({\boldsymbol{\xi }}) > 0} \\ 
			{\dot V({\boldsymbol{\xi }}) < 0} \\ 
			{{\text{ }}V\left( {{{\boldsymbol{\xi }}_0}} \right) = 0,} 
	\end{array}} \right.{\text{ }}\begin{array}{*{20}{c}}
		{\forall {\boldsymbol{\xi }} \in {\mathbb{R}^n},{\text{ }}{\boldsymbol{\xi }} \ne {{\boldsymbol{\xi }}_0}} \\ 
		{\forall {\boldsymbol{\xi }} \in {\mathbb{R}^n},{\text{ }}{\boldsymbol{\xi }} \ne {{\boldsymbol{\xi }}_0}} \\ 
		{\dot V\left( {{{\boldsymbol{\xi }}_0}} \right) = 0.} 
	\end{array}
\end{equation}	
It is easy to verify that condition~(\ref{eq:StabilityCons}) (c) is satisfied due to the form~(\ref{eq:ConFV}). Notice that $\dot V({\boldsymbol{\xi }}) = \nabla V({\boldsymbol{\xi }}) \cdot {\boldsymbol{\dot \xi }} =  - {\left\| {\nabla V({\boldsymbol{\xi }})} \right\|^2}$, condition~(\ref{eq:StabilityCons}) can be expressed as follows:
\begin{subequations}
	\label{eq:stabilitycons}
	\begin{align}
		g\left( {\boldsymbol{\xi }} \right) > g\left( {{{\boldsymbol{\xi }}_0}} \right) + \nabla g\left( {{{\boldsymbol{\xi }}_0}} \right) \cdot \left( {{\boldsymbol{\xi }} - {{\boldsymbol{\xi }}_0}} \right),{\text{  }}{\boldsymbol{\xi }} \ne {{\boldsymbol{\xi }}_0}\\
		\left\| {\nabla g\left( {\boldsymbol{\xi }} \right) - \nabla g\left( {{{\boldsymbol{\xi }}_0}} \right)} \right\|^2 > 0,{\text{  }}{\boldsymbol{\xi }} \ne {{\boldsymbol{\xi }}_0}.
	\end{align}
\end{subequations}
Next we will describe how to construct a suitable function $g\left( {\boldsymbol{\xi }} \right)$ in a reasonable way, based on GP.

For the demonstration task, different demonstration trajectories differ, and our goal is for the robot to move along a trajectory that represents a common feature of the demonstration trajectories.	
In Section~\ref{subsubsec:Step1}, we generate the integral path that can represent the geometrical features of the demonstration trajectories.
Based on the generated integration path, the potential energy function ${V_p}$ is constructed in Section~\ref{subsubsec:Step2}. By suitable construction, the DS corresponding to ${V_p}$ is conservative and has symmetric attraction behavior.
However, many parameters of ${V_p}$ remain to be further defined and optimized.
In Section~\ref{subsubsec:Step3}, ${V_p}$ is optimized so that the actual motion velocity generated based on GP is consistent with the reference velocity $\left\{ {{\boldsymbol{\dot \xi }}_i^{{\text{ref}}}} \right\}_{i = 1}^N$.

\subsection{Generation of the Integral Path}\label{subsubsec:Step1}
In this section, we obtain the integral path based on the function ${V_1}$ constructed by GP.

\subsubsection{\texorpdfstring{Generation of the Initial Function ${V_0}$}{}}
$\\$
We choose a quadratic function as the initial function ${V_0}\left( {\boldsymbol{\xi }} \right)$ (see Figure~\ref{fig:V0L}), which and its corresponding vector field ${{\mathbf{f}}_0}\left( {\boldsymbol{\xi }} \right)$ are as follows:
\begin{subequations}
	\begin{align}
		{V_0}\left( {\boldsymbol{\xi }} \right) = {g_0}\left( {\boldsymbol{\xi }} \right) = {\left( {{\boldsymbol{\xi }} - {{\boldsymbol{\xi }}_0}} \right)^T}{\mathbf{P}}\left( {{\boldsymbol{\xi }} - {{\boldsymbol{\xi }}_0}} \right),\\
		{{\mathbf{f}}_0}\left( {\boldsymbol{\xi }} \right) =  - \nabla {V_0}\left( {\boldsymbol{\xi }} \right) =  - 2{\mathbf{P}}\left( {{\boldsymbol{\xi }} - {{\boldsymbol{\xi }}_0}} \right),
	\end{align}
\end{subequations}	
where ${\mathbf{P}}$ is a positive definite matrix to be optimized. It is optimized by solving the following optimization problem.

\begin{equation}
	\mathop {\min }\limits_{{\mathbf{P}} \succ 0} {\text{ }}\sum\limits_{i = 1}^N {{{\left\| { - 2{\mathbf{P}}\left( {{\boldsymbol{\xi }}_i^{{\text{ref}}} - {{\boldsymbol{\xi }}_0}} \right) - {\boldsymbol{\dot \xi }}_i^{{\text{ref}}}} \right\|}^2}}. 
\end{equation}
$\left\{ {{\boldsymbol{\xi }}_i^{{\text{ref}}},{\boldsymbol{\dot \xi }}_i^{{\text{ref}}}} \right\}_{i = 1}^N$ is the reference trajectories, as indicated by the red dots in Figure~\ref{fig:GenIntPathL}. Through optimization, the value of ${{\mathbf{f}}_0}\left( {\boldsymbol{\xi }} \right)$ on the reference path ${{\boldsymbol{\xi }}_i^{{\text{ref}}}}$ is as close as possible to the reference velocity ${{\boldsymbol{\dot \xi }}_i^{{\text{ref}}}}$.
The optimal ${\mathbf{P}}$ is denoted as ${{{\mathbf{P}}^*}}$, and the corresponding function is expreesed as ${V_0^*}\left( {\boldsymbol{\xi }} \right) = {\left( {{\boldsymbol{\xi }} - {{\boldsymbol{\xi }}_0}} \right)^T}{\mathbf{P}}^*\left( {{\boldsymbol{\xi }} - {{\boldsymbol{\xi }}_0}} \right)$.

\subsubsection{\texorpdfstring{Generation of the Composite Function ${V_1}$}{}}
$\\$
To make the vector field exhibit symmetric attraction with respect to the demonstration trajectory, we utilize the properties of GP to construct the notch function. 
We set all target values at the training points ${\boldsymbol{X}_{{\text{Ref}}}} = [{\boldsymbol{\xi }}_1^{{\text{ref}}}{\text{ }}{\boldsymbol{\xi }}_2^{{\text{ref}}}{\text{ }} \cdots {\text{ }}{\boldsymbol{\xi }}_N^{{\text{ref}}}]$ to 1 (i.e., ${\mathbf{y}} = {{\mathbf{I}}_{N \times 1}}$). By the property of the squared exponential covariance function $k^{\boldsymbol{\theta }}({{\boldsymbol{\xi }},{{\boldsymbol{\xi }}^{'}}})$, the value of $gp\left( {\boldsymbol{\theta},{\boldsymbol{X}_{{\text{Ref}}}},{{\mathbf{I}}_{N \times 1}},{\boldsymbol{\xi }}} \right)$ decays to zero as ${\boldsymbol{\xi }}$ moves away from the demonstration trajectories ${\boldsymbol{X}_{{\text{Ref}}}}$. 
Therefore, we construct the notch function ${V_{{\text{tank}}}}\left( {{\boldsymbol{\xi }},a,{\boldsymbol{\theta} _1}} \right)$ (see Figure~\ref{fig:VtankL}) based on the following structure:
\begin{equation}
	\label{eq:Vtank}
	{V_{{\text{tank}}}} = \left( {1 - {r_a}\left( {\boldsymbol{\xi }} \right)} \right)\left( {1 + \varepsilon  - gp\left( {{\boldsymbol{\theta} _1},{\boldsymbol{X}_{{\text{Ref}}}},{{\mathbf{I}}_{N \times 1}},{\boldsymbol{\xi }}} \right)} \right),
\end{equation}
where $\varepsilon$ is a small positive number such that the value of ${V_{{\text{tank}}}}$ is constantly non-negative. ${r_a}\left( {\boldsymbol{\xi }} \right) = \exp \left( { - a{{\left\| {{\boldsymbol{\xi }} - {{\boldsymbol{\xi }}_0}} \right\|}^2}} \right)$ is the radial basis function centered on ${{\boldsymbol{\xi }}_0}$. The coefficient $\left( {1 - {r_a}\left( {\boldsymbol{\xi }} \right)} \right)$ makes the values of the function ${V_{{\text{tank}}}}$ and its derivative at ${{\boldsymbol{\xi }}_0}$ are zero.

Based on the initial function ${V_{0}}$ and the notch function ${V_{{\text{tank}}}}$, we obtain the composite function ${V_1}$ (see Figure~\ref{fig:V1L}):
\begin{equation}
	\begin{aligned}
		{V_1}\left( {{\boldsymbol{\xi }},a,b,{\boldsymbol{\theta} _1}} \right) \hfill 
		&= {g_1}\left( {{\boldsymbol{\xi }},a,b,{\boldsymbol{\theta} _1}} \right) \hfill \\
		&= {V_0}^*\left( {\boldsymbol{\xi }} \right) + b{V_{{\text{tank}}}}\left( {{\boldsymbol{\xi }},a,{\boldsymbol{\theta} _1}} \right).
	\end{aligned}
\end{equation}

By solving the following optimization problem, we can then obtain the values of the coefficients to be determined:
\begin{equation}
	\begin{aligned}
		\mathop {\min }\limits_{\left\{ {a,b,{\boldsymbol{\theta} _1}} \right\}} {\text{ }}&\sum\limits_{i = 1}^N {{{\left\| { - \nabla {V_1}\left( {{\boldsymbol{\xi }}_i^{{\text{ref}}},a,b,{\boldsymbol{\theta} _1}} \right) - {\boldsymbol{\dot \xi }}_i^{{\text{ref}}}} \right\|}^2}}  \hfill \\
		\mathit{s.t.}~&a \in \left[ {{a^l},{a^u}} \right]{\text{,    }} \hfill \\
		&b \in \left[ {{b^l},{b^u}} \right]{\text{,   }} \hfill \\
		&{\boldsymbol{\theta} _1} \in \boldsymbol{\Theta} .
	\end{aligned} 
\end{equation}
$\nabla {V_1}$ can be obtained analytically for the solution of the optimization, and we will explain this in detail later. The optimal coefficients are denoted as ${\left\{ {a^*,b^*,{\boldsymbol{\theta} _1 ^*}} \right\}}$.

\subsubsection{Generation of the Integral Path}
$\\$
For ${\boldsymbol{\xi }} \in {\mathbb{R}^n}$, $\boldsymbol{X = }[{{\boldsymbol{\xi }}_1}{\text{ }}{{\boldsymbol{\xi }}_2}{\text{ }} \cdots {\text{ }}{{\boldsymbol{\xi }}_N}] \in {\mathbb{R}^{n \times N}}$ and ${\mathbf{y}} \in {\mathbb{R}^{N \times 1}}$, the posterior mean $gp\left( {\boldsymbol{\xi }} \right)$ and its gradient $\nabla gp\left( {\boldsymbol{\xi }} \right)$ can be expressed as
\begin{subequations}
	\begin{align}
		gp\left( {\boldsymbol{\xi }} \right) 
		&= {K^{\boldsymbol{\theta}} }{\left( {{\boldsymbol{\xi }},\boldsymbol{X}} \right)_{1 \times N}}{\left[ {{K^{\boldsymbol{\theta}} }(\boldsymbol{X},\boldsymbol{X}) + \sigma _n^2{\mathbf{I}}} \right]^{ - 1}}_{N \times N}{{\mathbf{y}}_{N \times 1}},\\
		\nabla gp\left( {\boldsymbol{\xi }} \right) 
		&= {\left[ K \right]_{1 \times N \times n}}{\left[ {{K^{\boldsymbol{\theta}} }(\boldsymbol{X},\boldsymbol{X}) + \sigma _n^2{\mathbf{I}}} \right]^{ - 1}}_{N \times N}{{\mathbf{y}}_{N \times 1}},
	\end{align}
\end{subequations}
where ${\left[ K \right]_{1 \times N \times n}}$ is a third-order tensor, and its	components can be expressed analytically as	
\begin{equation}
	{\left[ K \right]_{1j}} = {\nabla _{\boldsymbol{\xi }}}{K^{\boldsymbol{\theta}} }{\left( {{\boldsymbol{\xi }},{{\boldsymbol{\xi }}_j}} \right)_{n \times 1}}.
\end{equation}	
Further, the integral vector field ${{\mathbf{f}}_1}({\boldsymbol{\xi }})$ can be obtained analytically by taking the gradient of ${V_1}\left( {{\boldsymbol{\xi }},{a^*},{b^*},{\boldsymbol{\theta} _1}^*} \right)$.
The integral path which represents the geometrical features of the demonstration trajectory can be obtained by integrating over the vector field ${{\mathbf{f}}_1}({\boldsymbol{\xi }})$ (see Figure~\ref{fig:IntPathL}):
\begin{subequations}
	\begin{align}
		{\boldsymbol{\dot \xi }} = {{\mathbf{f}}_1}({\boldsymbol{\xi }}) 
		&=  - \nabla {V_1}\left( {{\boldsymbol{\xi }},{a^*},{b^*},{\boldsymbol{\theta} _1}^*} \right),\\
		\left\{ {{\boldsymbol{\xi }}_i^{{\text{sim}}}} \right\}_{i = 1}^{{N_s}} 
		&= {\text{integ}}\left( {{{\mathbf{f}}_1}({\boldsymbol{\xi }}),{{{\boldsymbol{\bar \xi }}}_{{\text{init}}}},[{t_s},{t_e}]} \right).
	\end{align}
\end{subequations}	
${{{\boldsymbol{\bar \xi }}}_{{\text{init}}}}$ is the average starting point of the demonstration trajectories, which is used as the initial integration point. $[{t_s},{t_e}]$ is the integration interval. $\left\{ {{\boldsymbol{\xi }}_i^{{\text{sim}}}} \right\}_{i = 1}^{{N_s}} $ is the set of equally spaced sampled points (control points) on the integration path.
To enhance the subsequent optimization results in Section~\ref{subsubsec:Step3}, the reference value of the demonstration trajectory can be updated to $\left\{ {{\boldsymbol{\xi }}_i^{{\text{ref}}},{\boldsymbol{\dot \xi }}_i^{{\text{ref}}}} \right\}_{i = 1}^{{N_s}} = \left\{ {{\boldsymbol{\xi }}_i^{{\text{sim}}},{{\mathbf{f}}_1}\left( {{\boldsymbol{\xi }}_i^{{\text{sim}}}} \right)} \right\}_{i = 1}^{{N_s}}$.

Through the above process, we obtain the integral path based on GP. It is worth noting that we can obtain the integral path by other efficient methods, such as the LPV-DS ${{\mathbf{f}}_{{\text{LPV}}}}\left( {\boldsymbol{\xi }} \right)$~\citep{figueroa2018phyconsis}:
\begin{equation}
	\left\{ {{\boldsymbol{\xi }}_i^{{\text{sim}}}} \right\}_{i = 1}^{{N_s}} = {\text{integ}}\left( {{{\mathbf{f}}_{{\text{LPV}}}}({\boldsymbol{\xi }}),{{{\boldsymbol{\bar \xi }}}_{{\text{init}}}},[{t_s},{t_e}]} \right).
\end{equation}
For different types of demonstration trajectories, different methods have different results. The appropriate method can be selected according to the specific case.

\subsection{Construction of the Potential Function}\label{subsubsec:Step2}

\begin{figure*}[!ht]
	\begin{center}
		\subfigure[\label{fig:XSimAll}]
		{\includegraphics[width=0.5\columnwidth]{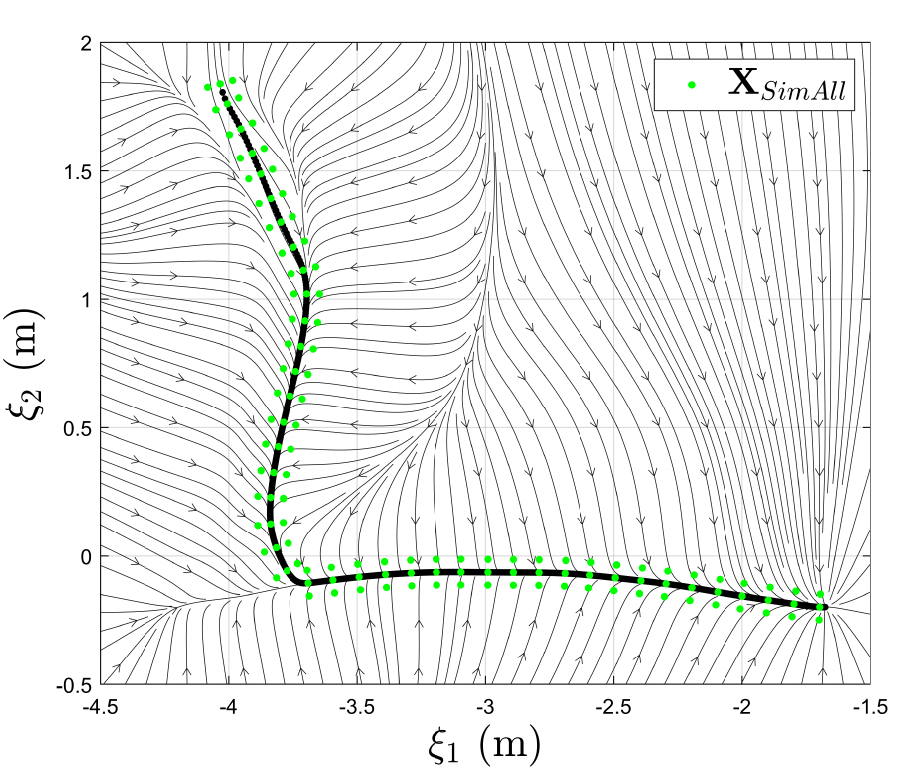}}%
		\hspace{0.00cm}
		\subfigure[\label{fig:V21L}]
		{\includegraphics[width=0.5\columnwidth]{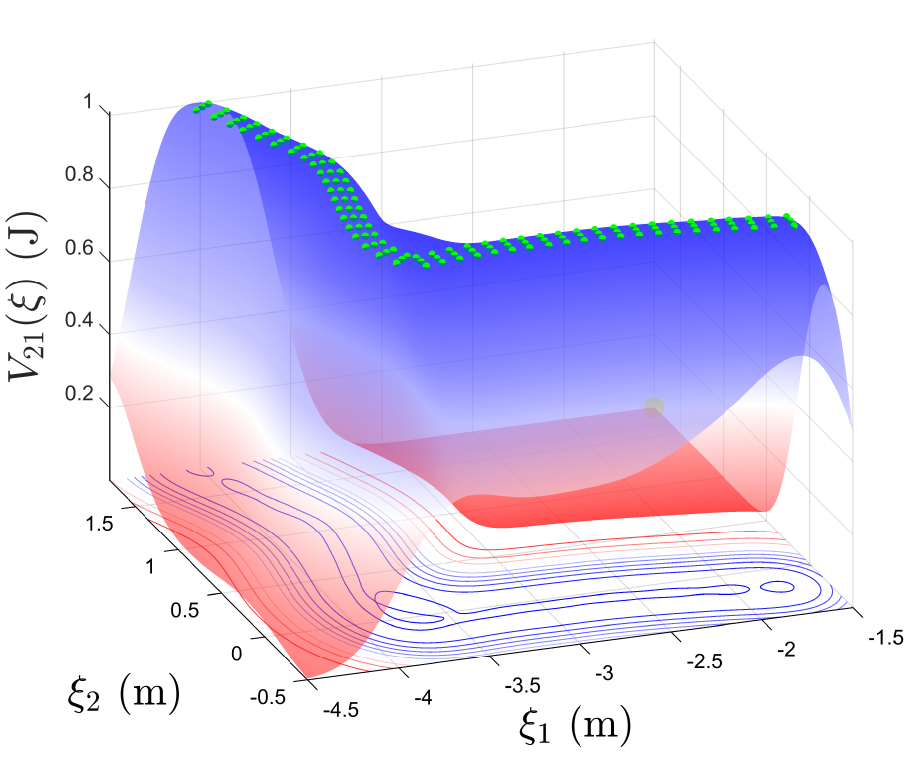}}
		\vspace{0.00cm}		
		\subfigure[\label{fig:V22L}]
		{\includegraphics[width=0.5\columnwidth]{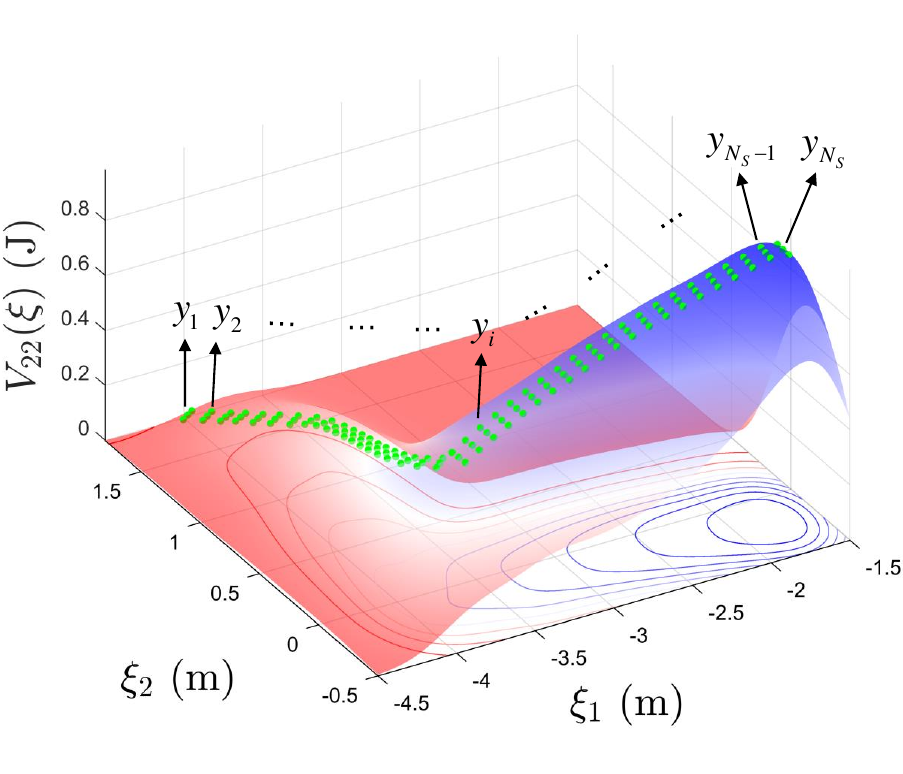}}%
		\hspace{0.00cm}
		\subfigure[\label{fig:V2L}]
		{\includegraphics[width=0.5\columnwidth]{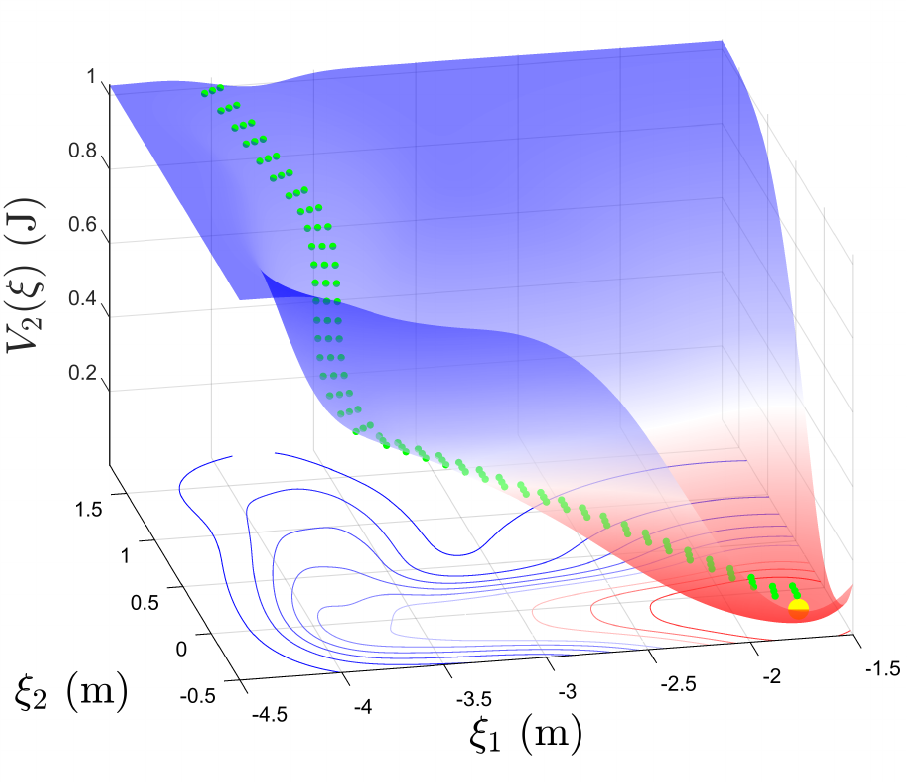}}
		\vspace{0.00cm}	
	\end{center}
	\caption{\label{fig:GPRamp} {Schematic of GP-Ramp function ${V_2}\left( {\boldsymbol{\xi }} \right)$ generation in Section~\ref{subsubsubsec:Step21}. The green dots in (a) denote the expanded point set ${\boldsymbol{X}_{{\text{SimAll}}}}$ and the black line denotes the integration path. (b) and (c) are the essential functions that make up (d). (a) The expanded point set ${\boldsymbol{X}_{{\text{SimAll}}}}$. (b) The function ${V_{21}}\left( {\boldsymbol{\xi }} \right)$. (c) The modulation function ${V_{22}}\left( {\boldsymbol{\xi }} \right)$. (d) The GP-Ramp function ${V_{2}}\left( {\boldsymbol{\xi }} \right)$.}}
\end{figure*}

In Section~\ref{subsubsec:Step1}, we directly obtain the notch function ${V_{{\text{tank}}}}\left( {\boldsymbol{\xi }} \right)$ based on the demonstration path $\left\{ {{\boldsymbol{\xi }}_i^{{\text{ref}}}} \right\}_{i = 1}^N$. Since the actual demonstration trajectory has some noise and perturbation, it is difficult for the notch function to flexibly characterize an arbitrary velocity profile $\left\{ {{\boldsymbol{\dot \xi }}_i^{{\text{ref}}}} \right\}_{i = 1}^N$ and satisfy the stability constraints~(\ref{eq:stabilitycons}).

In this section, we will utilize the control points $\left\{ {{\boldsymbol{\xi }}_i^{{\text{sim}}}} \right\}_{i = 1}^{{N_s}} $ obtained in Section~\ref{subsubsec:Step1} to construct the potential energy function ${V_p}$ corresponding to the target DS ${{\mathbf{f}}_p} =  - {\nabla _{\boldsymbol{\xi }}}{V_p}$. $\left\{ {{\boldsymbol{\xi }}_i^{{\text{sim}}}} \right\}_{i = 1}^{{N_s}} $ obtained from the integration has more regular geometric features, making ${V_p}$ smooth and easy to regulate.

\subsubsection{\texorpdfstring{Construction of the GP-Ramp Function ${V_2}$}{}} \label{subsubsubsec:Step21}
$\\$
We first introduce the construction of functions for the low-dimensional motion case ${\boldsymbol{\xi }} \in {\mathbb{R}^2}$, and for more complex cases we will present them in Section~\ref{sec:generalization}.

We wish the robot to move at a reference velocity along the integration path while resisting perturbation in the vertical direction of the reference velocity. To make the potential energy function more smooth and regular in the vertical direction of the reference velocity, we can expand the sampling points in that direction.
We denote all control points as the matrix
${\boldsymbol{X}_{{\text{sim}}}}\boldsymbol{ = }[{\boldsymbol{\xi }}_1^{{\text{sim}}}{\text{ }}{\boldsymbol{\xi }}_2^{{\text{sim}}}{\text{ }} \cdots {\text{ }}{\boldsymbol{\xi }}_{{N_s}}^{{\text{sim}}}]$. For each point ${\boldsymbol{\xi }}$ on the integration path, the corresponding velocity is
${{\mathbf{f}}_1}({\boldsymbol{\xi }}) = {\left[ {f_1^x({\boldsymbol{\xi }}){\text{ }}f_1^y({\boldsymbol{\xi }})} \right]^T}$, and its perpendicular direction is
${{\mathbf{n}}}({\boldsymbol{\xi }}) = \frac{{{{\mathbf{m}}}({\boldsymbol{\xi }})}}{{\left\| {{{\mathbf{m}}}({\boldsymbol{\xi }})} \right\|}}$, 
where
${{\mathbf{m}}}({\boldsymbol{\xi }}) = {\left[ { - f_1^y({\boldsymbol{\xi }}){\text{ }}f_1^x({\boldsymbol{\xi }})} \right]^T}$.
We denote the normal corresponding to all control points as ${\mathbf{N}} = [{{\mathbf{n}}}({\boldsymbol{\xi }}_1^{{\text{sim}}}){\text{ }}{{\mathbf{n}}}({\boldsymbol{\xi }}_2^{{\text{sim}}}){\text{ }} \cdots {\text{ }}{{\mathbf{n}}}({\boldsymbol{\xi }}_{{N_s}}^{{\text{sim}}})]$. The expanded point set (see the green dots in Figure~\ref{fig:XSimAll}) can be denoted as ${\boldsymbol{X}_{{\text{SimAll}}}} = [{\boldsymbol{X}_{{\text{sim}}}}{\text{, }}{\boldsymbol{X}_{{\text{sim}}}}{\text{ + }}\mu {\mathbf{N}}{\text{, }}{\boldsymbol{X}_{{\text{sim}}}} - \mu {\mathbf{N}}]$, where $\mu$ is a distance factor. Here, we triple the set of control points, and other expansions are also optional.
In GP model, the observed target values ${\mathbf{y}} = {\left[ {{y_{1{\text{ }}}}{y_2} \cdots {\text{ }}{y_{{N_s}}}} \right]^T}$ corresponding to the training points ${\boldsymbol{X}_{{\text{sim}}}}$ also needs to be tripled:
\begin{equation}
	\label{eq:tripley}
	{{\mathbf{y}}_{{\text{total}}}} = \left[ {\begin{array}{*{20}{c}}
			{\mathbf{y}} \\ 
			{\mathbf{y}} \\ 
			{\mathbf{y}} 
	\end{array}} \right] = \left[ {\begin{array}{*{20}{c}}
			{\mathbf{I}} \\ 
			{\mathbf{I}} \\ 
			{\mathbf{I}} 
	\end{array}} \right]{\mathbf{y}}.
\end{equation}
The expanded target values (\ref{eq:tripley}) means that every three points $\left\{ {{\boldsymbol{\xi }}_i^{{\text{sim}}},{\boldsymbol{\xi }}_i^{{\text{sim}}} + \mu {\mathbf{n}}({\boldsymbol{\xi }}_i^{{\text{sim}}}),{\boldsymbol{\xi }}_i^{{\text{sim}}} - \mu {\mathbf{n}}({\boldsymbol{\xi }}_i^{{\text{sim}}})} \right\}$ correspond to the same target value ${y_i}$, as shown in Figure~\ref{fig:V22L}.

Denote ${\mathbf{\Gamma }}  = {\left[ {0{\text{ 0 }} \cdots {\text{ 0 1}}} \right]_{1 \times {N_s}}}$, then ${y_{{N_s}}} = {\mathbf{\Gamma y}}$. The GP-Ramp function ${V_2}$ is constructed as follows:
\begin{equation}
	\label{eq:V2}
	{V_2}\left( {\boldsymbol{\xi },{\boldsymbol{\theta} _2},{\mathbf{y}}} \right) = \left( {1 - {r_a}\left( {\boldsymbol{\xi }} \right)} \right)\left( {\varepsilon  + {\mathbf{\Gamma }}{\mathbf{y}} - {V_{21}}\left( {\boldsymbol{\xi }} \right){V_{22}}\left( {\boldsymbol{\xi },{\boldsymbol{\theta} _2},{\mathbf{y}}} \right)} \right),
\end{equation}
where
\begin{equation}
	\label{eq:V21}
	{V_{21}}\left( {\boldsymbol{\xi }} \right) = gp\left( {{\boldsymbol{\theta} _1},{\boldsymbol{X}_{{\text{SimAll}}}},{{\mathbf{I}}_{3{N_s} \times 1}},{\boldsymbol{\xi }}} \right),
\end{equation}
\begin{equation}
	\label{eq:V22}
	\begin{aligned}
		&{V_{22}}\left( {\boldsymbol{\xi },{\boldsymbol{\theta} _2},{\mathbf{y}}} \right) 
		= gp\left( {{\boldsymbol{\theta} _2},{\boldsymbol{X}_{{\text{SimAll}}}},{{\mathbf{y}}_{{\text{total}}}},{\boldsymbol{\xi }}} \right) \\
		&= {{K^{{\boldsymbol{\theta} _2}}}\left( {{\boldsymbol{\xi }},{\boldsymbol{X}_{{\text{SimAll}}}}} \right){{\left[ {{K^{{\boldsymbol{\theta} _2}}}({\boldsymbol{X}_{{\text{SimAll}}}},{\boldsymbol{X}_{{\text{SimAll}}}}) + \sigma _n^2{\mathbf{I}}} \right]}^{ - 1}}\left[ {\begin{array}{*{20}{c}}
					{\mathbf{y}} \\ 
					{\mathbf{y}} \\ 
					{\mathbf{y}} 
			\end{array}} \right]}.
	\end{aligned}
\end{equation}
The function ${V_{21}}$ is obtained by setting all the target values ${\mathbf{y}}$ to 1, as shown in Figure~\ref{fig:V21L}. ${V_{21}}$ is used as a penalty factor to make ${V_2}$ smoother on the integration path.
The function ${V_{22}}$ realizes the adjustment of the slope of ${V_2}$ on the integration path by modulating the corresponding target value ${\mathbf{y}}$ at the control point ${\boldsymbol{X}_{{\text{sim}}}}$, as shown in Figure~\ref{fig:V22L}.
The coefficient $\left( {1 - {r_a}\left( {\boldsymbol{\xi }} \right)} \right)$ makes the values of the function ${V_2}$ and its derivative at ${{\boldsymbol{\xi }}_0}$ are zero, as shown in Figure~\ref{fig:V2L}.

By observing (\ref{eq:V2})-(\ref{eq:V22}), an important fact is that ${V_2}$ is a linear function with respect to ${\mathbf{y}}$. This property will be used in the solution of subsequent optimization problem.

Considering that the objective of ${V_2}$ is to make the robot move along the integration path and stop at the equilibrium point $\boldsymbol{\xi}_0$, the following constraints on the objective value are given as
${\text{0}} \leqslant {y_1}<{y_2}<{y_3}< \cdots <{y_{{N_s} - 1}} = {y_{{N_s}}}$.
It is written in matrix form as follows:
\begin{equation}
	{{\mathbf{\Gamma }}_1}{\mathbf{y}} 
	= {\left[ {\begin{array}{*{20}{c}}
				0&0& \cdots &0&1&{ - 1} 
		\end{array}} \right]_{1 \times {N_s}}}{\mathbf{y}} = {\mathbf{0}},
\end{equation}
\begin{equation}
	\begin{aligned}
	&{{\mathbf{\Gamma }}_2}{\mathbf{y}} \\ 
	&= {\left[ {\begin{array}{*{20}{c}}
				1&{ - 1}&0&0& \cdots &0&0 \\ 
				0&1&{ - 1}&0& \cdots &0&0 \\ 
				0&0&1&{ - 1}& \ddots & \vdots & \vdots  \\ 
				\vdots & \ddots & \ddots & \ddots & \ddots &0& \vdots  \\ 
				0& \cdots &0&0&1&{ - 1}&0 
		\end{array}} \right]_{\left( {{N_s} - 2} \right) \times {N_s}}}{\mathbf{y}} < {\mathbf{0}}.
	\end{aligned}
\end{equation}

\subsubsection{\texorpdfstring{Construction of the Base Function $\Phi $}{}}
$\\$
The GP-Ramp function ${V_2}$ in Section~\ref{subsubsubsec:Step21} has flexible slope regulation. However, due to the property of the covariance function 
$k^{\boldsymbol{\theta }}({{\boldsymbol{\xi }},{{\boldsymbol{\xi }}^{'}}})$, ${V_2}$ becomes flat as $\boldsymbol{\xi}$ moves away from the demonstration trajectory, as shown in Figure~\ref{fig:V2L}. In this section, we construct the base function $\Phi $ to "attract" the distant state point $\boldsymbol{\xi}$. Moreover, the function can exhibit variable stiffness properties on the demonstration trajectory to exhibit different symmetric attractiveness for different regions.

Based on the control points $\left\{ {{\boldsymbol{\xi }}_i^{{\text{sim}}}} \right\}_{i = 1}^{{N_s}} $, the base function is constructed as
\begin{equation}
	\Phi ({\boldsymbol{\xi }})
	=\sum\limits_{i = 1}^{{N_s}} {{{\tilde \omega }_i}({\boldsymbol{\xi }}){\phi _i}({\boldsymbol{\xi }})},
\end{equation}
where
\begin{subequations}
	\begin{align}
		{\phi _i}({\boldsymbol{\xi }}) 
		&=\frac{1}{2}{\left( {{\boldsymbol{\xi }} - {{\boldsymbol{\xi }}_i^{{\text{sim}}}}} \right)^T}{{\mathbf{S}}_i}\left( {{\boldsymbol{\xi }} - {{\boldsymbol{\xi }}_i^{{\text{sim}}}}} \right),\\
		{\omega _i}({\boldsymbol{\xi }})
		&= \exp \left( { - \frac{1}{{2{\delta _i}^2}}{{\left( {{\boldsymbol{\xi }} - {{\boldsymbol{\xi }}_i^{{\text{sim}}}}} \right)}^T}\left( {{\boldsymbol{\xi }} - {{\boldsymbol{\xi }}_i^{{\text{sim}}}}} \right)} \right),\\
		{\tilde \omega _i}({\boldsymbol{\xi }})
		&= \frac{{{\omega _i}({\boldsymbol{\xi }})}}{{\sum\limits_{j = 1}^{{N_s}} {{\omega _j}({\boldsymbol{\xi }})} }}.
	\end{align}
\end{subequations}	
${{\mathbf{S}}_i}$ can be regarded as the equivalent stiffness at the control point $\boldsymbol{\xi}_i^{{\text{sim}}}$, and the tuning of the attraction stiffness can be realized flexibly by changing the value of ${{\mathbf{S}}_i}$. $\delta_i$ is the length-scale factor.

The gradient of the base function $\Phi ({\boldsymbol{\xi }})$ is easily derived analytically for subsequent parameter optimization and computation of the DS:
\begin{equation}
	\begin{aligned}
		\nabla \Phi ({\boldsymbol{\xi }}) 
		&= \sum\limits_{i = 1}^{{N_s}} {{{\tilde \omega }_i}({\boldsymbol{\xi }}){{\mathbf{S}}_i}\left( {{\boldsymbol{\xi }} - {{\boldsymbol{\xi }}_i^{{\text{sim}}}}} \right)}  \hfill \\
		& - \sum\limits_{i = 1}^{{N_s}} {\frac{1}{{{\delta _i}^2}}{{\tilde \omega }_i}({\boldsymbol{\xi }})\left( {{\phi _i}({\boldsymbol{\xi }}) - \Phi ({\boldsymbol{\xi }})} \right)\left( {{\boldsymbol{\xi }} - {{\boldsymbol{\xi }}_i^{{\text{sim}}}}} \right)}.
	\end{aligned}
\end{equation}		

\subsubsection{\texorpdfstring{Construction of the Total Potential Function ${V_p}$}{}}
$\\$
From the previous description, base functioon $\Phi $ attracts the robot toward the integration (demonstration) path, and next ${V_2}$ moves the robot along the integration path at the reference velocity.
Thus, the total potential function ${V_p}$ (see Figure~\ref{fig:LyapL}) and the corresponding target DS ${{\mathbf{f}}_p}$ (see Figure~\ref{fig:ConDSL}) can be expressed as
\begin{subequations}
	\begin{align}
		{V_p}\left( {{\boldsymbol{\xi }},{\boldsymbol{\theta} _2},{\mathbf{y}}} \right) 
		&= {g_p}\left( {{\boldsymbol{\xi }},{\boldsymbol{\theta} _2},{\mathbf{y}}} \right) 
		= \Phi ({\boldsymbol{\xi }}) + {V_{2}}\left( {{\boldsymbol{\xi }},{\boldsymbol{\theta} _2},{\mathbf{y}}} \right)\\
		{{\mathbf{f}}_p}\left( {{\boldsymbol{\xi }},{\boldsymbol{\theta} _2},{\mathbf{y}}} \right) 
		&=  - {\nabla _{\boldsymbol{\xi }}}{V_p}\left( {{\boldsymbol{\xi }},{\boldsymbol{\theta} _2},{\mathbf{y}}} \right).
	\end{align}
\end{subequations}

\subsection{Solution of the Optimization Problem}\label{subsubsec:Step3}

\begin{figure*}[!h]
	\begin{center}
		\subfigure[\label{fig:LyapL}]
		{\includegraphics[width=0.60\columnwidth]{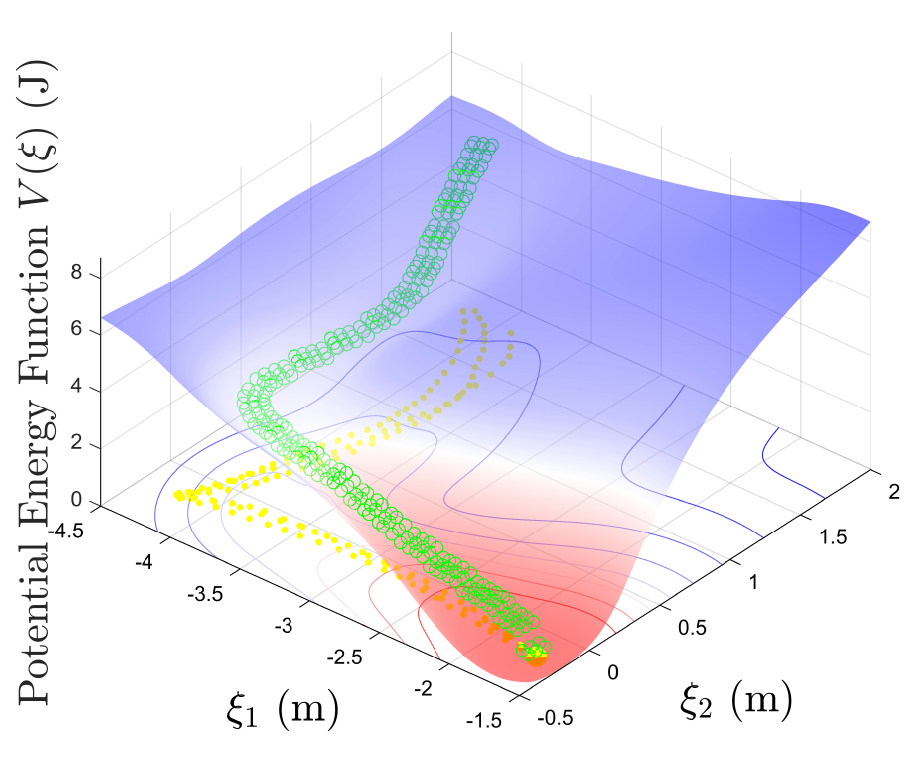}}%
		\hspace{0.00cm}
		\subfigure[\label{fig:ConDSL}]
		{\includegraphics[width=0.60\columnwidth]{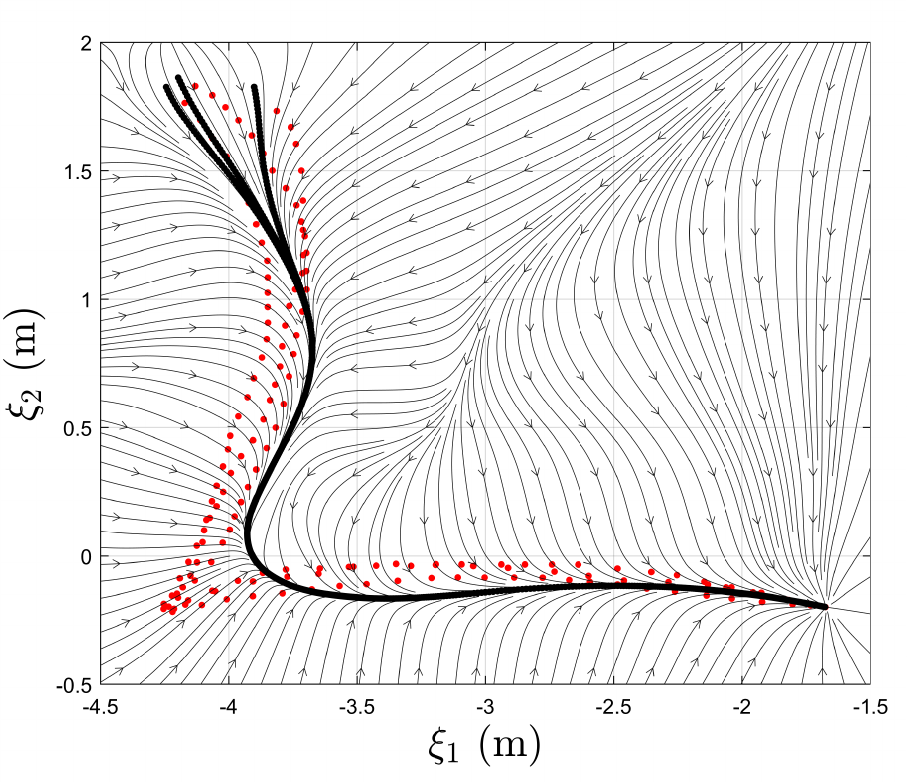}}
		\hspace{0.00cm}
		\subfigure[\label{fig:VelConDSL}]
		{\includegraphics[width=0.8\columnwidth]{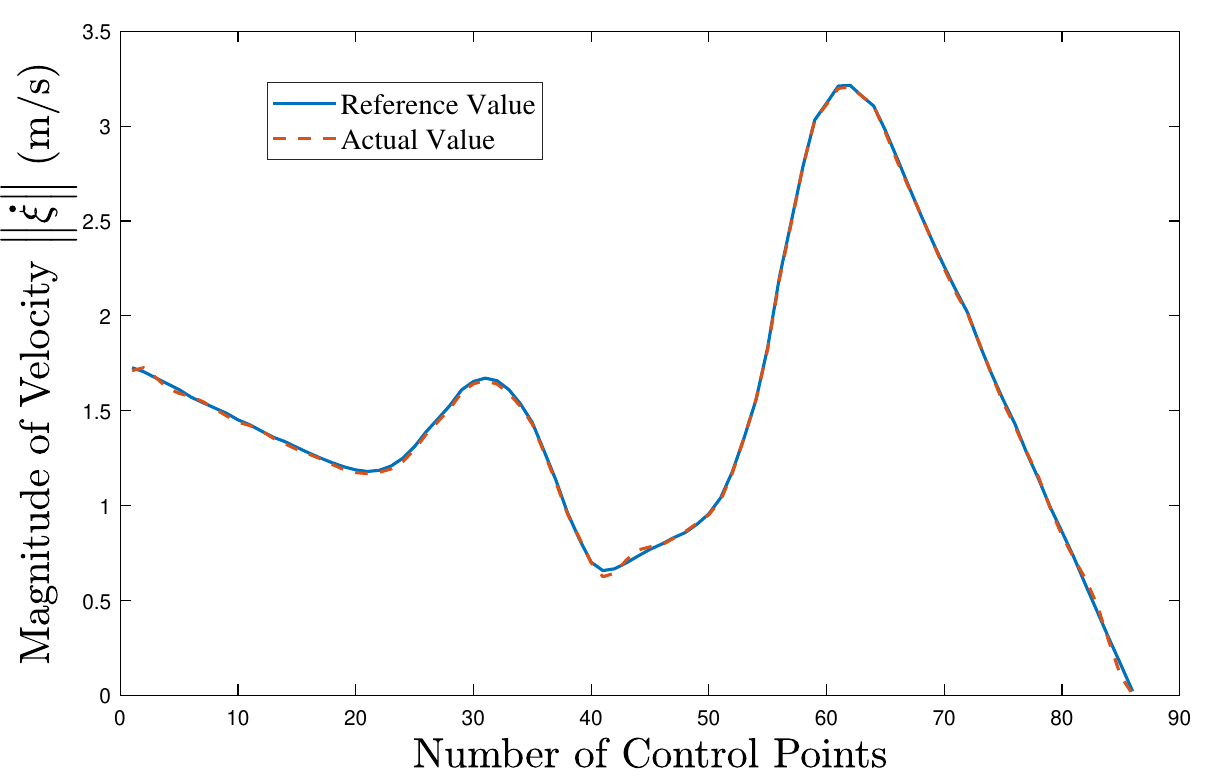}}
		\vspace{0.00cm}		
	\end{center}
	\caption{\label{fig:GPDSL} {Simulation of DS generation on the L-shaped trajectory. The process and details of the DS generation based on GP are displayed in (a), (b) and (c). The red dots represent demonstration trajectories. (a) The total potential function ${V_p}$ based on GP. (b) DS ${\boldsymbol{\dot \xi }} = {{\mathbf{f}}_p}\left( {{\boldsymbol{\xi }},{{\boldsymbol{\theta }}_2},{\mathbf{y}}} \right)$ based on GP. (c) The velocity comparison. The actual velocity represents the result of learning through the GP.}}
\end{figure*}

As can be seen from the previous subsections, the potential energy function ${V_p}\left( {{\boldsymbol{\xi }},{\boldsymbol{\theta} _2},{\mathbf{y}}} \right)$ can be flexibly controlled by adjusting the corresponding target values ${\mathbf{y}}$ at control points $\left\{ {{\boldsymbol{\xi }}_i^{{\text{sim}}}} \right\}_{i = 1}^{{N_s}} $ as well as the hyperparameters ${\boldsymbol{\theta} _2}$ of the GP. 

By the previous construction, $\nabla g_p\left( {{{\boldsymbol{\xi }}_0}} \right)=\mathbf{0}$ and $g_p\left( {{{\boldsymbol{\xi }}_0}} \right)=0$. The stability constraints~(\ref{eq:stabilitycons}) are simplified to
\begin{subequations}
	\label{eq:stabilitycons1}
	\begin{align}
		g_p\left( {\boldsymbol{\xi }} \right) > 0,{\text{  }}{\boldsymbol{\xi }} \ne {{\boldsymbol{\xi }}_0}\\
		\left\| {\nabla g_p\left( {\boldsymbol{\xi }} \right)} \right\| > 0,{\text{  }}{\boldsymbol{\xi }} \ne {{\boldsymbol{\xi }}_0}.
	\end{align}
\end{subequations}
The goal of the optimization is to get as much consistency as possible between the constructed DS ${\boldsymbol{\dot \xi }}_i^{{\text{GP}}} = {{\mathbf{f}}_p}\left( {{\boldsymbol{\xi }}_i^{{\text{ref}}},{\boldsymbol{\theta} _2},{\mathbf{y}}} \right)$ and the reference velocity ${\boldsymbol{\dot \xi }}_i^{{\text{ref}}}$. 
We select the key region $\mathcal{C} \subset {\mathbb{R}^n}$ to impose stability constraint (\ref{eq:stabilitycons1}). Constraint (\ref{eq:stabilitycons1}) is applied to the sampled points $\left\{ {{\boldsymbol{\xi }}_i^{{\text{sam}}}} \right\}_{i = 1}^{{N_a}}$ by sampling $N_a$ suitable points in the region $\mathcal{C}$.
The well-chosen sampling of points $\left\{ {{\boldsymbol{\xi }}_i^{{\text{sam}}}} \right\}_{i = 1}^{{N_a}}$ can generalize the incorporated discrete constraints to the continuous region $\mathcal{C}$~\citep{Lemme2014neuralds}.
Combining the above description and also considering the monotonicity constraint on $\mathbf{y}$, the optimization problem can be built in the following form:
\begin{equation}
	\label{opt:optim0}
	\begin{aligned}
		\mathop {\min }\limits_{\left\lbrace {\boldsymbol{\theta} _2}, \mathbf{y}\right\rbrace } {\text{  }}
		&\frac{1}{N}\sum\limits_{i = 1}^N {{{\left\| { - \nabla {V_p}\left( {{\boldsymbol{\xi }}_i^{{\text{ref}}},{\boldsymbol{\theta} _2},{\mathbf{y}}} \right) - {\boldsymbol{\dot \xi }}_i^{{\text{ref}}}} \right\|}^2}} \\
		\mathit{s.t.}
		~&{{\mathbf{\Gamma }}_1}{\mathbf{y}} = {\mathbf{0}},\\
		&{{\mathbf{\Gamma }}_2}{\mathbf{y}} < {\mathbf{0}},\\
		&{g_p}\left( {{\boldsymbol{\xi }}_j^{{\text{sam}}},{\boldsymbol{\theta} _2},{\mathbf{y}}} \right) > 0,{\text{ }}\left( {j = 1,2, \cdots ,{N_a}} \right) \\
		&\left\| {\nabla {g_p}\left( {{\boldsymbol{\xi }}_j^{{\text{sam}}},{\boldsymbol{\theta} _2},{\mathbf{y}}} \right)} \right\| > 0.{\text{ }}\left( {j = 1,2, \cdots ,{N_a}} \right)
	\end{aligned} 
\end{equation}    
The objective function and constraints of the optimization problem (\ref{opt:optim0}) are highly nonlinear functions with respect to the optimization variables ${\left\lbrace {\boldsymbol{\theta} _2}, \mathbf{y}\right\rbrace }$. Solving them directly is tedious and time-consuming.

Notice that $V_2$ is a linear function of ${\mathbf{y}}$, so $V_p$ and its gradient ${\nabla _{\boldsymbol{\xi }}}{V_p}$ are also linear functions with respect to ${\mathbf{y}}$, which can be expressed as:
\begin{subequations}
	\label{eq:lineareq}
	\begin{align}
		{V_p}\left( {{\boldsymbol{\xi }},{\boldsymbol{\theta} _2},{\mathbf{y}}} \right) 
		&= {\mathbf{a}}\left( {{\boldsymbol{\xi }},{\boldsymbol{\theta} _2}} \right){\mathbf{y}} + {\mathbf{b}}\left( {{\boldsymbol{\xi }},{\boldsymbol{\theta} _2}} \right),\\
		{{\mathbf{f}}_p}\left( {{\boldsymbol{\xi }},{\boldsymbol{\theta} _2},{\mathbf{y}}} \right) 
		&=  - {\nabla _{\boldsymbol{\xi }}}{V_p}\left( {{\boldsymbol{\xi }},{\boldsymbol{\theta} _2},{\mathbf{y}}} \right) = {\mathbf{A}}\left( {{\boldsymbol{\xi }},{\boldsymbol{\theta} _2}} \right){\mathbf{y}} + {\mathbf{B}}\left( {{\boldsymbol{\xi }},{\boldsymbol{\theta} _2}} \right),
	\end{align}
\end{subequations}
where ${\mathbf{a}}\left( {{\boldsymbol{\xi }},{\boldsymbol{\theta} _2}} \right)$, ${\mathbf{b}}\left( {{\boldsymbol{\xi }},{\boldsymbol{\theta} _2}} \right)$, ${\mathbf{A}}\left( {{\boldsymbol{\xi }},{\boldsymbol{\theta} _2}} \right)$ and ${\mathbf{B}}\left( {{\boldsymbol{\xi }},{\boldsymbol{\theta} _2}} \right)$ are the coefficients of linear equations (\ref{eq:lineareq}).

Therefore, when fixing ${\boldsymbol{\theta} _2}$, the above optimization problem can be regarded as a quadratic constrained quadratic programming (QCQP) problem ${{\mathbf{P}}_1}$ with respect to ${\mathbf{y}}$:
\begin{equation}
	\label{opt:optim1}
	\begin{aligned}
		\mathop {\min }\limits_{\mathbf{y}} {\text{  }}
		&\frac{1}{N}\sum\limits_{i = 1}^N {{{\left\| {{\mathbf{A}}\left( {{\boldsymbol{\xi }}_i^{{\text{ref}}},{\boldsymbol{\theta} _2}} \right){\mathbf{y}} + {\mathbf{B}}\left( {{\boldsymbol{\xi }}_i^{{\text{ref}}},{\boldsymbol{\theta} _2}} \right) - {\boldsymbol{\dot \xi }}_i^{{\text{ref}}}} \right\|}^2}}  \\
		\mathit{s.t.}
		~&{{\mathbf{\Gamma }}_1}{\mathbf{y}} = {\mathbf{0}},\\
		&{{\mathbf{\Gamma }}_2}{\mathbf{y}} < {\mathbf{0}},\\
		&{\mathbf{a}}\left( {{\boldsymbol{\xi }}_j^{{\text{sam}}},{\boldsymbol{\theta} _2}} \right){\mathbf{y}} + {\mathbf{b}}\left( {{\boldsymbol{\xi }}_j^{{\text{sam}}},{\boldsymbol{\theta} _2}} \right) > 0,{\text{ }}\left( {j = 1, \cdots ,{N_a}} \right) \\
		&{\left\| {{\mathbf{A}}\left( {{\boldsymbol{\xi }}_j^{{\text{sam}}},{\boldsymbol{\theta} _2}} \right){\mathbf{y}} + {\mathbf{B}}\left( {{\boldsymbol{\xi }}_j^{{\text{sam}}},{\boldsymbol{\theta} _2}} \right)} \right\|^2} > 0.{\text{ }}\left( {j = 1,\cdots,{N_a}} \right)
	\end{aligned} 
\end{equation}
The above problem ${{\mathbf{P}}_1}$ can be solved by using the interior-point algorithm with the fmincon solver in MATLAB.

\begin{figure}[!ht]
	\centering
	\includegraphics[width=0.7\columnwidth]{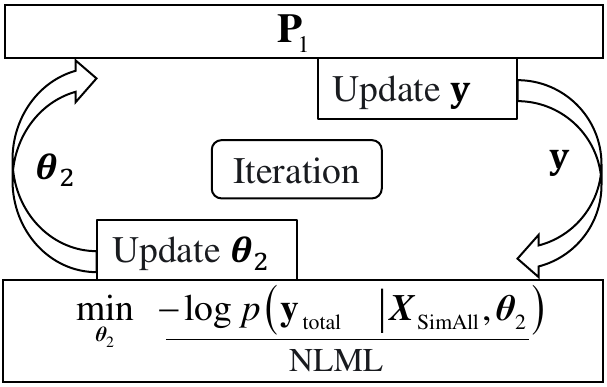}
	\caption{Schematic diagram of the iterative optimization process. Update ${\mathbf{y}}$ by solving the QCQP optimization problem ${{\mathbf{P}}_1}$ (\ref{opt:optim1}) and subsequently update ${\boldsymbol{\theta} _2}$ by solving the optimization problem (\ref{opt:optim2}).}
	\label{fig:OptimIter}
\end{figure}

When we update ${\mathbf{y}}$ by solving ${{\mathbf{P}}_1}$, the hyperparameter ${\boldsymbol{\theta} _2}$ can be subsequently updated by minimizing the negative logarithmic marginal likelihood (NLML):
\begin{equation}
	\label{opt:optim2}
	\mathop {\min }\limits_{{\boldsymbol{\theta} _2}} {-\text{ log }}p\left( {\begin{array}{*{20}{c}}
			{{{\mathbf{y}}_{{\text{total}}}}}&{\left| {{\boldsymbol{X}_{{\text{SimAll}}}},{\boldsymbol{\theta} _2}} \right.} 
	\end{array}} \right). 
\end{equation}
We use the GPML toolbox~\citep{williams2006gaussian} to solve the above optimization problem (\ref{opt:optim2}).
It is worth noting that when ${\boldsymbol{\theta} _2}$ is updated, the optimization model ${{\mathbf{P}}_1}$ (\ref{opt:optim1}) is also updated. Thus, the original optimization problem (\ref{opt:optim0}) is decomposed into an iterative optimization of the parameters ${\mathbf{y}}$ and ${\boldsymbol{\theta} _2}$, as shown in Figure~\ref{fig:OptimIter}. The optimization process ends when the number of iterations reaches the maximum or the change in the optimization parameters ${\left\lbrace {\boldsymbol{\theta} _2}, \mathbf{y}\right\rbrace }$ is less than the set threshold. A series of simulation results in this paper illustrate the effectiveness of the proposed optimization method.

After optimization, the actual velocity ${\boldsymbol{\dot \xi }}_i^{{\text{GP}}} = {{\mathbf{f}}_p}\left( {{\boldsymbol{\xi }}_i^{{\text{ref}}},{\boldsymbol{\theta} _2},{\mathbf{y}}} \right)$ at ${{\boldsymbol{\xi }}_i^{{\text{ref}}}}$ is highly consistent with the reference velocity ${\boldsymbol{\dot \xi }}_i^{{\text{ref}}}$, as shown in Figure~\ref{fig:VelConDSL}. The above results illustrate the effectiveness of the GP-based approach.

%% file: sections/5_general_method.tex

\begin{figure*}[!h]
	\begin{center}
		\subfigure[\label{fig:3DSinkRef}]
		{\includegraphics[width=0.5\columnwidth]{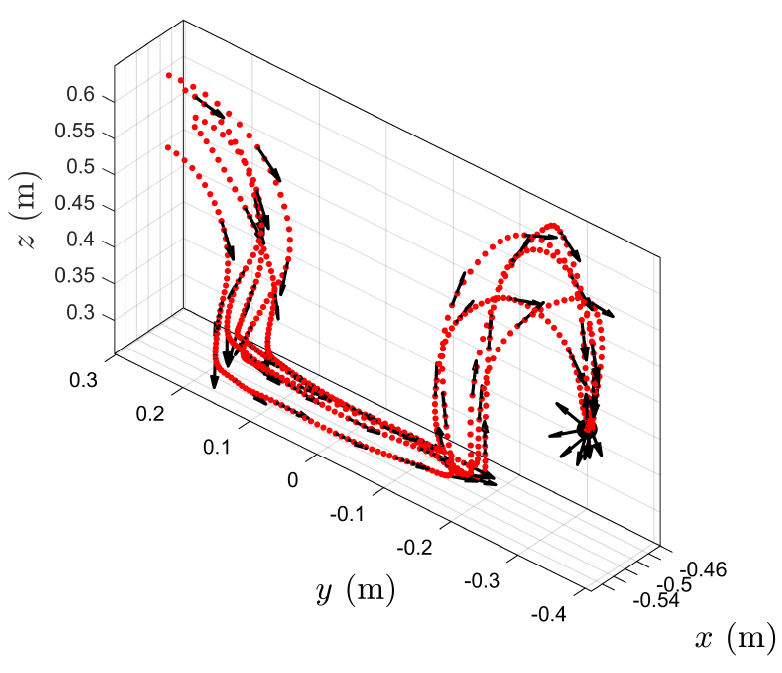}}%
		\hspace{0.00cm}
		\subfigure[\label{fig:3DSinkXSimAll}]
		{\includegraphics[width=0.5\columnwidth]{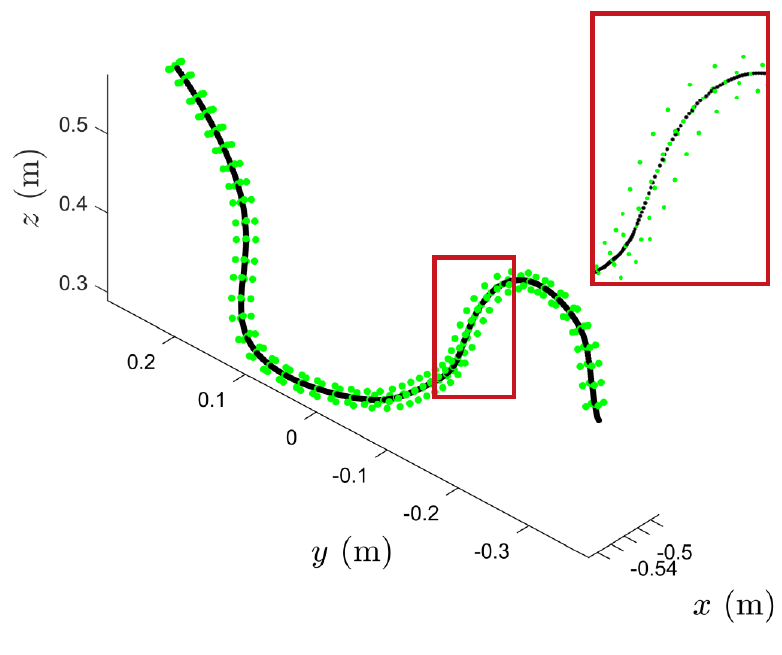}}
		\vspace{0.00cm}		
		\subfigure[\label{fig:3DSinkLyap}]
		{\includegraphics[width=0.5\columnwidth]{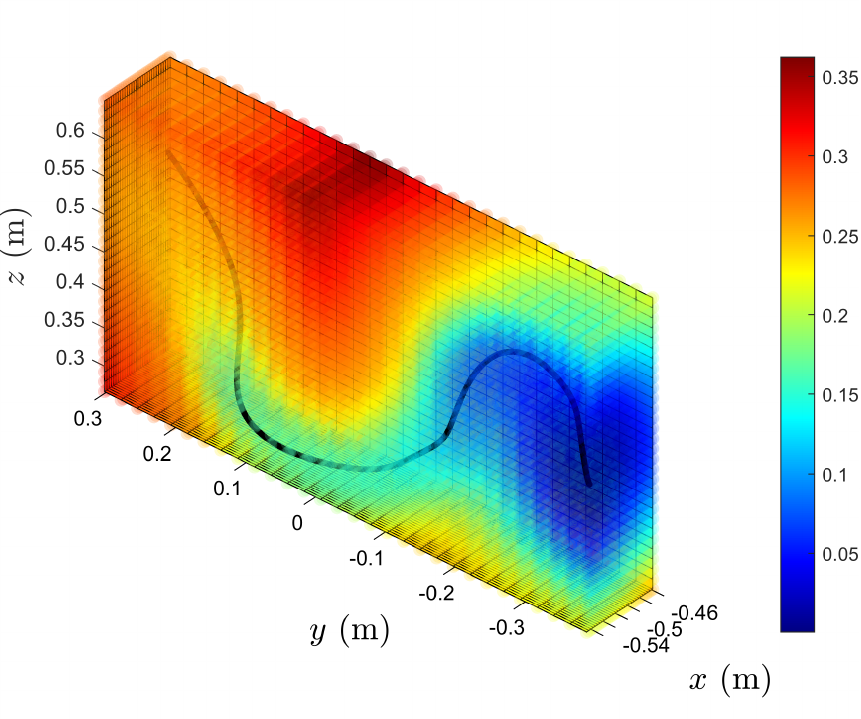}}%
		\hspace{0.00cm}
		\subfigure[\label{fig:3DSinkDS}]
		{\includegraphics[width=0.5\columnwidth]{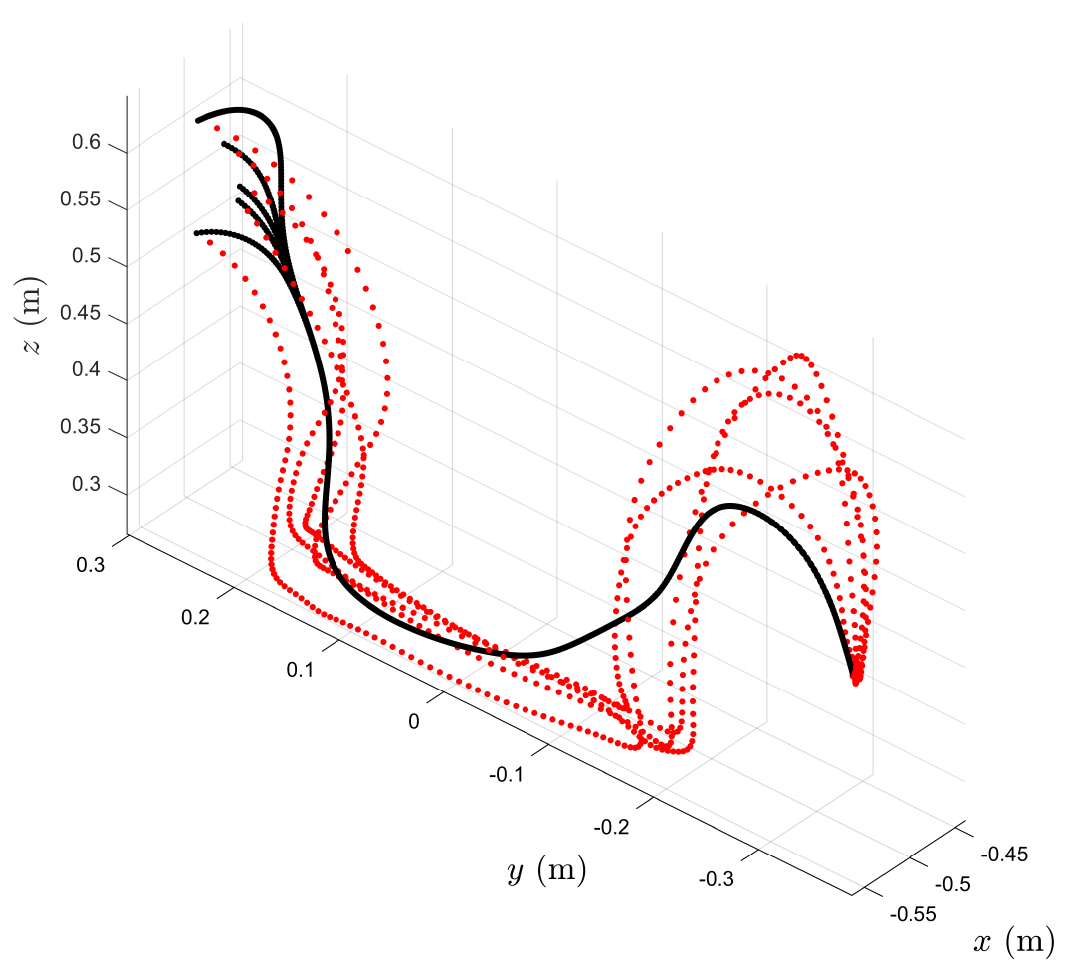}}
		\vspace{0.00cm}	
	\end{center}
	\caption{\label{fig:3DSink} {Schematic of the process of DS generation in high-dimensional space ${\mathbb{R}^{3}}$. The red dots represent demonstration trajectories. The green dots in (b) denote the expanded point set ${\boldsymbol{X}_{{\text{SimAll}}}}$. (a) Reference demonstration trajectory in ${\mathbb{R}^{3}}$. (b) The expanded point set ${\boldsymbol{X}_{{\text{SimAll}}}}$. (c) Potential energy function $V_p$ generated in ${\mathbb{R}^{3}}$. The color indicates the value of the potential energy. (d) DS generated in ${\mathbb{R}^{3}}$. The black lines represent integral curves from random starting points.}}
\end{figure*}

\section{Generalizability of the Proposed Methodology}\label{sec:generalization}
The method for generating conservative DS in low-dimensional flat space is presented in Section~\ref{sec:Conservative DS}.
In this section, we generalize the method introduced in Section~\ref{sec:Conservative DS} to make the method applicable to different manifolds as well as special motion trajectories.
\subsection{The High-Dimension Space }\label{subsubsec:HighDim}
After expanding the set of control points for the case of high-dimensional spaces, the method in Section~\ref{sec:Conservative DS} can be naturally generalized to high-dimensional linear spaces.

Similarly, we denote all control points as the matrix
${\boldsymbol{X}_{{\text{sim}}}}\boldsymbol{ = }[{\boldsymbol{\xi }}_1^{{\text{sim}}}{\text{ }}{\boldsymbol{\xi }}_2^{{\text{sim}}}{\text{ }} \cdots {\text{ }}{\boldsymbol{\xi }}_{{N_s}}^{{\text{sim}}}]$.
For each point ${\boldsymbol{\xi }} \in {\mathbb{R}^n}$ on the integration path, the corresponding velocity is
${{\mathbf{f}}_1}({\boldsymbol{\xi }}) \in {\mathbb{R}^n}$, and there exist $n-1$ linearly independent vectors in its perpendicular direction. Depending on the specific situation, $l$  vertical vectors ${{\mathbf{n}}_i}({\boldsymbol{\xi }}) \bot {{\mathbf{f}}_1}({\boldsymbol{\xi }})$ are selected $\left( {i = 1, \cdots ,l}<{n-1} \right)$. We denote the normal matrix corresponding to the $i$th perpendicular direction as
${{\mathbf{N}}_i} = [{{\mathbf{n}}_i}({\boldsymbol{\xi }}_1^{{\text{sim}}}){\text{ }}{{\mathbf{n}}_i}({\boldsymbol{\xi }}_2^{{\text{sim}}}){\text{ }} \cdots {\text{ }}{{\mathbf{n}}_i}({\boldsymbol{\xi }}_{{N_s}}^{{\text{sim}}})] \text{  } \left( {i = 1, \cdots ,l} \right)$.
Then the expanded control point set (see the green dots in Figure~\ref{fig:3DSinkXSimAll}) can be denoted as
${\boldsymbol{X}_{{\text{SimAll}}}} = [{\boldsymbol{X}_{{\text{sim}}}}{\text{, }}{\boldsymbol{X}_{{\text{sim}}}}{\text{ + }}\mu {{\mathbf{N}}_1}{\text{, }} \cdots {\text{, }}{\boldsymbol{X}_{{\text{sim}}}}{\text{ + }}\mu {{\mathbf{N}}_l}]$.
In GP model, the observed target values ${\mathbf{y}} = {\left[ {{y_{1{\text{ }}}}{y_2} \cdots {\text{ }}{y_{{N_s}}}} \right]^T}$ corresponding to the training points ${\boldsymbol{X}_{{\text{sim}}}}$ also needs to be expanded:
\begin{equation}
	\label{eq:expandy}
	{{\mathbf{y}}_{total}} = {\left[ {\begin{array}{*{20}{c}}
				{\mathbf{y}} \\ 
				\vdots  \\ 
				{\mathbf{y}} 
		\end{array}} \right]_{(l+1) \cdot {N_s} \times 1}} = {\left[ {\begin{array}{*{20}{c}}
				{\mathbf{I}} \\ 
				\vdots  \\ 
				{\mathbf{I}} 
		\end{array}} \right]_{(l+1) \cdot {N_s} \times {N_s}}}{\mathbf{y}}
\end{equation}
The expanded target values (\ref{eq:expandy}) means that every $l+1$ points $\left\{ {{\boldsymbol{\xi }}_i^{{\text{sim}}},{\boldsymbol{\xi }}_i^{{\text{sim}}} + \mu {\mathbf{n}_1}({\boldsymbol{\xi }}_i^{{\text{sim}}}), \cdots,{\boldsymbol{\xi }}_i^{{\text{sim}}} + \mu {\mathbf{n}_l}({\boldsymbol{\xi }}_i^{{\text{sim}}})} \right\}$ correspond to the same target value ${y_i}$, as shown in Figure~\ref{fig:3DSinkLyap}.

For the sake of demonstration, take the 3-dimensional Cartesian space as an example, as shown in Figure~\ref{fig:3DSink}. In this case, we select $l = 4$ for control point expansion, as shown in Figure~\ref{fig:3DSinkXSimAll}. GP-based generated potential function still generates effective attraction regions around the demonstration path in 3D space, as shown in Figure~\ref{fig:3DSinkLyap}. The initial point of motion is randomly selected, and the motion trajectory first rapidly converges to the demonstration trajectory, followed by motion to the equilibrium point. This exhibits a clear symmetric attractiveness of the GP-generated DS, as shown in Figure~\ref{fig:3DSinkDS}.

\subsection{The Rotation motion}
The trajectories in the aforementioned linear space ${\mathbb{R}^n}$ describe translational motion. The motion of the robot system can be described as a smooth curve on the submanifold of $\text{SE(3)}$ as the system undergoes rotation.
To generate a conservative DS on SE(3), we first introduce the following lemma.
\begin{figure*}[!ht]
	\begin{center}
		\subfigure[\label{fig:RotDsgen}]
		{\includegraphics[width=0.55\columnwidth]{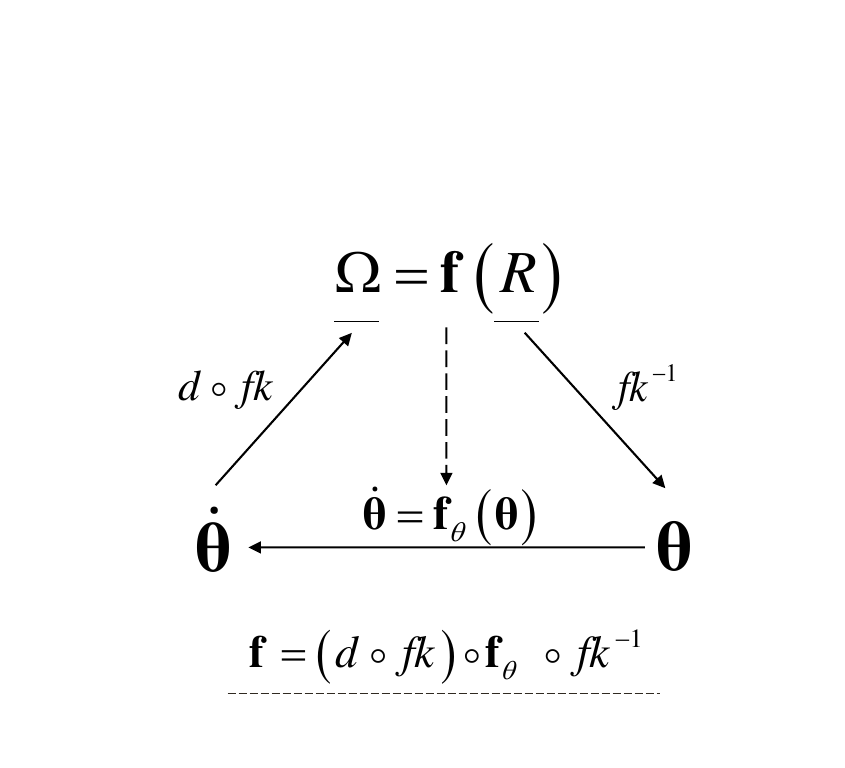}}
		\hspace{0.00cm}
		\subfigure[\label{fig:S2RefTra}]
		{\includegraphics[width=0.4\columnwidth]{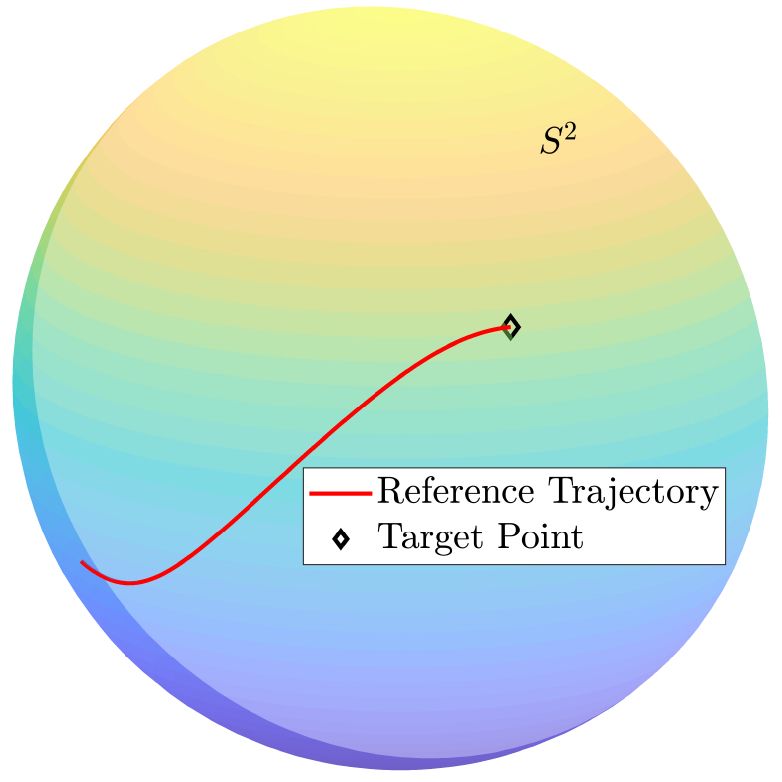}}%
		\vspace{0.00cm}		
		\subfigure[\label{fig:S2ThetaDS}]
		{\includegraphics[width=0.65\columnwidth]{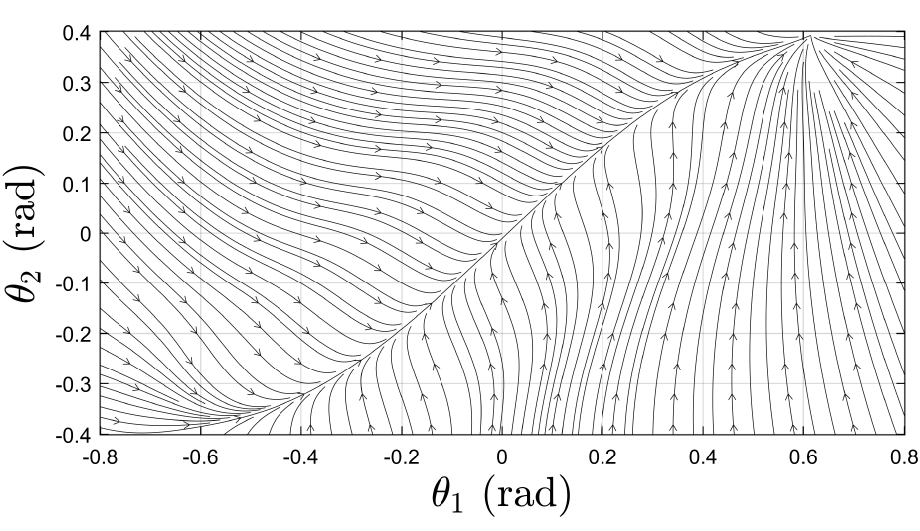}}%
		\hspace{0.00cm}
		\subfigure[\label{fig:S2ConDS}]
		{\includegraphics[width=0.4\columnwidth]{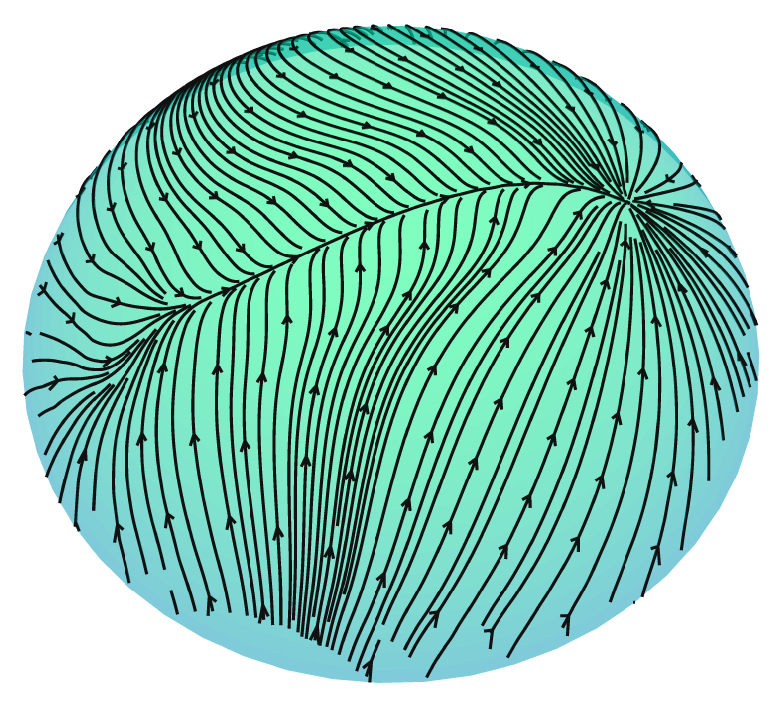}}
		\vspace{0.00cm}	
	\end{center}
	\caption{\label{fig:RotDsGen} {Schematic of the process of DS generation on $S^2$. (a) Schematic diagram of the mapping relationship between joint space DS and motion manifold DS. (b) Reference demonstration trajectory on $S^2$. (c) The conservative DS ${\boldsymbol{\dot \theta }} = {{\boldsymbol{f}}_\theta }\left( {\boldsymbol{\theta }} \right)$ in the linear joint space. (d) The conservative DS $\Omega  = {\mathbf{f}}\left( R \right)$ on $S^2$.}}
\end{figure*}

\begin{lemma}
	[Consistency of stiffness conservativeness~\citep{hou2024conserstiff}]
	\label{lem:consisconserv} 
	A symmetric and exact joint stiffness matrix ${{\mathbf{K}}_\theta }$ always corresponds to the conservative Cartesian stiffness matrix ${{\mathbf{K}}_C }$ on SE(3).
\end{lemma}
It follows from the Lemma~\ref{lem:consisconserv} that we can transform the problem to generate a conservative DS in the joint space and subsequently use the kinematic (statics) relations to generate a conservative DS on SE(3).

For demonstration, let's take two-degree-of-freedom rotation as an example. The posture $R$ of the robot at any moment can be considered to be on a two-dimensional sphere $S^2$. The reference trajectory of the robot can be represented as a curve on $S^2$, as shown in Figure~\ref{fig:S2RefTra}. Our goal is to generate a conservative DS $\Omega  = {\mathbf{f}}\left( R \right)$ on $S^2$, where $\Omega$ is the angular velocity. Denote the forward kinematics and differential kinematics of the robot as $fk\left(  \cdot  \right)$ and $d \circ fk\left(  \cdot  \right)$, respectively. Using the kinematic relationship, the learning from demonstration (LfD) on the original dataset $\left\{ {R_i^{{\text{ref}}},\Omega _i^{{\text{ref}}}} \right\}_{i = 1}^N$ can be transformed into the LfD on dataset $\left\{ {{\boldsymbol{\theta }}_i^{{\text{ref}}},{\boldsymbol{\dot \theta }}_i^{{\text{ref}}}} \right\}_{i = 1}^N$, where ${\boldsymbol{\theta }}$ and ${\boldsymbol{\dot \theta }}$ are the robot joint angles and angular velocities, respectively.
At this point, the goal translates into generating conservative DS ${\boldsymbol{\dot \theta }} = {{\boldsymbol{f}}_\theta }\left( {\boldsymbol{\theta }} \right)$ in the linear joint space, which can be achieved based on the method in Section~\ref{sec:Conservative DS}. Finally, the conservative DS ${\mathbf{f}}\left( R \right)$ on $S^2$ can be given based on ${{\boldsymbol{f}}_\theta }\left( {\boldsymbol{\theta }} \right)$:
\begin{equation}
	\label{eq:kinemap}
	{\mathbf{f}} = \left( {d \circ fk} \right) \circ {{\mathbf{f}}_\theta }{\text{ }} \circ f{k^{ - 1}}.
\end{equation}
The above algorithm flow is shown in Figure~\ref{fig:RotDsgen}. Figure~\ref{fig:S2ThetaDS} illustrates the conservative DS in the joint space, and the conservative DS on $S^2$ (see Figure~\ref{fig:S2ConDS}) can be further obtained based on the mapping relation (\ref{eq:kinemap}). As can be seen from the Figure~\ref{fig:S2ConDS}), the DS on $S^2$ is conservative and symmetrically attractive.

\subsection{\texorpdfstring{The Closed Trajectories ${\text{\&}}$ Self-Intersecting Trajectories}{}}\label{subsubsec:projectds}

\begin{figure*}[!ht]
	\begin{center}
		\subfigure[\label{fig:HigherDim}]
		{\includegraphics[width=2\columnwidth]{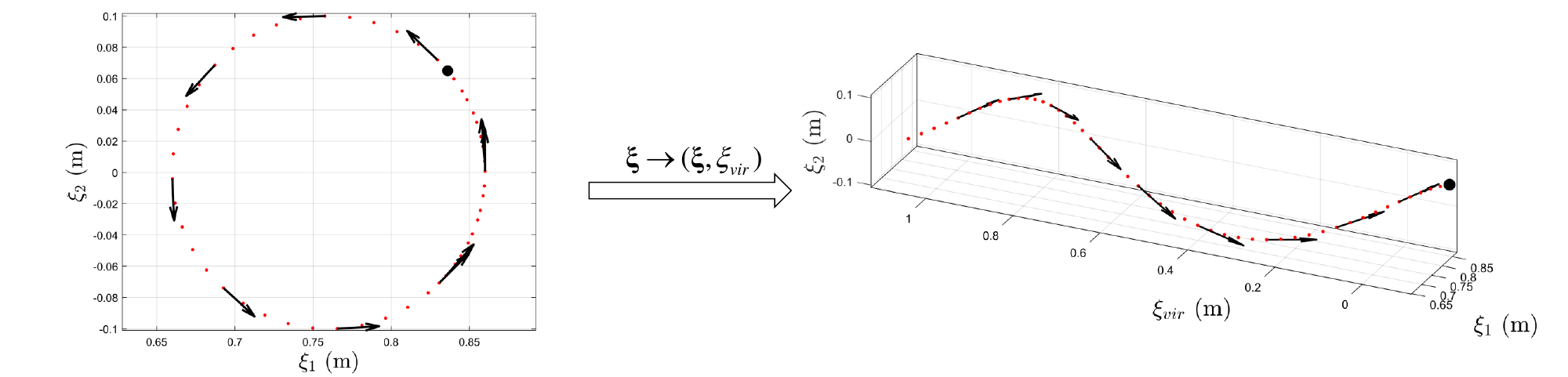}}%
		\hspace{3cm}
		\subfigure[\label{fig:LowerDim}]
		{\includegraphics[width=2\columnwidth]{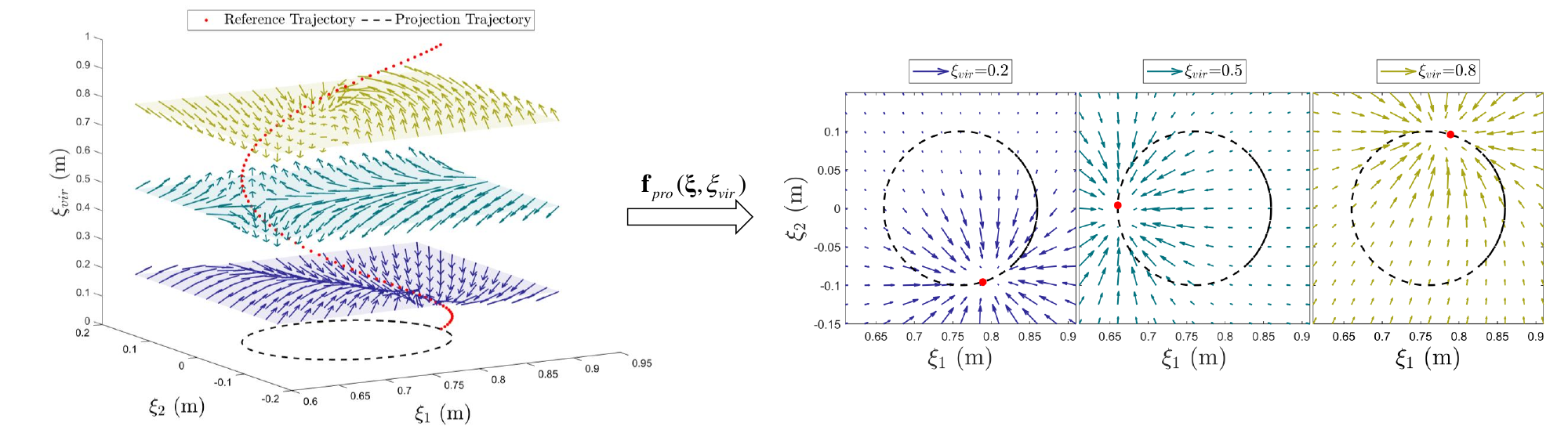}}
	\end{center}
	\caption{\label{fig:ClosedDsGen} {Schematic of the process of projected DS generation for the closed trajectories ${\text{\&}}$ self-intersecting trajectories. The red dots represent demonstration trajectories. The increase in spatial dimension in (a) is achieved by introducing the virtual coordinate ${\xi _{vir}}$. The three low-dimensional projected DSs corresponding to the high-dimensional DS at virtual coordinate ${\xi _{vir}} = \left[ {0.2,0.5,0.8} \right]$ are shown in (b). (a) Schematic of dimension increase. (b) Schematic of the projection for the high-dimensional dynamical system.}}
\end{figure*}

When the target trajectory is a closed or self-intersecting curve in ${\mathbb{R}^{n}}$, it is difficult to generate the DS by the method in Section~\ref{sec:Conservative DS} since the trajectory hardly corresponds to a single equilibrium point potential function defined on ${\mathbb{R}^{n}}$. However, we can transfer the problem to generate high-dimensional DS in the high-dimensional space ${\mathbb{R}^{n+1}}$.

\subsubsection{\texorpdfstring{Addition of Virtual Coordinates ${\xi _{vir}}$}{}}
$\\$
For closed or self-intersecting trajectories ${\boldsymbol{\xi }}$ in the original ${\mathbb{R}^{n}}$ space, it is difficult to find a single equilibrium point potential function $V_p$ corresponding to them. However, we can describe this trajectory in higher dimensional space ${\mathbb{R}^{n+1}}$ by adding virtual coordinates ${\xi _{vir}}$: ${\boldsymbol{\xi }} \to ({\boldsymbol{\xi }},{\xi _{vir}})$.
When introducing the virtual coordinate ${\xi _{vir}}$, we can design it as a unidirectional evolutionary coordinate on the high-dimensional space.
In Figure~\ref{fig:HigherDim}, the original closed trajectory in the two-dimensional plane is transformed into a unidirectional advancing helix in three-dimensional space.
Then this spiral trajectory is easy to find the corresponding potential function $V_p$ using the algorithm in Section~\ref{sec:Conservative DS}.

\subsubsection{\texorpdfstring{Generation of the DS in ${\mathbb{R}^{n+1}}$}{}}
$\\$
For the increased dimensional trajectory $({\boldsymbol{\xi }},{\xi _{vir}})$, we are able to generate the corresponding potential function $V\left( {{\boldsymbol{\xi }},{\xi _{vir}}} \right)$ using the algorithm in Section~\ref{sec:Conservative DS}.
Similarly, the high-dimensional DS ${\mathbf{ f}}({\boldsymbol{\xi }},{\xi _{vir}})$ can be calculated based on the potential function $V\left( {{\boldsymbol{\xi }},{\xi _{vir}}} \right)$:
\begin{equation}
	\label{eq:HighDimDS}
	{\mathbf{ f}}({\boldsymbol{\xi }},{\xi _{vir}}) =  - {\nabla _{({\boldsymbol{\xi }},{\xi _{vir}})}}V\left( {{\boldsymbol{\xi }},{\xi _{vir}}} \right).
\end{equation}

\subsubsection{The projection of High-Dimensional DS}
\label{subsubsubsec:projectds}
$\\$
Since the actual physical coordinates of the target trajectory ${\boldsymbol{\xi }}$ are in ${\mathbb{R}^{n}}$, after generating the high-dimensional DS ${\mathbf{ f}}({\boldsymbol{\xi }},{\xi _{vir}})$, we need to project the high-dimensional DS onto the low-dimensional space ${\mathbb{R}^{n}}$.
We first introduce the projection operator ${\pi _{(1, \ldots ,n)}}:{\mathbb{R}^m} \to {\mathbb{R}^n}$ as follows:
\begin{equation}
	{\pi _{(1, \ldots ,n)}}\left[ {\left( {{\xi_1}, \ldots ,{\xi_n}, \ldots ,{\xi_m}} \right)} \right] = \left( {{\xi_1}, \ldots ,{\xi_n}} \right).
\end{equation}
Therefore, the projected DS ${{\mathbf{f}}_{pro}}({\boldsymbol{\xi }},{\xi _{vir}})$ can be expressed as
\begin{equation}
	{{\mathbf{f}}_{pro}}({\boldsymbol{\xi }},{\xi _{vir}}) = {\pi _{(1, \ldots ,n)}}\left[ {{\mathbf{f}}({\boldsymbol{\xi }},{\xi _{vir}})} \right].
\end{equation}

In Figure~\ref{fig:LowerDim}, the physical space reference trajectory is shown as a black dashed line. By introducing the virtual coordinate, the reference trajectory is transformed into a red spiral line. The role of the projection operator ${\pi _{(1, \ldots ,n)}}$ is equivalent to taking the slice of the DS in 3D space corresponding to a specific virtual coordinate value.

We show three projection results for ${\xi _{vir}} = \left[ {0.2,0.5,0.8} \right]$ in Figure~\ref{fig:LowerDim}. It can be seen that the two-dimensional projection DS ${{\mathbf{f}}_{pro}}({\boldsymbol{\xi }},{\xi _{vir}})$ exhibits a symmetric attraction about the current target point (red point), which is a result induced by the symmetric attractiveness of the DS ${\mathbf{ f}}({\boldsymbol{\xi }},{\xi _{vir}})$ in the higher dimensional space. As the virtual coordinate ${\xi _{vir}}$ is evolving, the equivalent target red point is constantly changing, driving the robot to move along the closed circle trajectory in the low-dimensional space.

\subsubsection{Passivity-Based Control for Projection DS}
$\\$
The DS obtained in the above Section~\ref{subsubsubsec:projectds} is finally used for the control of the robot. The impedance control based on the projected DS ${\mathbf{ f}}({\boldsymbol{\xi }},{\xi _{vir}})$ is designed as follows:
\begin{equation}
	\label{eq:PassiveDSContr1}
	{{\boldsymbol{\tau }}_c} = {\mathbf{G}}({\boldsymbol{\xi }}) - {\mathbf{D}}({\boldsymbol{\xi }},{\xi _{vir}}){\boldsymbol{\dot \xi }} + {\lambda _1}{{\mathbf{f}}_{pro}}({\boldsymbol{\xi }},{\xi _{vir}}),
\end{equation}
where ${\mathbf{D}}({\boldsymbol{\xi }},{\xi _{vir}})= {\mathbf{Q}}({\boldsymbol{\xi }},{\xi _{vir}}){\boldsymbol{\Lambda Q}}{({\boldsymbol{\xi }},{\xi _{vir}})^T}$ is the damping matrix. ${\mathbf{Q}}({\boldsymbol{\xi }},{\xi _{vir}}) = \left[ {{{\mathbf{e}}_1}, \ldots ,{{\mathbf{e}}_n}} \right]$ is the orthogonal matrix where ${{\mathbf{e}}_1} = {{\mathbf{f}}_{pro}}({\boldsymbol{\xi }},{\xi _{vir}})/\left\| {{\mathbf{f}}_{pro}}({\boldsymbol{\xi }},{\xi _{vir}}) \right\|$.

The energy storage function is chosen as follows:
\begin{equation}
	\label{eq:StorageFun}
	W = \frac{1}{2}{{\boldsymbol{\dot \xi }}^T}{\mathbf{M}}({\boldsymbol{\xi }}){\boldsymbol{\dot \xi }} + {\lambda _1}V({\boldsymbol{\xi }},{\xi _{vir}}),
\end{equation}
then we have the following propsition.
\begin{proposition} 
	\label{pro:PassiveDSContr1}
	The system (\ref{eq:RigidDyn}) controlled by (\ref{eq:PassiveDSContr1}) is passive with regard to the input-output pair $( {{\boldsymbol{\dot \xi }},{{\boldsymbol{\tau }}_e}} )$ with the storage function (\ref{eq:StorageFun}). See Appendix~\ref{apd:PassiveDSContr1} for details.
\end{proposition}

\subsection{The Decomposition of Dynamical System}\label{subsubsec:DecompDS}
In the above sections, we introduced learning conservative DS from various demonstration trajectories. However, existing advanced methods often generate a broader class of non-conservative DS. For some cases, DS described using (polynomial) analytic expressions meet the task requirements well but tend to be non-conservative. The non-conservativeness of DS can cause the energy tanks to decay fast. Therefore, we propose an effective vector field decomposition strategy for non-conservative DS.	

For any ${\mathbf{f}}({\boldsymbol{\xi }})$, there are infinitely many ways to decompose it into a conservative part ${{\mathbf{f}}_c}({\boldsymbol{\xi }})$ and a non-conservative part ${{\mathbf{f}}_{nc}}({\boldsymbol{\xi }})$:
\begin{equation}
	{\mathbf{f}}({\boldsymbol{\xi }}) = {{\mathbf{f}}_c}({\boldsymbol{\xi }}) + {{\mathbf{f}}_{nc}}({\boldsymbol{\xi }}).
\end{equation}
However, a superior decomposition exists for a particular design of controller.

We use the computationally convenient polynomial function ${{\mathbf{\hat f}}_{\boldsymbol{\nu }}}({\boldsymbol{\xi }}) = {{\mathbf{f}}_c}({\boldsymbol{\xi }})$ (${\boldsymbol{\nu }}$ is the hyperparameters) as the conservative part. For the linear 3D spatial case ${\boldsymbol{\xi }} = {\left[ {\begin{array}{*{20}{c}}
			{{\xi _1}}&{{\xi _2}}&{{\xi _3}} 
	\end{array}} \right]^T} \in {\mathbb{R}^3}$, the simplest class of ${{\mathbf{\hat f}}_{\boldsymbol{\nu }}}({\boldsymbol{\xi }})$ is given as
\begin{equation}
	{{\mathbf{\hat f}}_{\boldsymbol{\nu }}}\left( {\boldsymbol{\xi }} \right) = {{\mathbf{f}}_{\boldsymbol{\nu }}}\left( {\boldsymbol{\xi }} \right) - {{\mathbf{f}}_{\boldsymbol{\nu }}}\left( {{{\boldsymbol{\xi }}_0}} \right),
\end{equation}
where 
\begin{equation}
	{{\mathbf{f}}_{\boldsymbol{\nu }}}\left( {\boldsymbol{\xi }} \right) = \left[ {\begin{array}{*{20}{c}}
			{\begin{array}{*{20}{c}}
					{{\nu _1}{{\left( {{\xi _1}} \right)}^4}/4 + {v_2}{{\left( {{\xi _1}} \right)}^3}/3 + {\nu _3}{{\left( {{\xi _1}} \right)}^2}/2 + {\nu _4}{\xi _1}} \\ 
					{{\nu _5}{{\left( {{\xi _2}} \right)}^4}/4 + {\nu _6}{{\left( {{\xi _2}} \right)}^3}/3 + {\nu _7}{{\left( {{\xi _2}} \right)}^2}/2 + {\nu _8}{\xi _2}} 
			\end{array}} \\ 
			{{\nu _9}{{\left( {{\xi _3}} \right)}^4}/4 + {\nu _{10}}{{\left( {{\xi _3}} \right)}^3}/3 + {\nu _{11}}{{\left( {{\xi _3}} \right)}^2}/2 + {\nu _{12}}{\xi _3}} 
	\end{array}} \right].
\end{equation}
See Appendix~\ref{apd:FittingFunction} for specific details.

Sampling $N_d$ points $\left\{ {{{\boldsymbol{\xi }}_i}} \right\}_{i = 1}^{{N_d}}$ uniformly in the critical region $\mathcal{C}$, one of the most direct indexes can be expressed as
\begin{equation}
	\label{eq:decompindex1}
	\mathop {\min }\limits_{\boldsymbol{\nu }} {J_1}({\boldsymbol{\nu }}) = \frac{1}{{{N_d}}}\sum\limits_{i = 1}^{{N_d}} {{{\left\| {{{{\mathbf{\hat f}}}_{\boldsymbol{\nu }}}\left( {{{\boldsymbol{\xi }}_i}} \right) - {\mathbf{f}}\left( {{{\boldsymbol{\xi }}_i}} \right)} \right\|}^2}}.
\end{equation}
(\ref{eq:decompindex1}) shows that we utilize ${{{\mathbf{\hat f}}}_{\boldsymbol{\nu }}}\left( {{{\boldsymbol{\xi }}}} \right)$ to fit the conservative component of the original DS ${\mathbf{f}}\left( {{{\boldsymbol{\xi }}}} \right)$. However, this decomposition does not fully utilize the structure of the controller and tends to be ineffective in some cases.

We analyze the passivity of the controller (\ref{eq:PassiveDSContr}) in terms of the following energy storage function
\begin{equation}
	W = \frac{1}{2}{{\boldsymbol{\dot \xi }}^T}{\mathbf{M}}({\boldsymbol{\xi }}){\boldsymbol{\dot \xi }}+{\lambda _1}{V_{{{\mathbf{f}}_{c}}}}({\boldsymbol{\xi }}),
\end{equation}
where ${V_{{{\mathbf{f}}_{nc}}}}({\boldsymbol{\xi }})$ is the potential function corresponding to ${{\mathbf{f}}_{c}}({\boldsymbol{\xi }})$: ${{\mathbf{f}}_{c}}({\boldsymbol{\xi }}) =  - \nabla {V_{{{\mathbf{f}}_{c}}}}({\boldsymbol{\xi }})$.

Combined with the rigid body dynamics equation (\ref{eq:RigidDyn}), we can obtain the rate of change of the energy storage function:
\begin{equation}
	\label{eq:storagerat}
	\begin{aligned}
		{\dot W}
		&={{{\boldsymbol{\dot \xi }}}^T}{\mathbf{M}}({\boldsymbol{\xi }}){\boldsymbol{\ddot \xi }} + \frac{1}{2}{{{\boldsymbol{\dot \xi }}}^T}{\mathbf{\dot M}}({\boldsymbol{\xi }}){\boldsymbol{\dot \xi }}+ {\lambda _1}\nabla {V_{{{\mathbf{f}}_{c}}}}{({\boldsymbol{\xi }})^T}{\boldsymbol{\dot \xi }} \\ 
		&={{{\boldsymbol{\dot \xi }}}^T}\left( {{{\boldsymbol{\tau }}_c} + {{\boldsymbol{\tau }}_e} - {\mathbf{C}}({\boldsymbol{\xi }},{\boldsymbol{\dot \xi }}){\boldsymbol{\dot \xi }} - {\mathbf{g}}({\boldsymbol{\xi }})} \right) \\
		&\quad + \frac{1}{2}{{{\boldsymbol{\dot \xi }}}^T}{\mathbf{\dot M}}({\boldsymbol{\xi }}){\boldsymbol{\dot \xi }}+ {\lambda _1}\nabla {V_{{{\mathbf{f}}_{c}}}}{({\boldsymbol{\xi }})^T}{\boldsymbol{\dot \xi }} \\
		&=\frac{1}{2}{{{\boldsymbol{\dot \xi }}}^T}({\mathbf{\dot M}} - 2{\mathbf{C}}){\boldsymbol{\dot \xi }} - {{{\boldsymbol{\dot \xi }}}^T}{\mathbf{D }\boldsymbol{\dot \xi}} + {{{\boldsymbol{\dot \xi }}}^T}{{\boldsymbol{\tau }}_e} \\
		&\quad + ({\lambda _1}{{{\boldsymbol{\dot \xi }}}^T}{\mathbf{f}}({\boldsymbol{\xi }})+ {\lambda _1}\nabla {V_{{{\mathbf{f}}_{c}}}}{({\boldsymbol{\xi }})^T}{\boldsymbol{\dot \xi }}) \\
		&=\left( { - {{{\boldsymbol{\dot \xi }}}^T}{\mathbf{D }\boldsymbol{\dot \xi}} + {\lambda _1}{{{\boldsymbol{\dot \xi }}}^T}{\mathbf{f}_{nc}}({\boldsymbol{\xi }})} \right) + {{{\boldsymbol{\dot \xi }}}^T}{{\boldsymbol{\tau }}_e}.
	\end{aligned}
\end{equation}
Notice the term $\left( { - {{{\boldsymbol{\dot \xi }}}^T}{\mathbf{D }\boldsymbol{\dot \xi}} + {\lambda _1}{{{\boldsymbol{\dot \xi }}}^T}{\mathbf{f}_{nc}}({\boldsymbol{\xi }})} \right)$ in (\ref{eq:storagerat}), so the energy tank is set as (\ref{eq:EnergyTankB}) and (\ref{eq:EnergyTank1B}). 

From the energy rate equation (\ref{eq:EnergyTank1B}), we can slow down the dissipation of the energy $s$ by minimizing ${{\boldsymbol{\dot \xi }}^T}{{\mathbf{f}}_{nc}}({\boldsymbol{\xi }})$. 
In general, it is assumed that the actual velocity of the robot ${\boldsymbol{\dot \xi }}$ tracks well the reference velocity ${\mathbf{f}}({\boldsymbol{\xi }})$ given by the controller.
Thus We have ${{\boldsymbol{\dot \xi }}^T}{{\mathbf{f}}_{nc}}({\boldsymbol{\xi }}) \approx {\mathbf{f}}{\left( {\boldsymbol{\xi }} \right)^T}\left( {{\mathbf{f}}\left( {\boldsymbol{\xi }} \right) - {{{\mathbf{\hat f}}}_{\boldsymbol{\nu }}}\left( {\boldsymbol{\xi }} \right)} \right)$. The composite index after considering the controller structure is finally expressed as (corespinding to the specific energy tank design (\ref{eq:EnergyTank1B}))
\begin{equation}
	\label{eq:decompindex}
	\begin{aligned}
		\mathop {\min }\limits_{\boldsymbol{\nu }} {J_2}({\boldsymbol{\nu }}) 
		&= {\omega _1}\frac{1}{{{N_d}}}\sum\limits_{i = 1}^{{N_d}} {{{\left\| {{{{\mathbf{\hat f}}}_{\boldsymbol{\nu }}}\left( {{{\boldsymbol{\xi }}_i}} \right) - {\mathbf{f}}\left( {{{\boldsymbol{\xi }}_i}} \right)} \right\|}^2}} \\
		&\quad + {\omega _2}\frac{1}{{{N_d}}}\sum\limits_{i = 1}^{{N_d}} {{\mathbf{f}}{{\left( {{{\boldsymbol{\xi }}_i}} \right)}^T}\left( {{\mathbf{f}}\left( {{{\boldsymbol{\xi }}_i}} \right) - {{{\mathbf{\hat f}}}_{\boldsymbol{\nu }}}\left( {{{\boldsymbol{\xi }}_i}} \right)} \right)},
	\end{aligned}
\end{equation}
where ${\boldsymbol{\omega }} = \left[ {{\omega _1},{\omega _2}} \right]$ is the weight vector. The validity of this index will be discussed and analyzed in Section~\ref{subsubsec:Exp3}. The design of the energy tank-based controller is not unique, and for other structures, the corresponding decomposition indexes can be designed according to the above process as well.

%% file: sections/6_simu_exp.tex
\begin{figure*}[!h]
	\begin{center}
		\subfigure[\label{fig:Exp1set}]
		{\includegraphics[width=0.95\columnwidth]{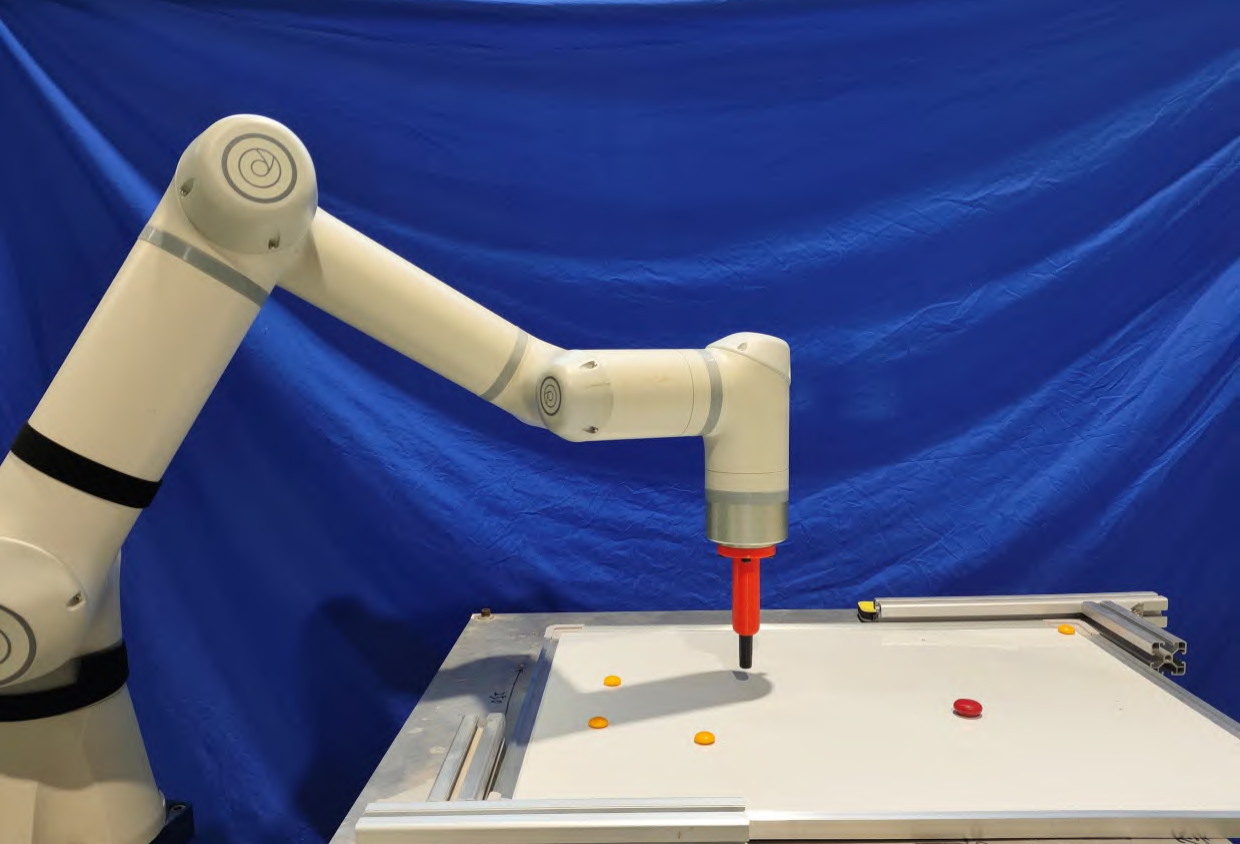}}%
		\hspace{0.01cm}
		\subfigure[\label{fig:Exp1WB}]
		{\includegraphics[width=0.85\columnwidth]{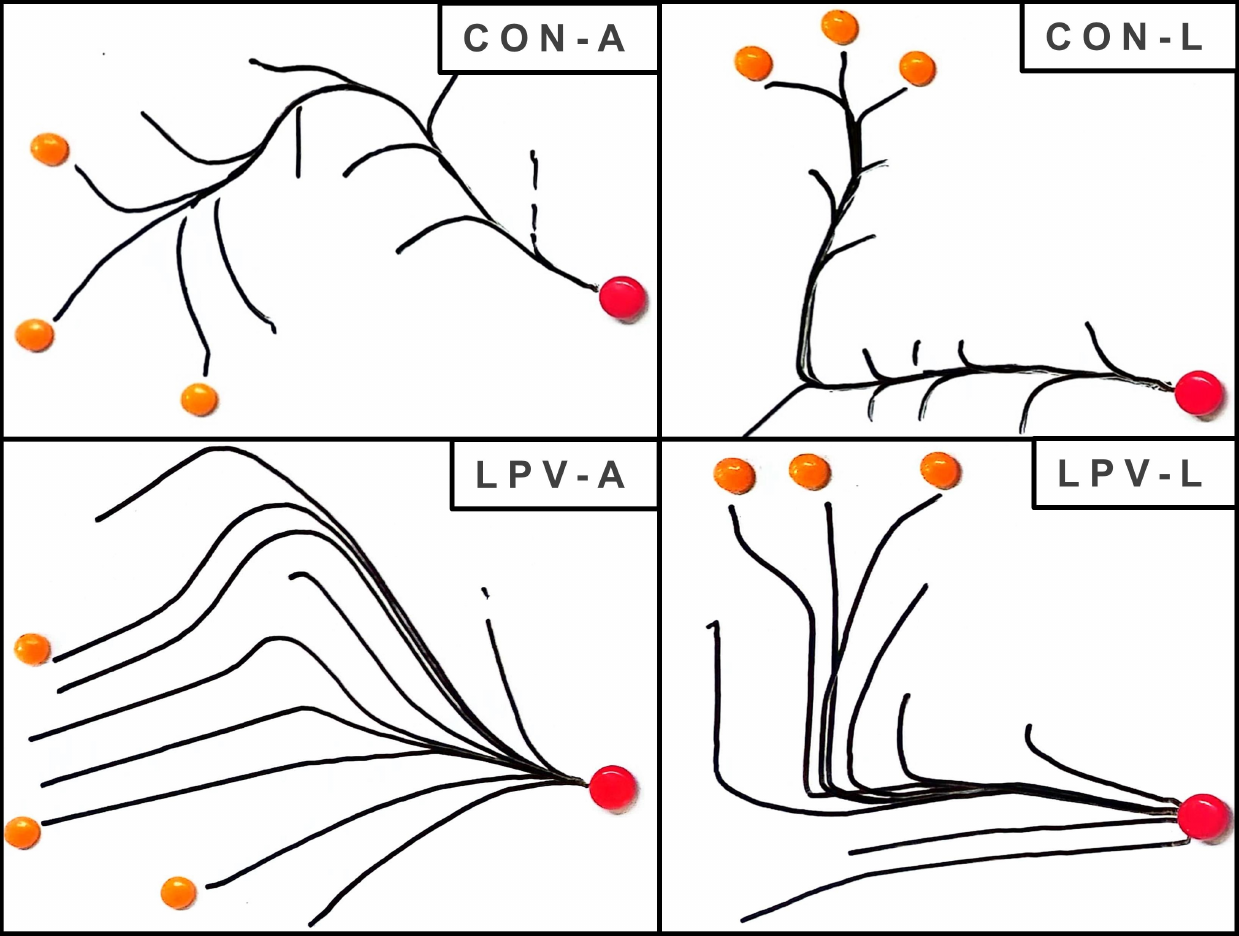}}
		\vspace{0.00cm}		
	\end{center}
	\caption{\label{fig:Exp1Set} {Schematic diagram of the experiment in Section~\ref{subsubsec:Exp1}. For clarity, the orange circle on the whiteboard indicates the start of the movement and the red circle indicates the end. (b) shows the comparative results of two different methods on two trajectories. (a) Experimental setup. (b) Experimental comparison results on the whiteboard.}}
\end{figure*}

\section{Simulation and Experimental Verification}\label{sec:Simu_Exp}
A series of simulations and experiments are conducted in this section to validate the methods proposed in Section~\ref{sec:Conservative DS} and Section~\ref{sec:generalization}. In Section~\ref{subsubsec:Exp1}, we test the proposed method on the classical LASA handwriting dataset~\citep{Khansari2011SEDS} to highlight the symmetric attractiveness and conservativeness of the proposed method. Next, environmental obstacles are imposed during the execution of robotic arm movements to verify the interaction safety in the framework of DS in Section~\ref{subsubsec:Exp2}. Subsequently, the effectiveness of the DS decomposition is verified in Section~\ref{subsubsec:Exp3} on more complex and twisted trajectories. Finally, for more specific closed ${\text{\&}}$ self-intersecting trajectories, the effectiveness of the DS decomposition is limited. Section~\ref{subsubsec:Exp4} demonstrates that the proposed method (projection conservative DS) still handles such cases well. The above experiments illustrate that the proposed method effectively improves the stability margin during control while having the potential to handle a wide range of scenarios.

\subsection{LASA Handwriting Motions}\label{subsubsec:Exp1}
The symmetric attraction and conservativeness of the DS generated in Section~\ref{sec:Conservative DS} are important in the actual motion execution. In this section, we illustrate this by performing a series of simulations and experiments on the LASA handwriting dataset.

Figure~\ref{fig:Exp1Set} illustrates the experimental setup.
The experiment was conducted on a 6-DOF collaborative robot. The robot is equipped with encoders and torque sensors to acquire joint position signals and joint torque, respectively. The communication between the upper computer and the robot is realized through TwinCAT 3 interface at 1 kHz.
To clearly show the robot's trajectory, we mounted a pen on the end of the robot arm and used a whiteboard to record the trajectory during the movement. Also, the orange circles were used to indicate the starting point of the motion, and a red circle was used to indicate the equilibrium point ${{\boldsymbol{\xi }}_0}$ corresponding to the DS.

\begin{figure*}[!h]
	\begin{center}
		\subfigure[\label{fig:LyapA}]
		{\includegraphics[width=0.85\columnwidth]{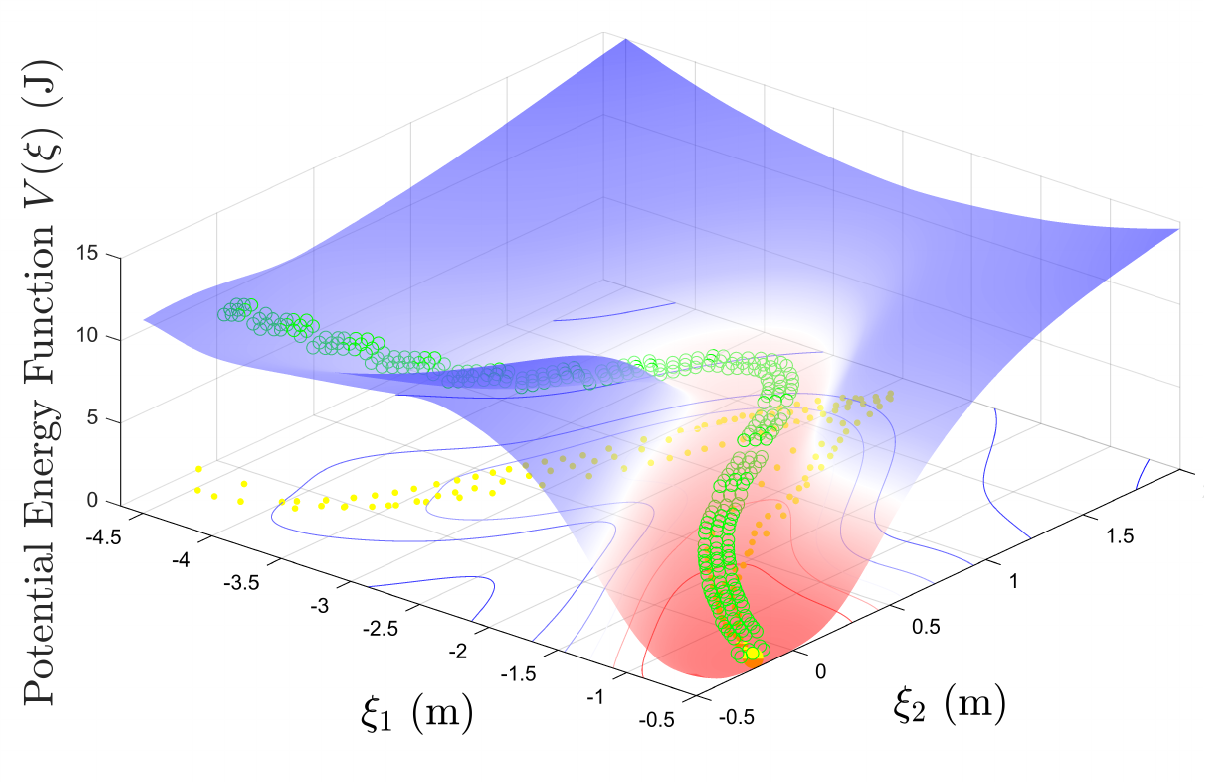}}%
		\hspace{0.01cm}
		\subfigure[\label{fig:CondsA}]
		{\includegraphics[width=0.85\columnwidth]{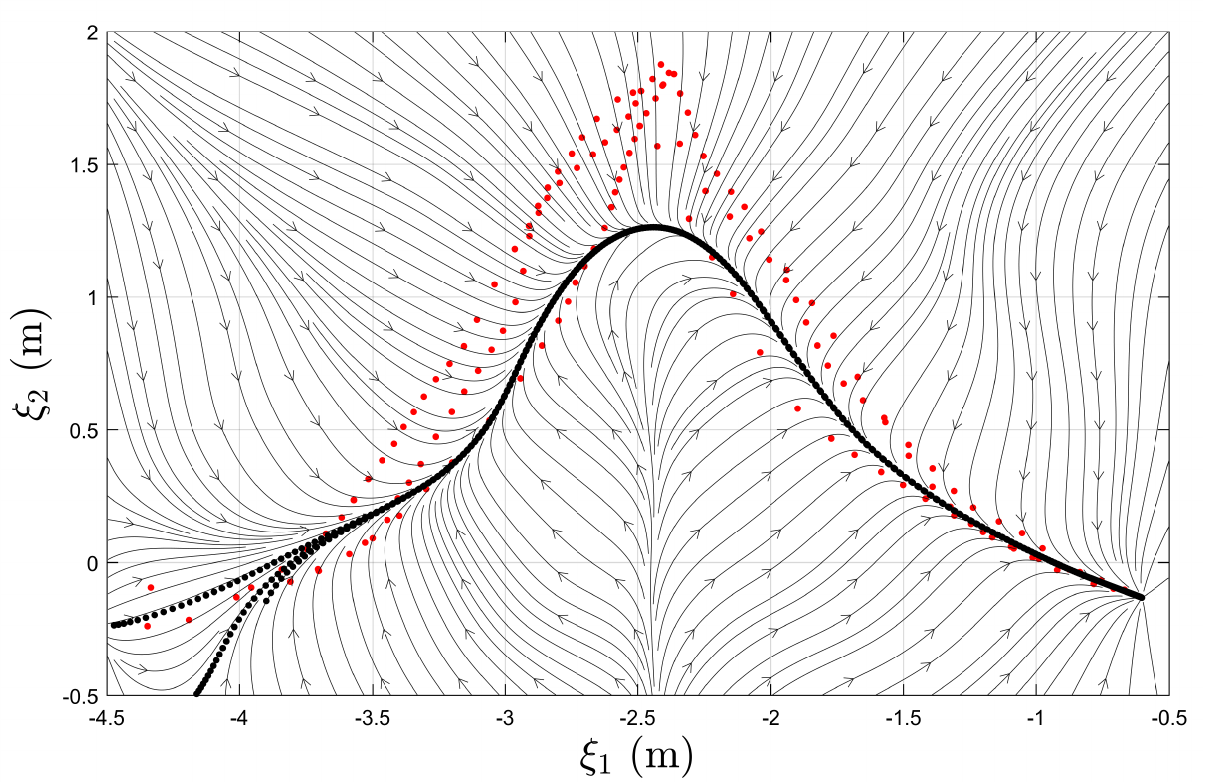}}
		\vspace{0.00cm}		
		\subfigure[\label{fig:VelConDSA}]
		{\includegraphics[width=0.85\columnwidth]{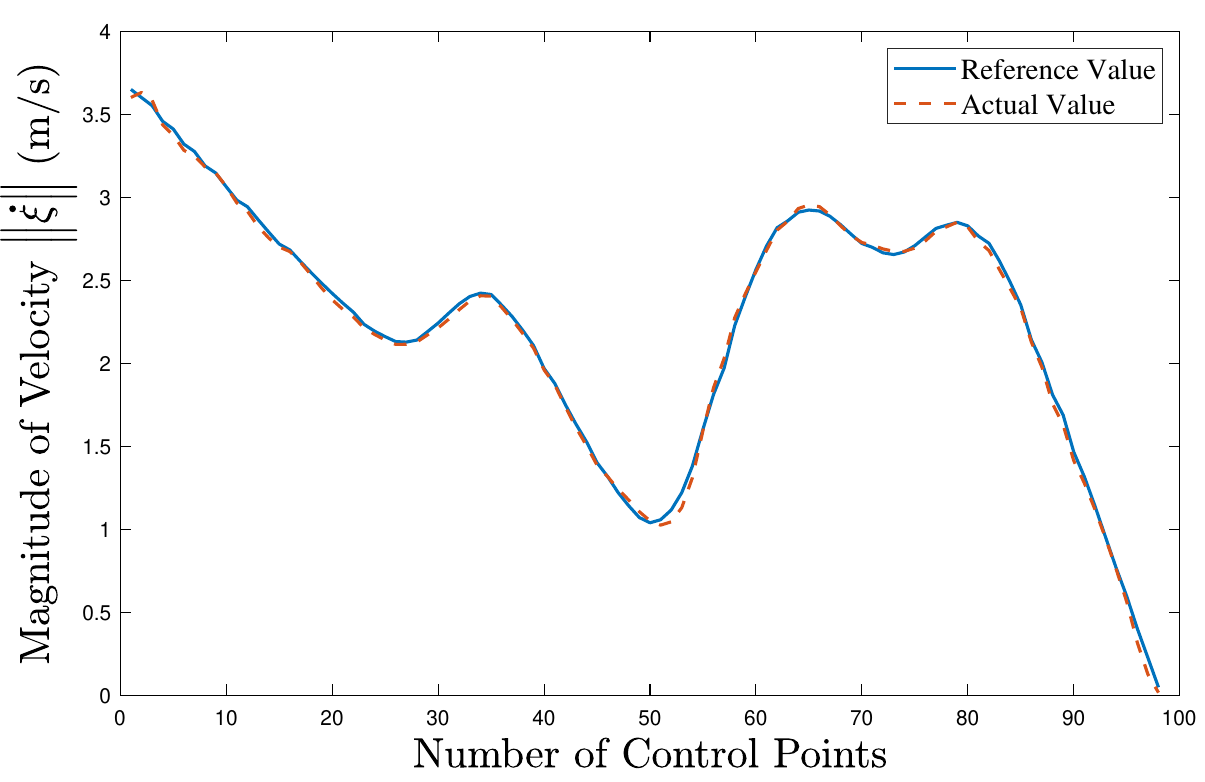}}%
		\hspace{0.01cm}
		\subfigure[\label{fig:DSLpvA}]
		{\includegraphics[width=0.85\columnwidth]{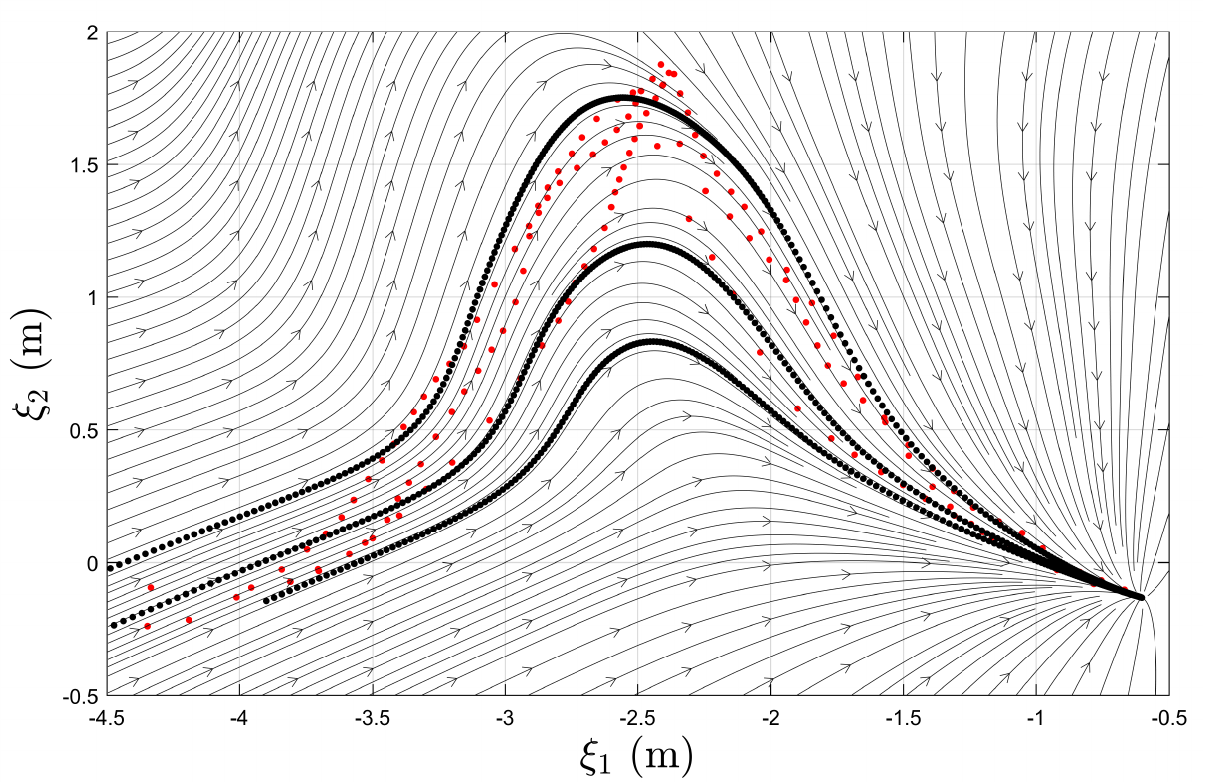}}
		\vspace{0.00cm}	
	\end{center}
	\caption{\label{fig:ConDSA} {Simulation of DS generation on the A-shaped trajectory. The process and details of the DS generation based on GP are displayed in (a), (b) and (c). To exemplify the property of symmetric attraction, (d) shows the LPV-DS as a comparison. The red dots and black lines in (b) and (d) represent the demonstration trajectories and integral curves of DS, respectively. (a) The potential function $V_p$ based on GP. (b) DS based on GP. (c) The velocity comparison. The actual velocity represents the result of learning through the GP. (d) LPV-DS for A-shaped trajectory.}}
\end{figure*}

Using the method in Section~\ref{sec:Conservative DS}, we generated the DS corresponding to the A-shaped trajectory, as shown in Figure~\ref{fig:LyapA}, \ref{fig:CondsA} and \ref{fig:VelConDSA}.
Based on GP, we generated the potential energy function $V_p$ corresponding to the A-shaped demonstration trajectory in Figure~\ref{fig:LyapA}.
By calculating the negative gradient for $V_p$, we then obtain the corresponding DS in Figure~\ref{fig:CondsA}.
With the optimization in Section~\ref{subsubsec:Step3}, the actual velocities of the vector field generated in Figure~\ref{fig:CondsA} are highly consistent with the reference velocities and a good fit is achieved, as shown in Figure~\ref{fig:VelConDSA}.
To illustrate the role of symmetric attraction on the improvement of trajectory accuracy and perturbation resistance, we simultaneously show the DS generated based on the LPV-DS method for the same A-shaped trajectory in Figure~\ref{fig:DSLpvA}. As can be seen in Figure~\ref{fig:CondsA}, due to the symmetric attractiveness of the proposed method, after the three initial starting points are selected, the trajectories are first attracted to the demonstration trajectory quickly and then move along the demonstration trajectory to the equilibrium point.

Figure~\ref{fig:Exp1WB} shows the results of the whiteboard experiments in which the robot uses the proposed method in this paper and the LPV-DS method for A-shaped and L-shaped trajectories, respectively.
It can be seen that, not only for the perturbation of the start point, but also after perturbations on various intervals on the path, the proposed method (CON-A, CON-L) enables the robot to quickly return to the demonstration trajectory, and subsequently continue to execute the motion. Through the above discussion, we demonstrate the symmetric attractiveness of the approach proposed in this paper with its importance.

In Section~\ref{sec:background}, we introduce that a conservative DS corresponds to the passivity of the control system. When the DS is non-conservative, the system is kept passive by introducing the energy tank. However, when the energy $s$ in the energy tank is depleted, the performance of the controller is affected in the following two cases: (1) the energy is depleted and the switching function $\beta({\boldsymbol{\xi }, s})$ is set to zero, at which time the system has difficulty moving along the target DS ${\mathbf{f}}({\boldsymbol{\xi }})$ but moves along the conservative part ${{\mathbf{f}}_c}({\boldsymbol{\xi }})$; and (2) the energy tank is switched between fast charging and discharging, and the switching function $\beta({\boldsymbol{\xi }, s})$ varies rapidly between zero and non-zero, which degrades the system dynamic performance.

\begin{figure*}[!h]
	\begin{center}
		\subfigure[\label{fig:EnergyTankA}]
		{\includegraphics[width=0.85\columnwidth]{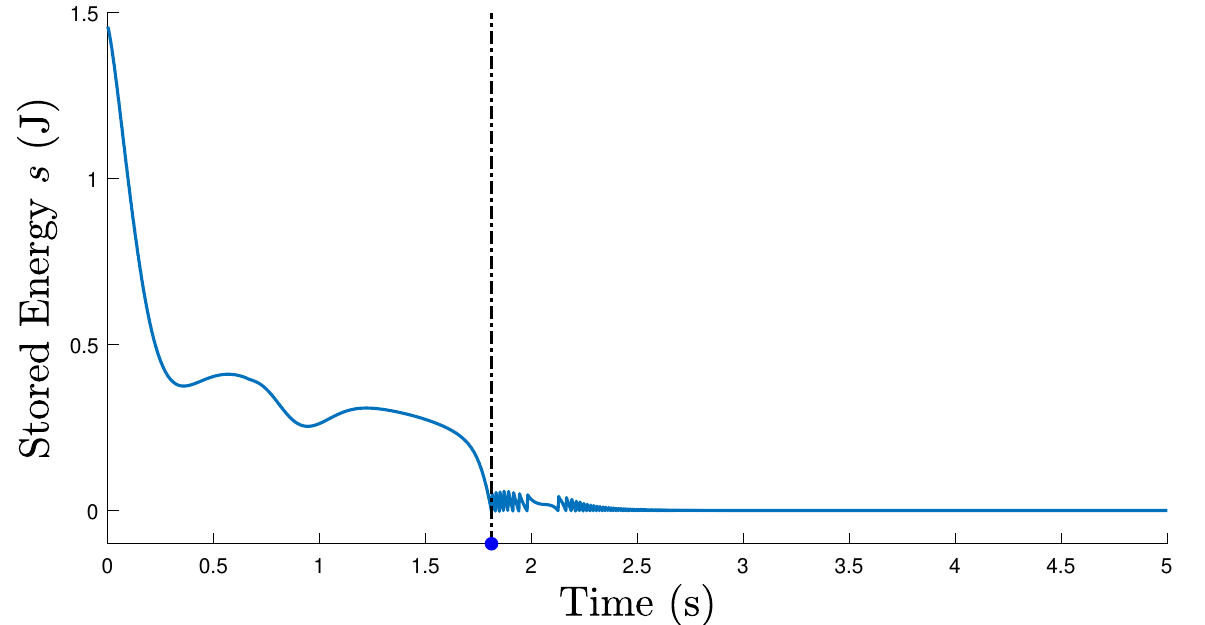}}%
		\hspace{0.01cm}
		\subfigure[\label{fig:PosA}]
		{\includegraphics[width=0.85\columnwidth]{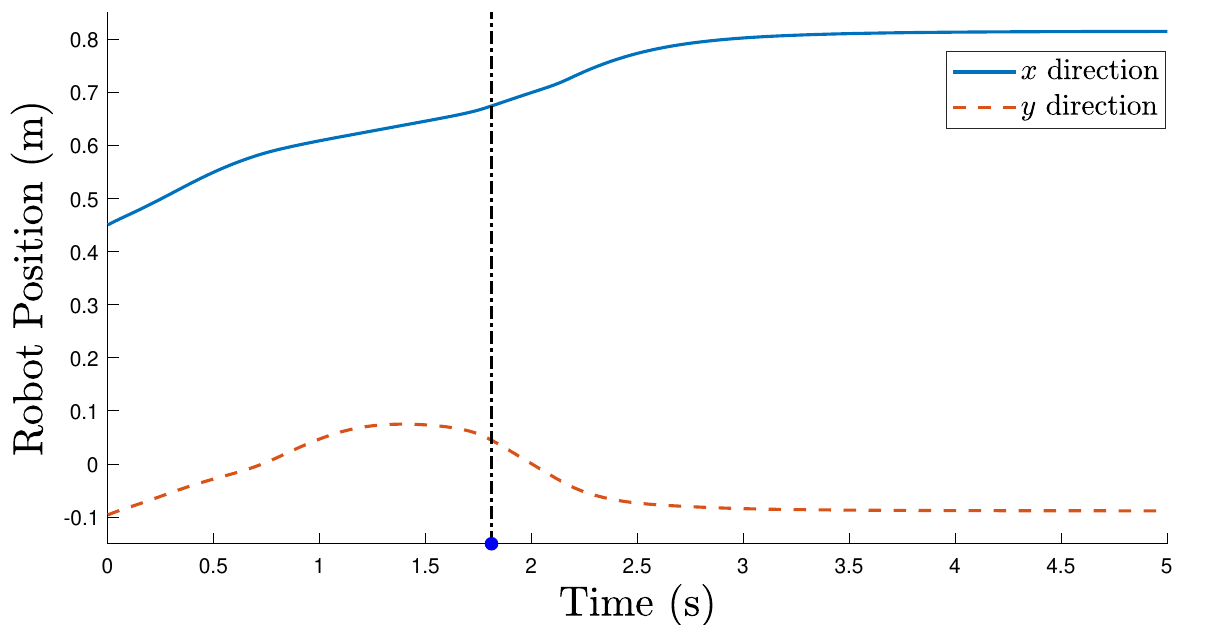}}
		\vspace{0.00cm}		
		\subfigure[\label{fig:VelA}]
		{\includegraphics[width=0.85\columnwidth]{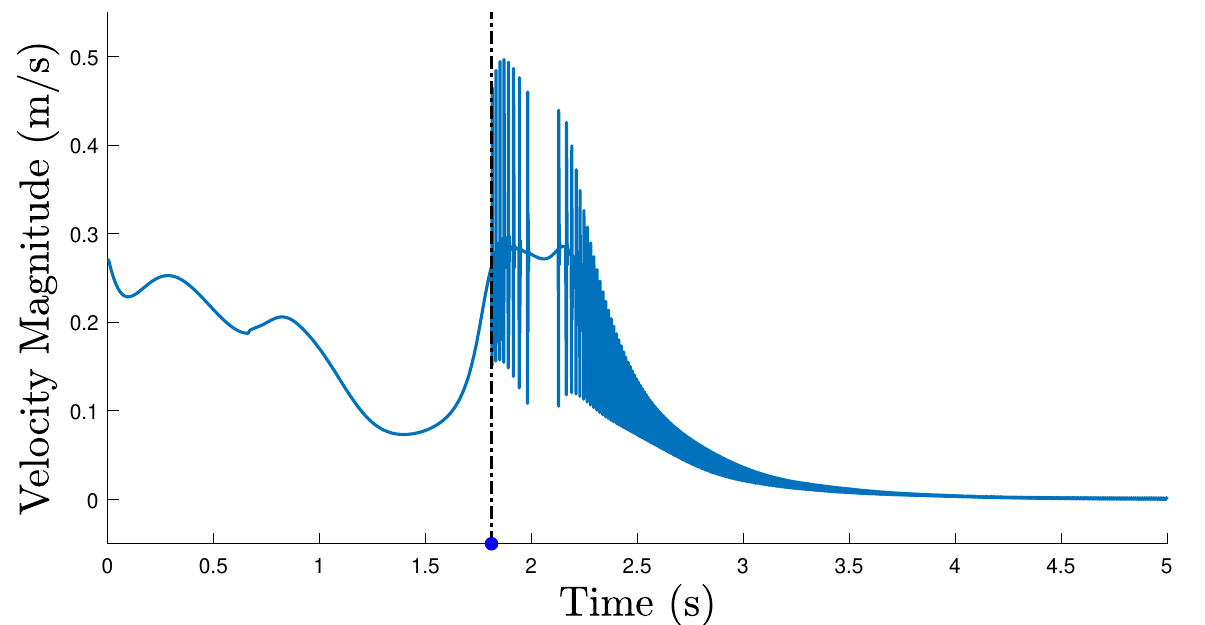}}%
		\hspace{0.01cm}
		\subfigure[\label{fig:TorqueA}]
		{\includegraphics[width=0.85\columnwidth]{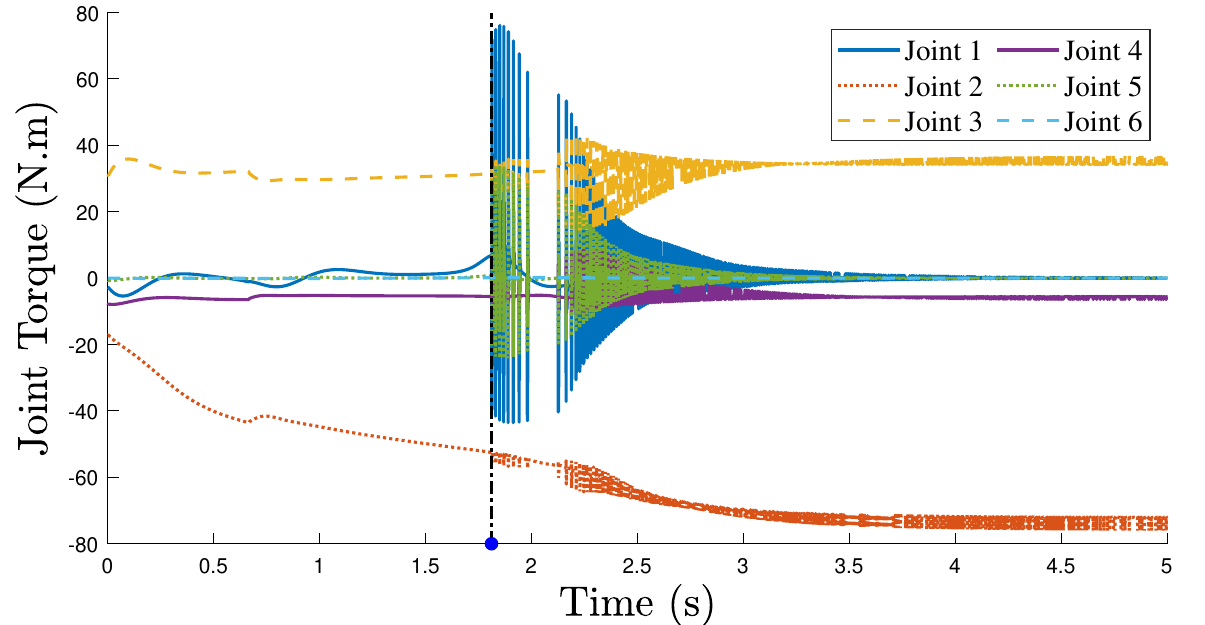}}
		\vspace{0.00cm}	
	\end{center}
	\caption{\label{fig:EnergyA} {Simulation of LPV-DS on the A-shaped trajectory under passive-DS controller. In all figures, the black dash-dot line indicates the moment when the energy of the energy tank is depleted. (a) Schematic diagram of the energy change curve of the energy tank. (b) The 2D position profile of robot end-effector. (c) The velocity profile of robot end-effector. (d) Schematic diagram of robot joint torque.}}
\end{figure*}

To illustrate the importance of conservativeness for the controller, a non-conservative DS is generated based on the LPV for the A-shaped trajectory. The result is observed in simulation to show the system's performance after energy depletion, as shown in Figure~\ref{fig:EnergyA}. The controller frequency is set to 1000 Hz in the simulation, which is consistent with the actual physical system.

Figure~\ref{fig:EnergyTankA} illustrates the energy evolution in the energy tank, which is depleted around 1.8 s due to the presence of the non-conservative component. Subsequently, due to the charging term ${{\boldsymbol{\dot \xi }}^T}{\mathbf{D}}({\boldsymbol{\xi }}){\boldsymbol{\dot \xi }}$ in (\ref{eq:EnergyTank1B}), the energy tank switches between fast charging and discharging, which is shown as the oscillation in Figure~\ref{fig:EnergyTankA}.
The oscillation of the energy $s$ leads to a rapid change in the switching function $\beta({\boldsymbol{\xi }, s})$, which, according to the controller (\ref{eq:EnergyTank1A}), directly leads to an oscillation of the control torque, as shown in Figure~\ref{fig:TorqueA}.
In actual physical systems, controlling torque oscillations can lead to torque divergence, often creating serious safety hazards. Therefore, the energy $s$ should be avoided depletion.
Although the position signal did not produce significant oscillations (see Figure~\ref{fig:PosA}), the oscillations in torque resulted in obvious oscillations in robot velocity (see Figure~\ref{fig:VelA}). Thus, energy depletion affects the dynamic performance of the robotic system. The DS proposed in Section~\ref{sec:Conservative DS} of this paper is conservative and therefore does not consume energy $s$ and lead to oscillations in the signal.
With the above discussion, we illustrate the importance of DS conservativeness (or DS decomposition) on the performance of control systems.

\subsection{Safety in Interaction Task}\label{subsubsec:Exp2}
\begin{figure*}[!h]
	\begin{center}
		\subfigure[\label{fig:Exp2set}]
		{\includegraphics[width=0.90\columnwidth]{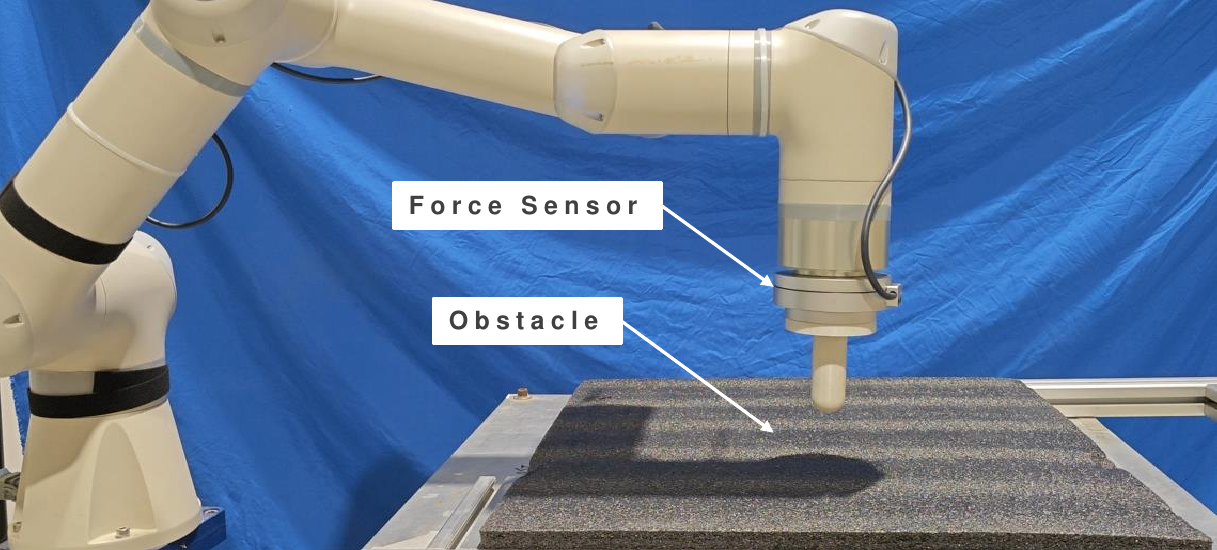}}%
		\hspace{0.00cm}		
		\subfigure[\label{fig:Exp2Set1}]
		{\includegraphics[width=1.05\columnwidth]{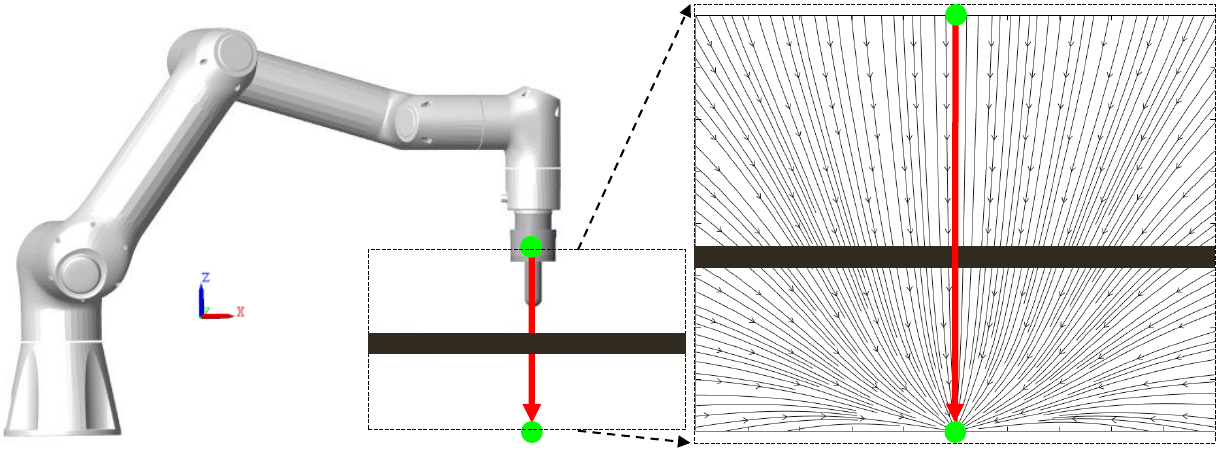}}%
		\vspace{0.00cm}	
	\end{center}
	\caption{\label{fig:Exp2Set} {Schematic diagram of the experiment in Section~\ref{subsubsec:Exp2}. (a) Experimental setup. (b) The schematic of the DS setup in the experiment.}}
\end{figure*}

In this section, we show that the DS framework exhibits interaction safety when the robot has a sudden collision with the environment. The details of the experimental setup are shown in Figure~\ref{fig:Exp2Set}.
A force sensor equipped at the end of the robot is used to detect the real-time interaction force as shown in  Figure~\ref{fig:Exp2set}. The obstacles are placed on the experimental platform. The DS ${\mathbf{f}}\left( {{{\boldsymbol{\xi }}}} \right)$ designed in the experiment is shown in Figure~\ref{fig:Exp2Set1}, and the reference trajectory is a vertical downward straight line crossing the obstacle.

\begin{figure*}[!h]
	\begin{center}
		\subfigure[\label{fig:ExpHeight}]
		{\includegraphics[width=0.85\columnwidth]{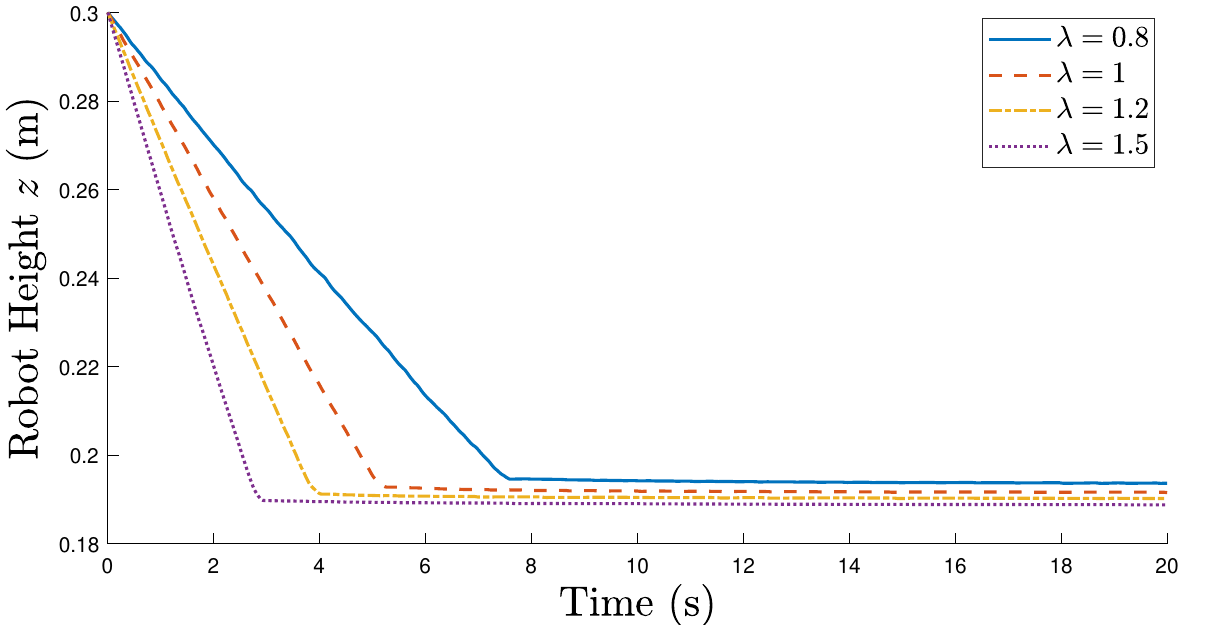}}%
		\hspace{0.01cm}
		\subfigure[\label{fig:ExpContactforce}]
		{\includegraphics[width=0.85\columnwidth]{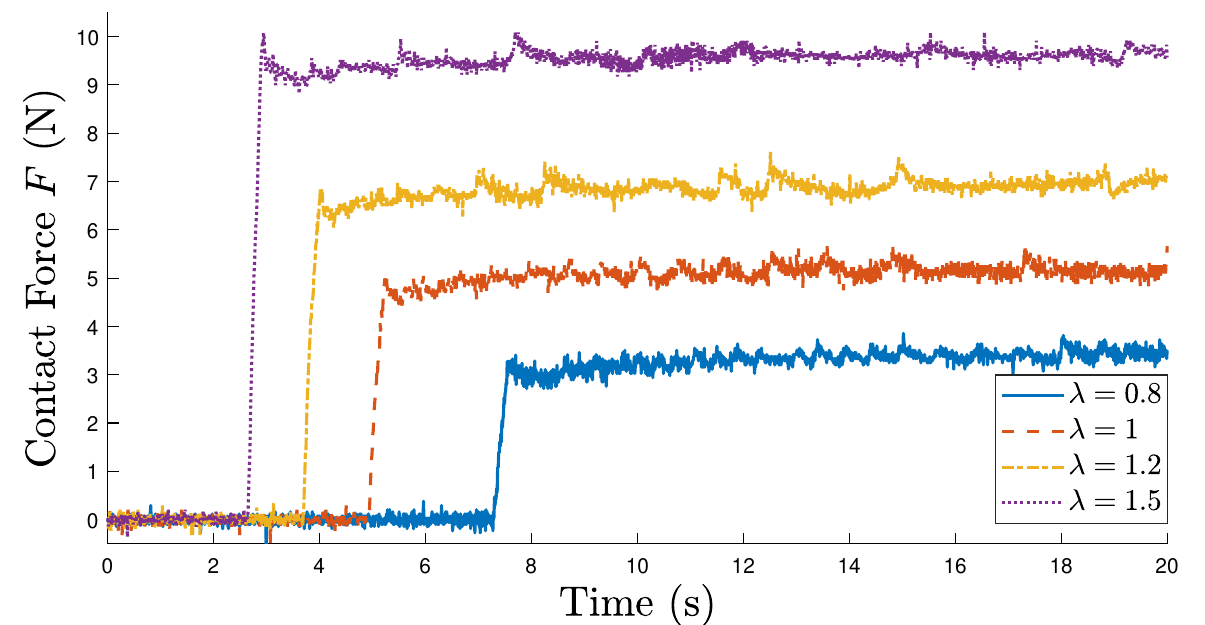}}
		\vspace{0.00cm}		
	\end{center}
	\caption{\label{fig:ExpContactForce} {Schematic of the results of the interaction task experiment. In the experiment, the robot collided with the environment at four different velocities. (a) The profile of robot end-effector height change. (b) The profile of robot-environment interaction force.}}
\end{figure*}

In the experiment, in order to detect the interaction of the robot with the environment at different collision velocities, we control the robot using a variable magnitude DS $\lambda {\mathbf{f}}\left( {{{\boldsymbol{\xi }}}} \right)$, where $\lambda$ is the gain coefficient.
We set up four different experimental conditions $\lambda = 0.8, 1, 1.2, 1.5$, and the robot starts moving from the same height in each test, and the height profile of the robot end is shown in Figure~\ref{fig:ExpHeight}.
The higher the gain coefficient $\lambda$, the larger the velocity of the vector field ${\mathbf{f}}\left( {{{\boldsymbol{\xi }}}} \right)$, the faster the robot's height decreases, and the larger the interaction force between the robot and the environment according to the controller (\ref{eq:EnergyTank1A}). As shown in Figure~\ref{fig:ExpHeight}, as the gain coefficient increases, the slope of the descent curve is larger and the robot is stabilized at a lower height due to the larger interaction forces. Figure~\ref{fig:ExpContactforce} demonstrates the interaction safety for different parameters with fast stabilization of constant interaction force after the robot collides with the environment. This is because when the robot stops after a collision with the environment, the torque value of the Passive-DS controller (\ref{eq:PassiveDSContr}) in the DS framework depends only on the current state ${\boldsymbol{\xi }}$ of the robot. Notice that, unlike conventional impedance control, when the robot stops due to an obstacle, the error between the actual and the reference positions gradually increases, leading to an increasing interaction force.

\subsection{Dynamical System Decomposition}\label{subsubsec:Exp3}
In practical applications and research, non-conservative DS is often widely used in trajectory generation and impedance control. However, due to the non-conservative property of DS, the performance of the controller suffers limitations when the energy tank is depleted. As shown in Section~\ref{subsubsec:DecompDS}, a reasonable DS decomposition avoids depletion of the energy tank and ensures the controller's performance while maintaining passivity. To illustrate the validity of the proposed decomposition index in Section~\ref{subsubsec:DecompDS}, we validate it through a series of experiments.

\begin{figure*}[!h]
	\begin{center}
		\subfigure[\label{fig:DSLpvSnake}]
		{\includegraphics[width=1.05\columnwidth]{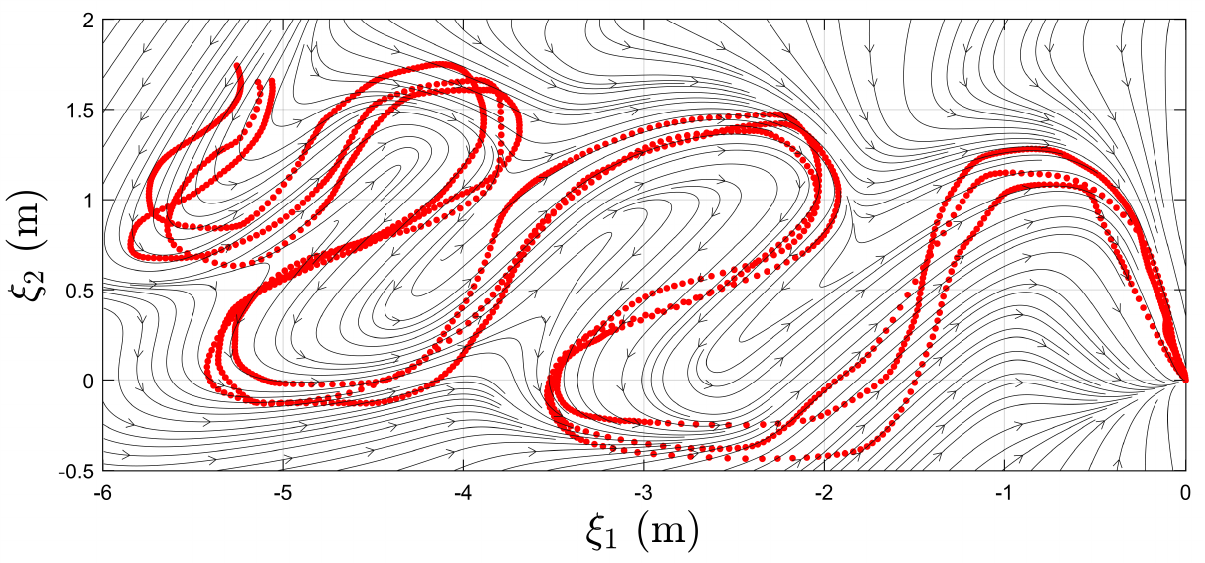}}%
		\hspace{0.01cm}
		\subfigure[\label{fig:ExpPathSnake}]
		{\includegraphics[width=0.845\columnwidth]{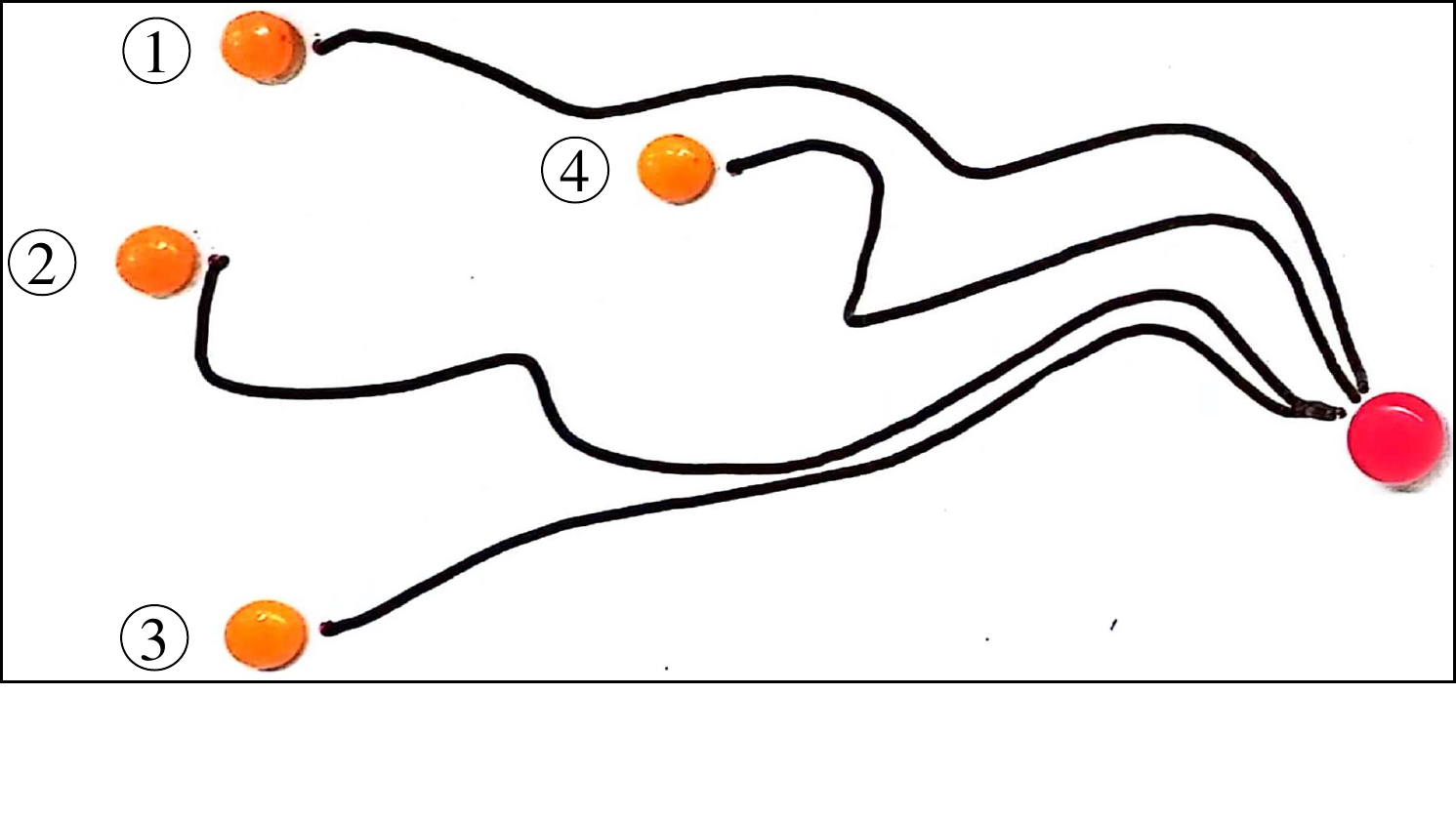}}
		\vspace{0.00cm}		
	\end{center}
	\caption{\label{fig:Exp3Set} {Schematic of the original (pre-decomposition) DS in Section~\ref{subsubsec:Exp3}. In the experiment the robot started from four fixed starting points (orange circles) while observing the consumption of energy $s$ in the energy tank. (a) DS corresponding to the snake trajectory. The red dots represent demonstration trajectories. (b) Schematic of the four  integral paths corresponding to the DS on the whiteboard.}}
\end{figure*}

The experimental setup is the same as in Section~\ref{subsubsec:Exp1}, where we learn a non-conservative DS based on the LPV method on a snake-shaped demonstration trajectory as shown in Figure~\ref{fig:DSLpvSnake}. The twisted character of the snake-shaped trajectory makes the generated DS strongly nonlinear, which is helpful to illustrate the effectiveness of the decomposition method.
We have chosen four representative motion starting points to show that the vector field decomposition is valid for different motion paths. As shown in Figure~\ref{fig:ExpPathSnake}, the four different starting points correspond to four motion paths. During the motion, we observe the change of energy $s$ in the energy tank in real time.

\begin{figure}[!ht]
	\centering
	\includegraphics[width=1\columnwidth]{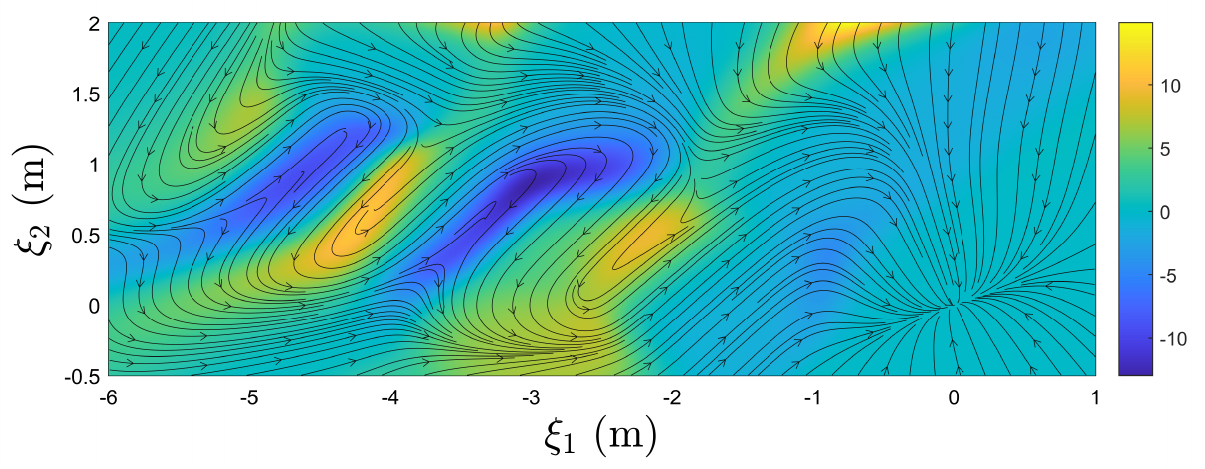}
	\caption{Schematic of the original (pre-decomposition) DS in Section~\ref{subsubsec:Exp3}. The color of each point in the figure indicates the value of the angular velocity of the vector field at the corresponding point.}
	\label{fig:SnakeOmega}
\end{figure}

To show the DS decomposition more clearly, we use the pseudocolor plot to show more information. Figure~\ref{fig:SnakeOmega} shows the pseudocolor plot of the DS in Figure~\ref{fig:DSLpvSnake}, and the color of each point in the plot indicates the magnitude of the angular velocity of the 2D vector field. Since the DS before decomposition is non-conservative, the colors show that the angular velocity in space is nonzero.

\begin{figure*}[!ht]
	\begin{center}
		\subfigure[\label{fig:SnakeOmegaCon10}]
		{\includegraphics[width=1\columnwidth]{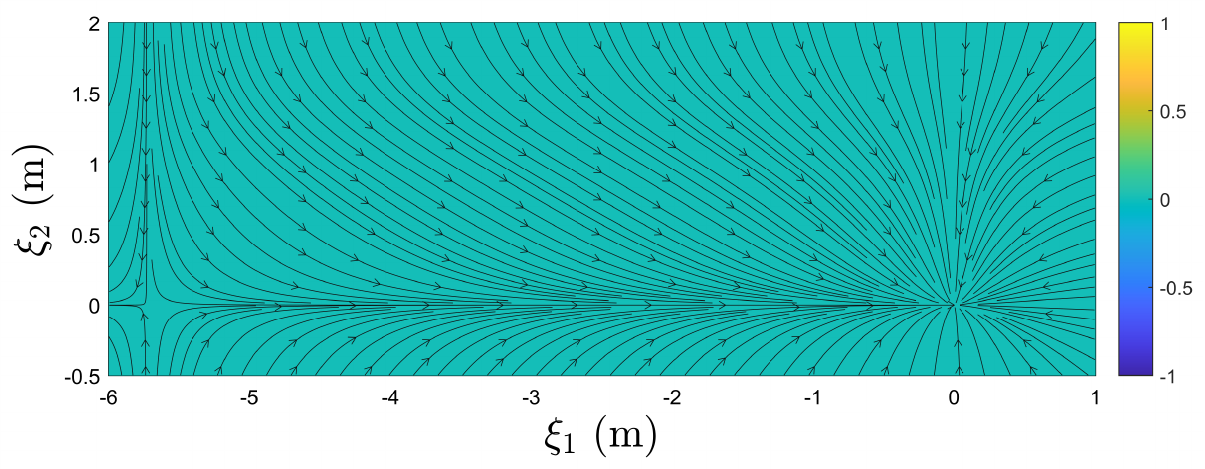}}%
		\hspace{0.01cm}
		\subfigure[\label{fig:SnakeOmegaNonCon10}]
		{\includegraphics[width=1\columnwidth]{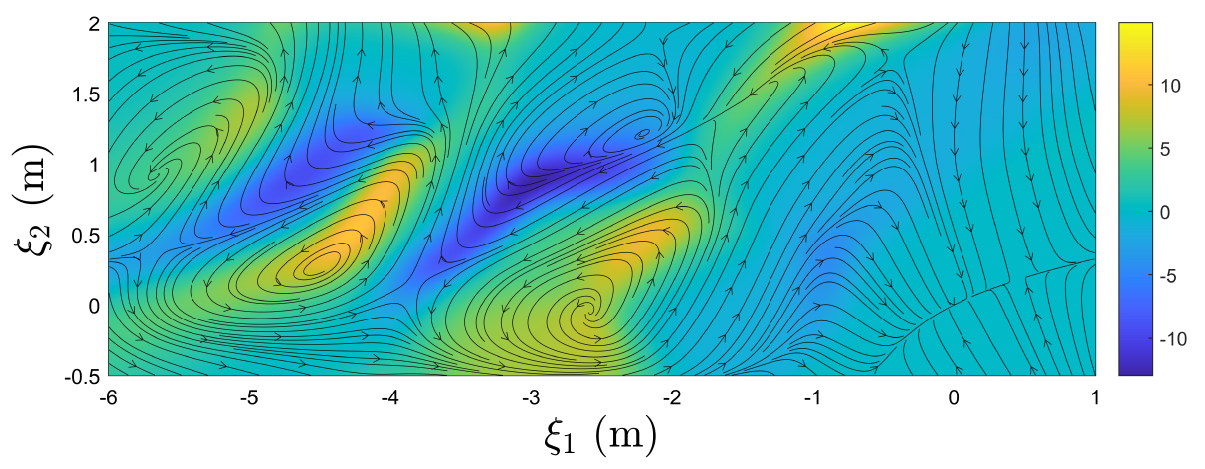}}
		\vspace{0.00cm}		
		\subfigure[\label{fig:SnakeOmegaCon11}]
		{\includegraphics[width=1\columnwidth]{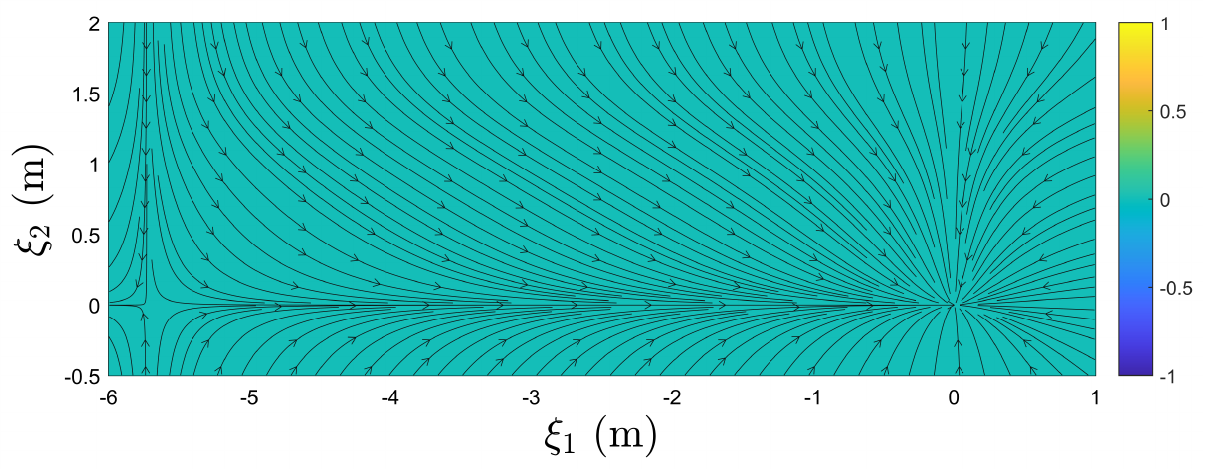}}%
		\hspace{0.01cm}
		\subfigure[\label{fig:SnakeOmegaNonCon11}]
		{\includegraphics[width=1\columnwidth]{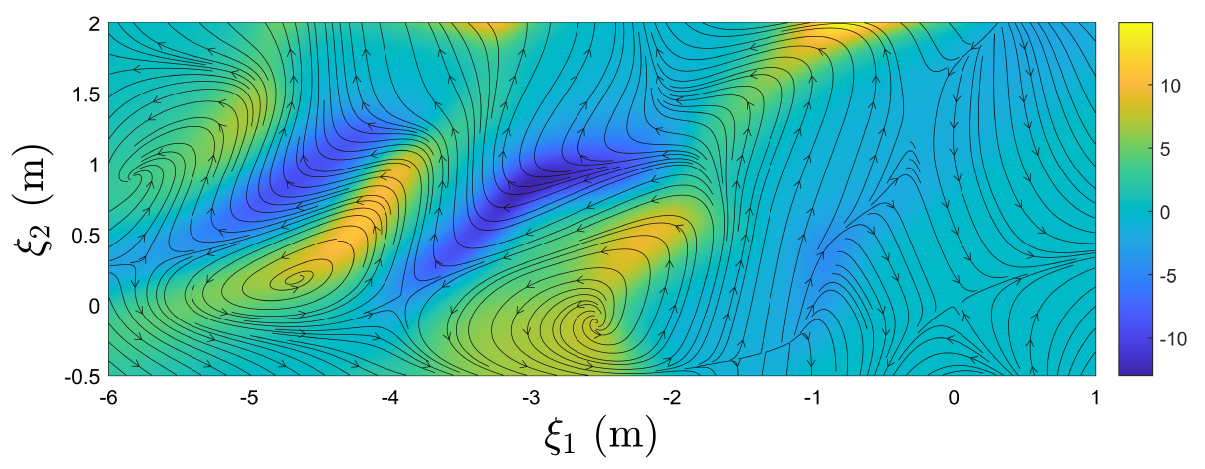}}
		\vspace{0.00cm}	
	\end{center}
	\caption{\label{fig:SnakeDecomp} {Schematic of DS decomposition in Section~\ref{subsubsec:Exp3}. The color of each point in the figure indicates the value of the angular velocity of the vector field at the corresponding point. Different decomposition methods and parameters correspond to different decomposition results: conservative DS (a) and non-conservative DS (b) correspond to ${\boldsymbol{\omega }} = \left[ {1,0} \right]$, and conservative DS (c) and non-conservative DS (d) correspond to ${\boldsymbol{\omega }} = \left[ {1,1} \right]$. (a) The conservative DS correspond to ${\boldsymbol{\omega }} = \left[ {1,0} \right]$.  (b) The non-conservative DS correspond to ${\boldsymbol{\omega }} = \left[ {1,0} \right]$. (c) The conservative DS correspond to ${\boldsymbol{\omega }} = \left[ {1,1} \right]$. (d) The non-conservative DS correspond to ${\boldsymbol{\omega }} = \left[ {1,1} \right]$.}}
\end{figure*}

We first decompose the vector field using the most intuitive index (\ref{eq:decompindex1}), when the weight vector is ${\boldsymbol{\omega }} = \left[ {1,0} \right]$. The conservative part and the non-conservative part are shown in Figure~\ref{fig:SnakeOmegaCon10} and Figure~\ref{fig:SnakeOmegaNonCon10}, respectively. The angular velocity of the conservative part is constant and equal to zero in space.

Further, we take the controller structure into account by choosing ${\boldsymbol{\omega }} = \left[ {1,1} \right]$ and using the index (\ref{eq:decompindex}) for the decomposition. The decomposed conservative and non-conservative parts are shown in Figure~\ref{fig:SnakeOmegaCon11} and Figure~\ref{fig:SnakeOmegaNonCon11}, respectively.

It is worth noting that the vector field streamlines in Figure~\ref{fig:SnakeOmegaNonCon10} and Figure~\ref{fig:SnakeOmegaNonCon11} are different, which means that different decompositions correspond to different conservative and non-conservative parts. However, for the two-dimensional vector field, the angular velocity of the non-conservative part after different ways of decomposition is fixed. Figure~\ref{fig:SnakeOmega}, Figure~\ref{fig:SnakeOmegaNonCon10} and Figure~\ref{fig:SnakeOmegaNonCon11} all have the same pseudocolor distribution. See Appendix~\ref{apd:AngularVelocity} for the details.

\begin{figure*}[!ht]
	\begin{center}
		\subfigure[\label{fig:EnergyTankNoDecmp}]
		{\includegraphics[width=0.65\columnwidth]{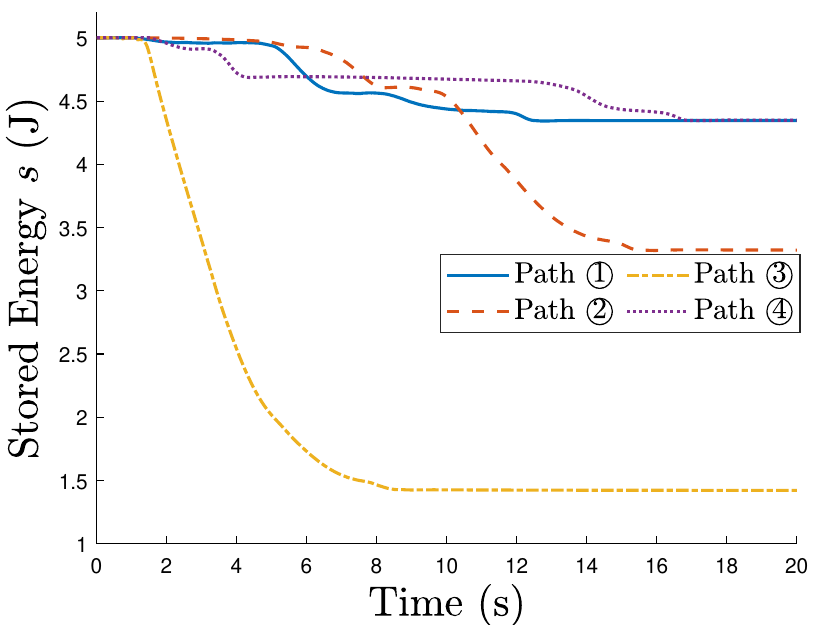}}%
		\hspace{0.01cm}
		\subfigure[\label{fig:EnergyTankDecmp10}]
		{\includegraphics[width=0.65\columnwidth]{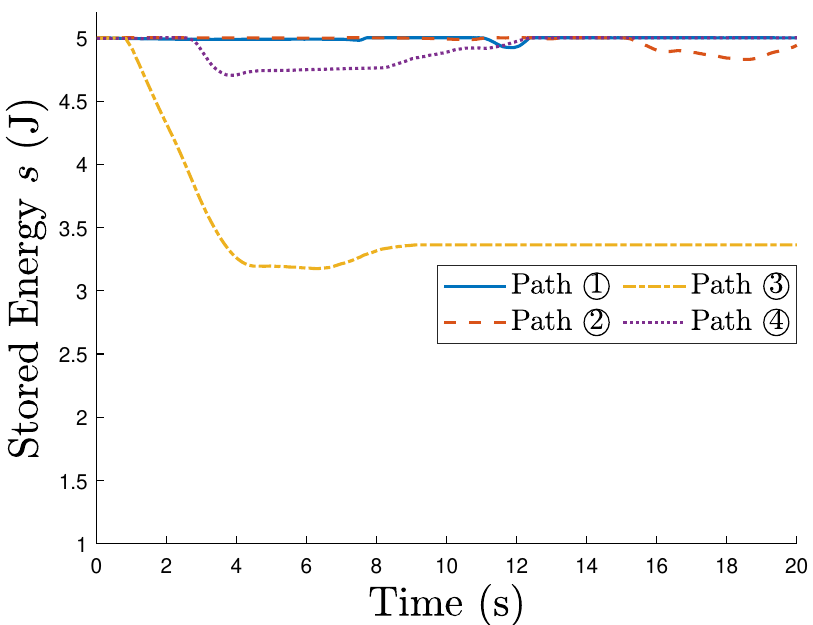}}
		\hspace{0.00cm}		
		\subfigure[\label{fig:EnergyTankDecmp11}]
		{\includegraphics[width=0.65\columnwidth]{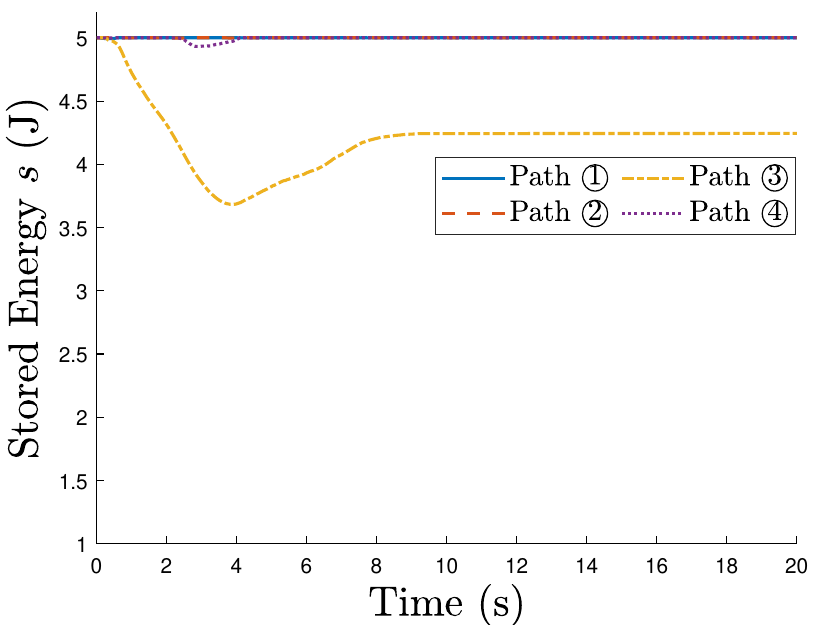}}%
		\vspace{0.00cm}	
	\end{center}
	\caption{\label{fig:EnergyTankSnake} {Schematic illustration of energy dissipation in the energy tank under different DS decomposition strategies. For each strategy, the robot moves along the representative four paths in Fig.~\ref{fig:ExpPathSnake}. (a) The profile of the energy change corresponding to the DS in Fig.~\ref{fig:DSLpvSnake} without decomposition. (b) The profile of the energy change corresponding to the decomposition strategy ${\boldsymbol{\omega }} = \left[ {1,0} \right]$. (c) The profile of the energy change corresponding to the decomposition strategy ${\boldsymbol{\omega }} = \left[ {1,1} \right]$.}}
\end{figure*}

In the experiment, we performed the experiments under three cases: 1) DS without decomposition. 2) Using the decomposition strategy ${\boldsymbol{\omega }} = \left[ {1,0} \right]$. 3) Using the decomposition strategy ${\boldsymbol{\omega }} = \left[ {1,1} \right]$. For each case, the robots move along the four paths in Figure~\ref{fig:ExpPathSnake} while the energy change in the energy tank is recorded as shown in Figure~\ref{fig:EnergyTankSnake}. The parameters of the energy tank are set to $\left[ {{s^l},{s^u}} \right]=\left[ 0,5 \right]$ (unit: J).

In Figure~\ref{fig:EnergyTankNoDecmp}, since the DS is not decomposed, the non-conservative part ${{\mathbf{f}}_{nc}}({\boldsymbol{\xi }})$ leads to a rapid energy depletion rate $\dot{s}$, and the energy curves corresponding to the four paths decrease rapidly.

In Figure~\ref{fig:EnergyTankDecmp10}, when the decomposition index (\ref{eq:decompindex1}) is used, the energy consumption rate decreases, and there is also an energy charging. This is because through rational decomposition, at some moments, the charging rate ${{\boldsymbol{\dot \xi }}^T}{\mathbf{D}}({\boldsymbol{\xi }}){\boldsymbol{\dot \xi }}$ in (\ref{eq:EnergyTank1B}) is greater than the consumption rate ${{{\boldsymbol{\dot \xi }}}^T}{\mathbf{f}_{nc}}({\boldsymbol{\xi }})$. At this time, the energy curves corresponding to all four pathways decrease more slowly than those in Figure~\ref{fig:EnergyTankNoDecmp}.

In Figure~\ref{fig:EnergyTankDecmp11}, after using the decomposition strategy ${\boldsymbol{\omega }} = \left[ {1,1} \right]$, the energy decreases at a slower rate compared to Figure~\ref{fig:EnergyTankDecmp10}, and most of the energy curves are finally maintained at the upper limit $s^u$.

Through the above discussion, we illustrate the rationality and effectiveness of the decomposition index. In particular, the index ${J_2}$ (\ref{eq:decompindex}) by considering the controller structure makes the decomposition more effective.

\subsection{Dynamical Systems for Closed ${\text{\&}}$ Self-intersecting Motions}\label{subsubsec:Exp4}
In Section~\ref{subsubsec:projectds}, we discuss the DS generation method for closed ${\text{\&}}$ self-intersecting motions.
In Section~\ref{subsubsubsec:Exp41}, we verify the validity of the method on planar motion. Further, the superiority of the proposed method over other methods is verified on curved surface motion in Section~\ref{subsubsubsec:Exp42}.
\subsubsection{Motion on the Plane}\label{subsubsubsec:Exp41}
\begin{figure*}[!ht]
	\begin{center}
		\subfigure[\label{fig:CircleSpiral}]
		{\includegraphics[width=0.28\columnwidth]{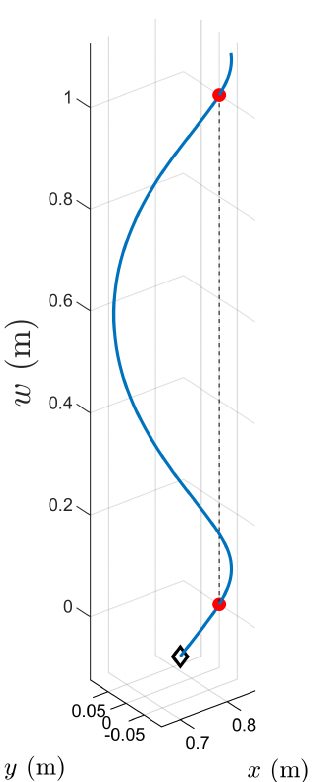}}%
		\hspace{0.01cm}
		\subfigure[\label{fig:CircleLyap}]
		{\includegraphics[width=0.36\columnwidth]{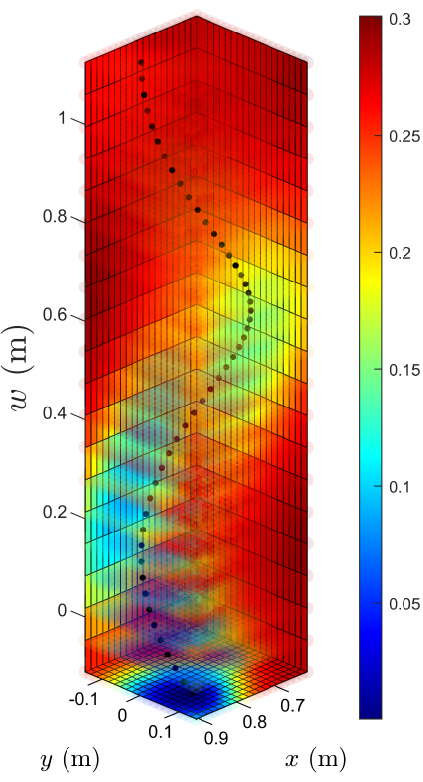}}
		\hspace{0.00cm}		
		\subfigure[\label{fig:3DCircleds}]
		{\includegraphics[width=0.27\columnwidth]{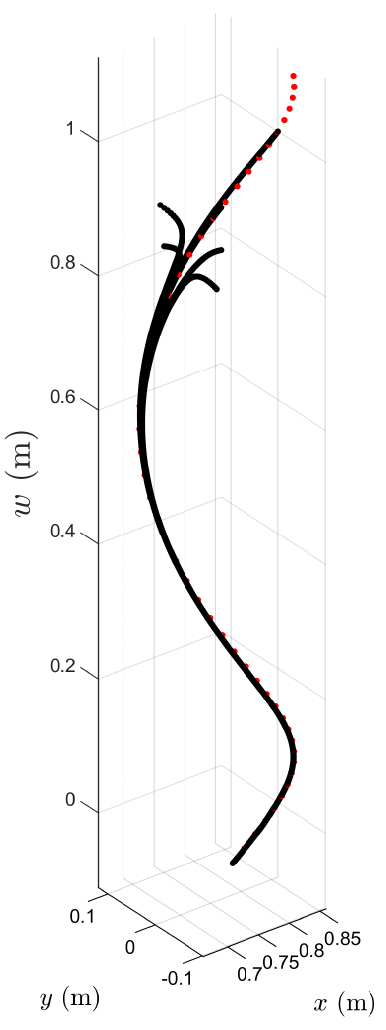}}
		\hspace{0.00cm}		
		\subfigure[\label{fig:VelConDSA3DCircle}]
		{\includegraphics[width=1\columnwidth]{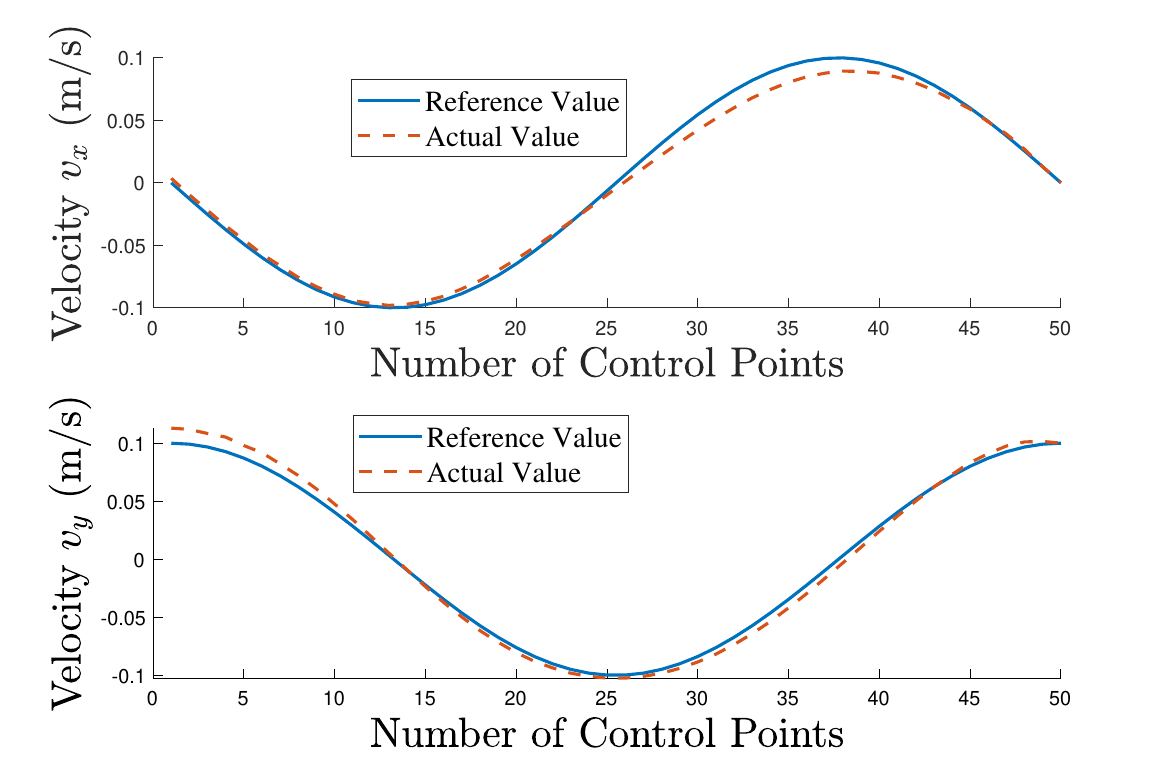}}%
		\vspace{0.00cm}	
	\end{center}
	\caption{\label{fig:3DCircleDS} {Schematic of the process of DS generation in high-dimensional space ${\mathbb{R}^{n+1}}$. $w$ is the virtual coordinate. (a) Reference trajectory in high-dimensional space. (b) Potential energy function generated in high-dimensional space. The color indicates the value of the potential energy function. (c) DS generated in high-dimensional space. The black line represents the integral curve. (d) The velocity comparison. The actual velocity represents the result of learning through the GP.}}
\end{figure*}

$\\$
We introduce the generation process of the projection DS using the closed circle trajectory in Section~\ref{subsubsec:projectds} as an example, as shown in Figure~\ref{fig:3DCircleDS}.

With the elevated dimension operation in Figure~\ref{fig:HigherDim}, the reference trajectory in the high-dimensional space is shown as the spiral in Figure~\ref{fig:CircleSpiral}, and the trajectory between the two red dots corresponds to the circular trajectory in the low dimension.
The potential energy function $V\left( {{\boldsymbol{\xi }},{\xi _{vir}}} \right)$ in Figure~\ref{fig:CircleLyap} is obtained by the iterative optimization process in Figure~\ref{fig:OptimIter}.
From Figure~\ref{fig:CircleLyap}, we can see that the potential energy $V\left( {{\boldsymbol{\xi }},{\xi _{vir}}} \right)$ at the demonstration path is much lower than that in other regions. Meanwhile, the potential energy decreases along the demonstration path and reaches the lowest potential energy at the equilibrium point $\boldsymbol{\xi}_0$. The difference in potential energy leads to the symmetric attractiveness of the generated DS, as shown in Figure~\ref{fig:3DCircleds}, where the black lines indicate the integration curves corresponding to different starting points. It should be noted that the projection DS ${{\mathbf{f}}_{pro}}({\boldsymbol{\xi }},{\xi _{vir}})$ is finally obtained by the projection operation of Figure~\ref{fig:LowerDim}. The velocity corresponding to the learned DS is shown in Figure~\ref{fig:VelConDSA3DCircle} to be in good agreement with the reference velocity.

Further, we perform experimental validation based on the projection DS ${{\mathbf{f}}_{pro}}({\boldsymbol{\xi }},{\xi _{vir}})$. The experimental setup is the same as in Section~\ref{subsubsec:Exp1}. We generate the projection DS ${{\mathbf{f}}_{pro}}({\boldsymbol{\xi }},{\xi _{vir}})$ based on the circle reference trajectory and the figure-of-eight reference trajectory, respectively. For the figure-of-eight self-intersecting trajectory, the DS generation process is the same as in Figure~\ref{fig:3DCircleds}.

\begin{figure*}[!ht]
	\begin{center}
		\subfigure[\label{fig:CircleWB}]
		{\includegraphics[width=0.36\columnwidth]{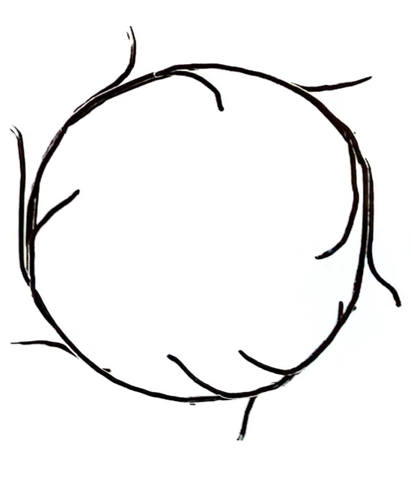}}%
		\hspace{0.01cm}
		\subfigure[\label{fig:CircleExp}]
		{\includegraphics[width=0.56\columnwidth]{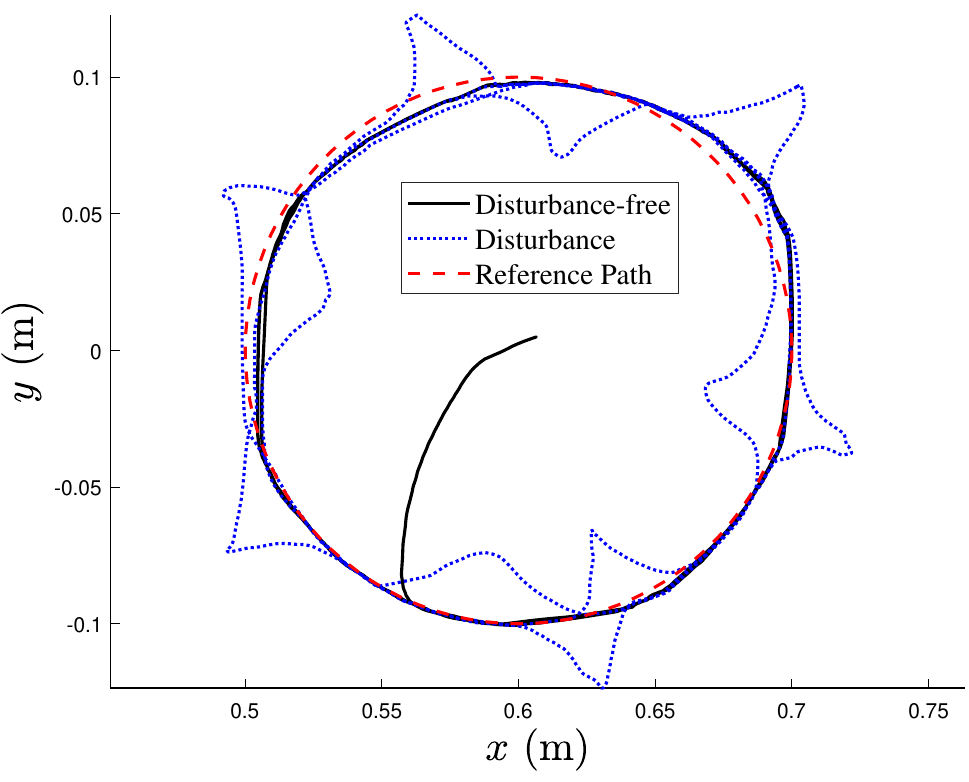}}
		\hspace{0.01cm}
		\subfigure[\label{fig:CircleExp3D}]
		{\includegraphics[width=0.45\columnwidth]{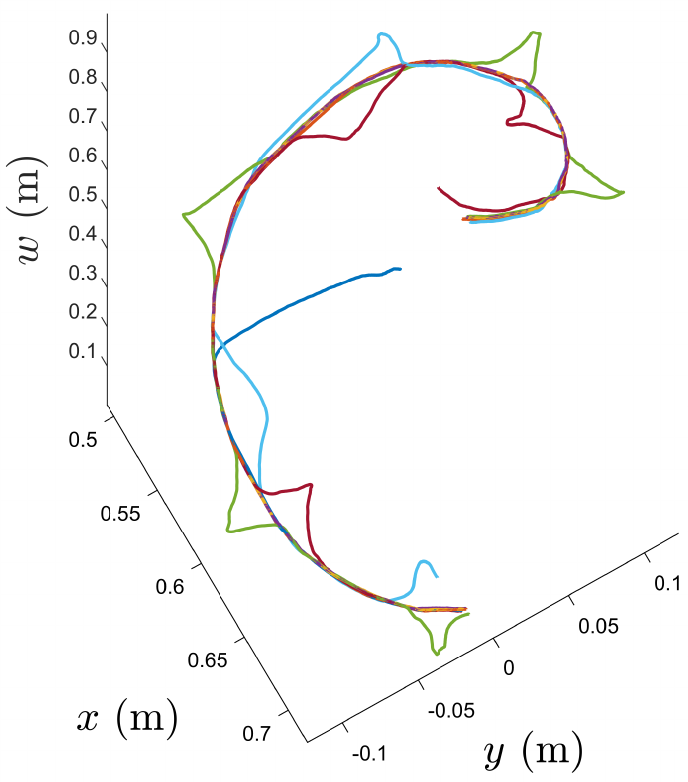}}
		\hspace{0.1cm}
		\vspace{0.00cm}		
		\subfigure[\label{fig:EightWB}]
		{\includegraphics[width=0.64\columnwidth]{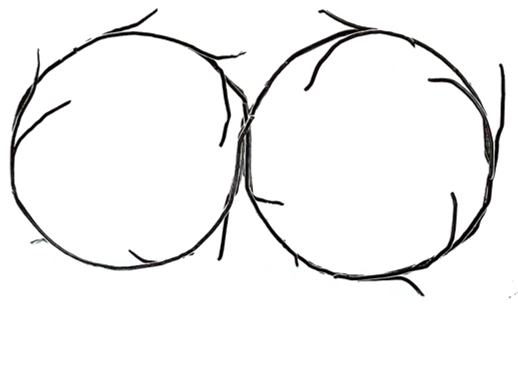}}%
		\hspace{0.01cm}
		\subfigure[\label{fig:EightExp}]
		{\includegraphics[width=0.7\columnwidth]{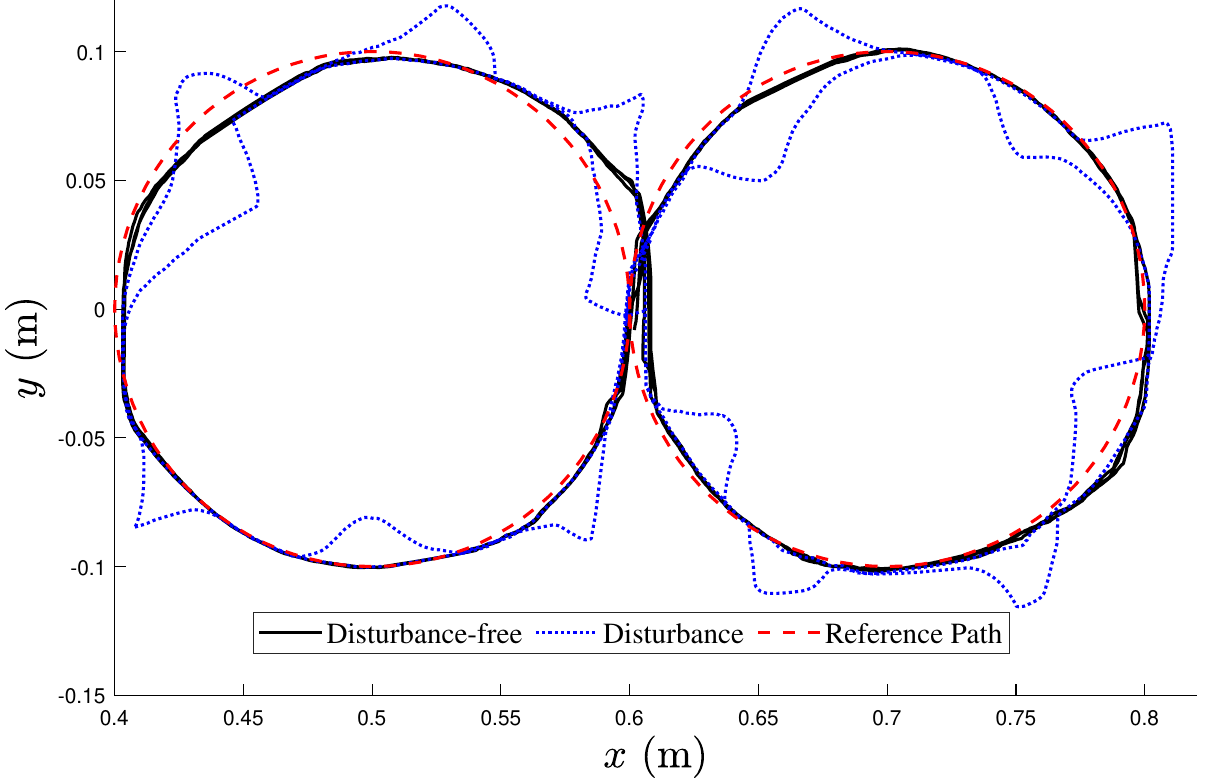}}
		\vspace{0.00cm}	
	\end{center}
	\caption{\label{fig:CEExp} {The experimental results for closed ${\text{\&}}$ self-intersecting motions on the plane. The result of the closed circle motion is shown in (a), (b) and (c). The results of the self-intersecting figure-of-eight movement are displayed in (d) and (e). (a) Trajectories of circular motion on the whiteboard under environmental disturbances. (b) Motion trajectories of circular motions under environmental disturbances. (c) Motion trajectories corresponding to circular motion in higher dimensional space ${\mathbb{R}^{n+1}}$ under environmental disturbances. $w$ is the virtual coordinate. (d) Trajectories of figure-of-eight movement on the whiteboard under environmental disturbances. (e) Motion trajectories of figure-of-eight movement under environmental disturbances.}}
\end{figure*}

Figure~\ref{fig:CircleWB}, Figure~\ref{fig:CircleExp}, and Figure~\ref{fig:CircleExp3D} demonstrate the experimental results for the circle demonstration trajectory. During the experiment, perturbations were applied by frequent human drags. Figure~\ref{fig:CircleWB} shows the robot's trajectory on the whiteboard. Thanks to the symmetric attractiveness, the DS is extremely resistant to perturbations. Figure~\ref{fig:CircleExp} shows the robot motion position data $(x,y)$ collected in the experiment. It can be seen that the actual robot execution trajectory is consistent with the ideal reference circle, and the slight difference comes from the friction between the robot and the whiteboard, as well as other disturbances. To visualize the projection DS, Figure~\ref{fig:CircleExp3D} shows the trajectory evolution of the system in high-dimensional space $(x,y,w)$. The dynamical system exhibits symmetric attraction with respect to the high-dimensional spiral curve, and when perturbed, the system converges along the spiral, which is visualized as convergence to the circular path in the actual physical space $(x,y)$.
Figures~\ref{fig:EightWB} and \ref{fig:EightExp} demonstrate the experimental results for the figure-of-eight demonstration trajectory. For this self-intersecting trajectory, the experimental results show that the proposed projection DS has an excellent anti-perturbation capability while guaranteeing path accuracy.
At the same time, Proposition~\ref{pro:PassiveDSContr1} ensures the passivity of the system under the control of the projection DS ${{\mathbf{f}}_{pro}}({\boldsymbol{\xi }},{\xi _{vir}})$, which facilitates the realization of safe and reliable interaction tasks.

\subsubsection{Motion on the Surface}\label{subsubsubsec:Exp42}
\begin{figure*}[!h]
	\begin{center}
		\subfigure[\label{fig:Exp4set}]
		{\includegraphics[width=0.62\columnwidth]{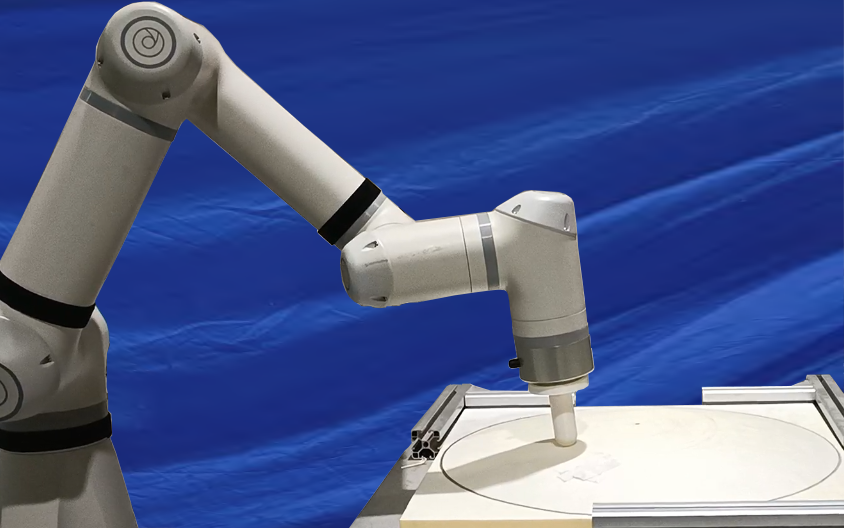}}%
		\hspace{0.00cm}
		\subfigure[\label{fig:SurfaceDs}]
		{\includegraphics[width=0.50\columnwidth]{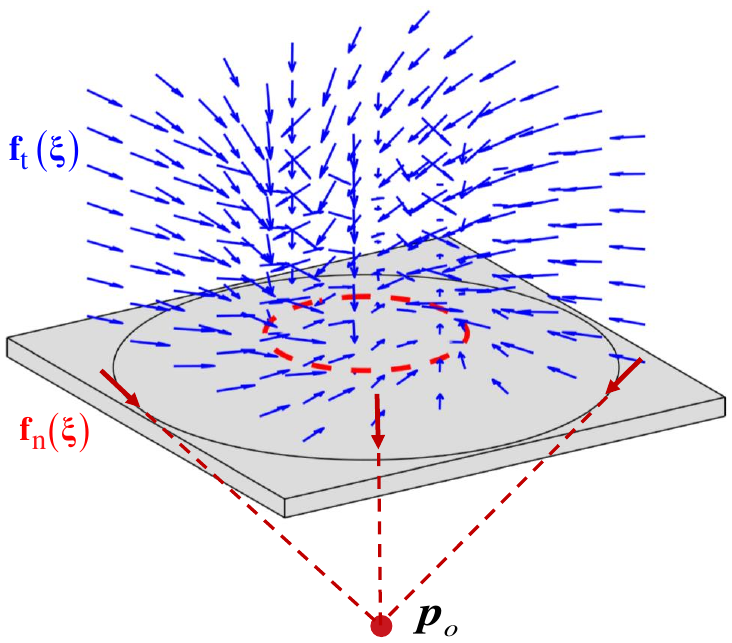}}
		\hspace{0.00cm}
		\subfigure[\label{fig:CurSOmega}]
		{\includegraphics[width=0.80\columnwidth]{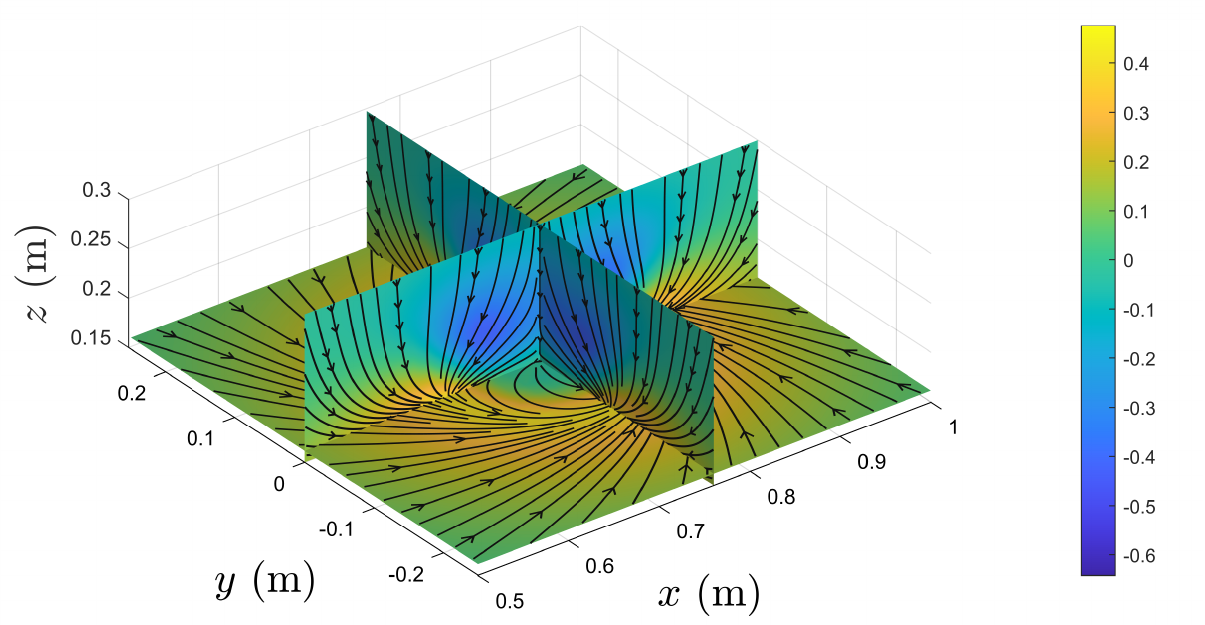}}
		\vspace{0.00cm}		
	\end{center}
	\caption{\label{fig:Exp4Set} {Schematic diagram of the experiment in Section~\ref{subsubsubsec:Exp42}. (a) Experimental setup. (b) The schematic of the original (pre-decomposition) DS in the experiment. ${{\mathbf{f}}_t}\left( {\boldsymbol{\xi }} \right)$ is the nominal DS, and ${{\mathbf{f}}_n}\left( {\boldsymbol{\xi }} \right)$ is the normal modulated DS. (c) The schematic of the original (pre-decomposition) DS ${{\mathbf{f}}_t}\left( {\boldsymbol{\xi }} \right)$ in Section~\ref{subsubsubsec:Exp42}. For clarity, we use three mutually perpendicular slices to show the 3D vector field and its angular velocity distribution. The color of each point in the figure indicates the value of the angular velocity of the vector field at the corresponding point.}}
\end{figure*}

$\\$
In this section, we study the scenario as a robot performing closed motion on a curved surface.
The experimental setup is shown in Figure~\ref{fig:Exp4set}, with the robot moving on a spherical surface. The sphere center is $\boldsymbol{p}_o = {\left[ {\begin{array}{*{20}{c}}{{x_o}}&{{y_o}}&{{z_o}}\end{array}} \right]^T}$.
We want the robot to be in contact with the surface as it moves tangentially along the surface. Therefore, the DS can be designed as follows:
\begin{equation}
	{\mathbf{f}}({\boldsymbol{\xi }}) = {{\mathbf{f}}_t}({\boldsymbol{\xi }}){\text{ + }}{{\mathbf{f}}_n}({\boldsymbol{\xi }}),
\end{equation}
where ${{\mathbf{f}}_t}({\boldsymbol{\xi }})$ is the nominal DS that performs the surface motion and ${{\mathbf{f}}_n}({\boldsymbol{\xi }})$ is the normal modulation DS that keeps the robot in contact with the surface, as shown in Figure~\ref{fig:SurfaceDs}.

The normal modulation DS ${{\mathbf{f}}_n}({\boldsymbol{\xi }})$ is always perpendicular to the surface:
\begin{equation}
	{{\mathbf{f}}_n}({\boldsymbol{\xi }}) = \frac{{{F_d}}}{{{\lambda _1}}}{\mathbf{n}}({\boldsymbol{\xi }}),
\end{equation}
where $F_d$ is the expected contact force and ${\mathbf{n}}({\boldsymbol{\xi }})$ is the surface normal.

As shown by the red dashed line in Figure~\ref{fig:SurfaceDs}, the circle on the sphere with center $\boldsymbol{p}_c = {\left[ {\begin{array}{*{20}{c}}{{x_c}}&{{y_c}}&{{z_c}}\end{array}} \right]^T}$ and radius $r = 0.1\text{ } \text{m}$ was chosen as the reference trajectory in the experiment. We first define the following DS describing the circular motion:
\begin{subequations}
	\begin{align}
		{\mathbf{f}_{{\text{cir}}}}({\boldsymbol{\xi }}) = \left( {\begin{array}{*{20}{c}}
				{{v_t}\left( {r - R} \right)x - {v_w}y} \\ 
				{{v_t}\left( {r - R} \right)y + {v_w}x} \\ 
				{{v_z}\left( {{z_c} - z} \right)} 
		\end{array}} \right),\\
		R = \sqrt {{{\left( {x - {x_c}} \right)}^2} + {{\left( {y - {y_c}} \right)}^2}},
	\end{align}
\end{subequations}
where ${\boldsymbol{\xi }} = {\left[ {\begin{array}{*{20}{c}}
			x&y&z 
	\end{array}} \right]^T}$ is the robot end position.
$v_t$, $v_w$ and $v_z$ denote the normal, tangential and $z$-direction velocity coefficients, respectively.

The nominal DS ${\mathbf{f}_t}({\boldsymbol{\xi }})$ is constructed as follows:
\begin{subequations}
	\label{eq:nominds}
	\begin{align}
		{\mathbf{f}_t}({\boldsymbol{\xi }}) = {v_0}\frac{{{{{\mathbf{\hat f}}_{cir}}({\boldsymbol{\xi }})}}}{{\left\| {{{\mathbf{\hat f}}_{cir}}({\boldsymbol{\xi }})} \right\|}},\\
		{{\mathbf{\hat f}}_{cir}}({\boldsymbol{\xi }}) = {\mathbf{f}_{{\text{cir}}}}({\boldsymbol{\xi }}) - \left( {{{\mathbf{n}}^T}({\boldsymbol{\xi }}){\mathbf{f}_{{\text{cir}}}}({\boldsymbol{\xi }})} \right){\mathbf{n}}({\boldsymbol{\xi }}),
	\end{align}
\end{subequations}
where $v_0$ denotes the magnitude of the tangential velocity of the end-effector. In this experiment, set $F_d = 5\text{ } \text{N}$ and $v_0 = 0.1\text{ } \text{m/s}$.

Since the modulated DS can be considered conservative, we next focus on the analysis of the nominal DS. The pseudocolor plot of the nominal DS ${\mathbf{f}_t}({\boldsymbol{\xi }})$ is illustrated in Figure~\ref{fig:CurSOmega}. It can be seen that the angular velocity of ${\mathbf{f}_t}({\boldsymbol{\xi }})$ is not constant zero and is therefore non-conservative.

\begin{figure*}[!ht]
	\begin{center}
		\subfigure[\label{fig:CurSOmegaCon10}]
		{\includegraphics[width=0.9\columnwidth]{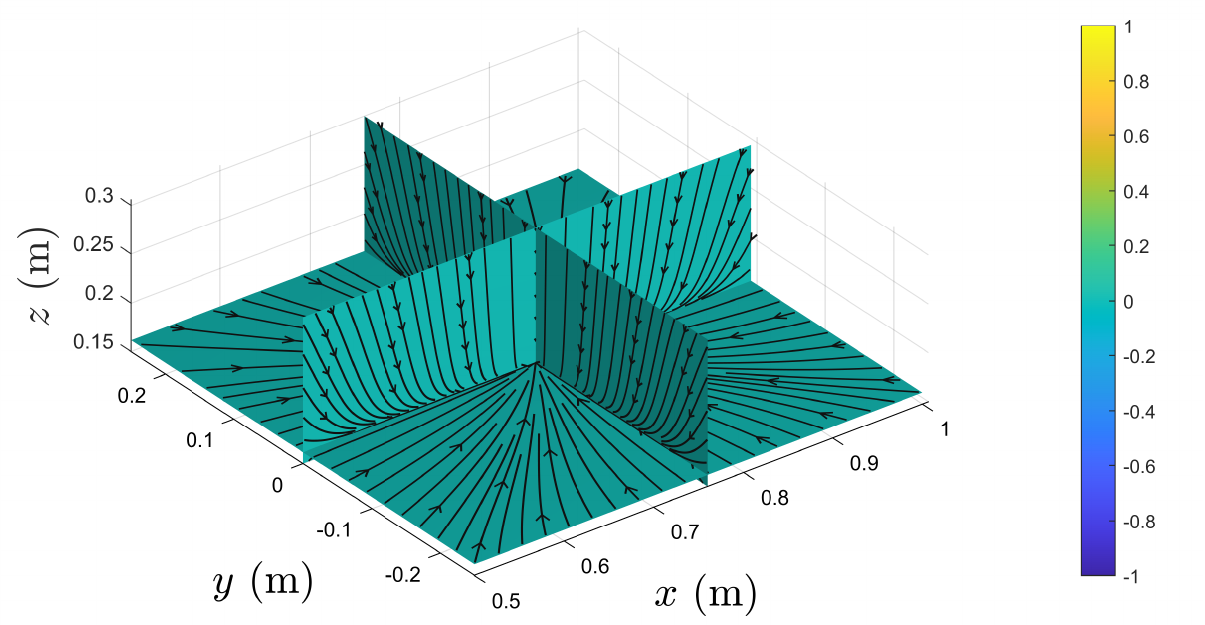}}%
		\hspace{0.00cm}
		\subfigure[\label{fig:CurSOmegaNonCon10}]
		{\includegraphics[width=0.9\columnwidth]{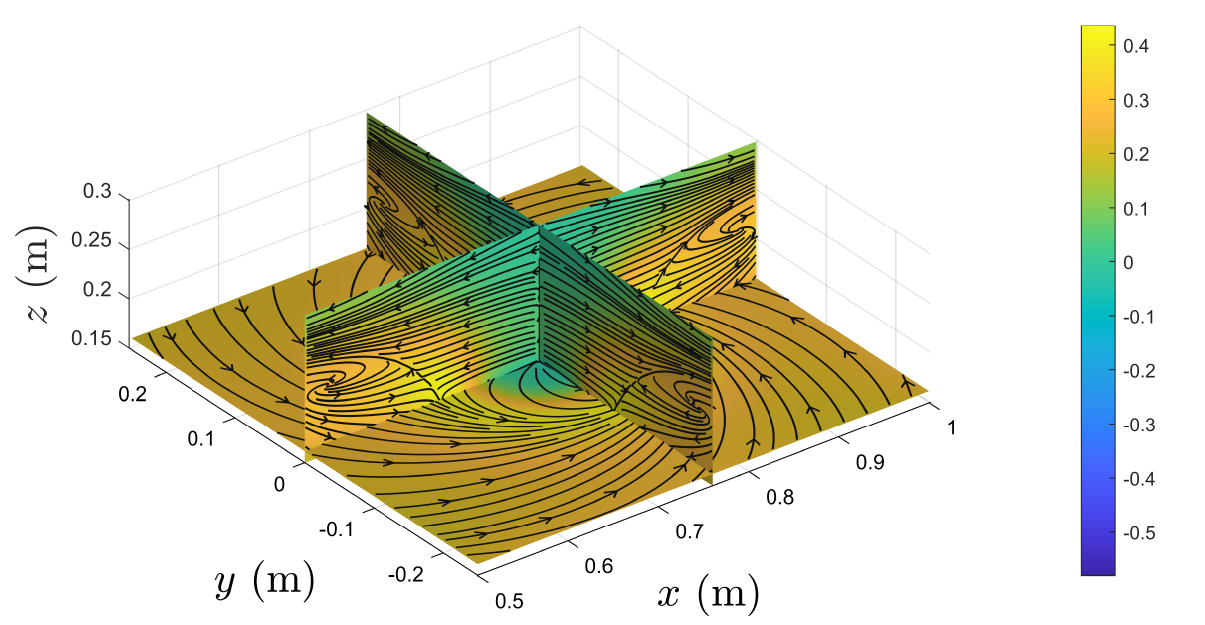}}
		\vspace{0.00cm}		
		\subfigure[\label{fig:CurSOmegaCon11}]
		{\includegraphics[width=0.9\columnwidth]{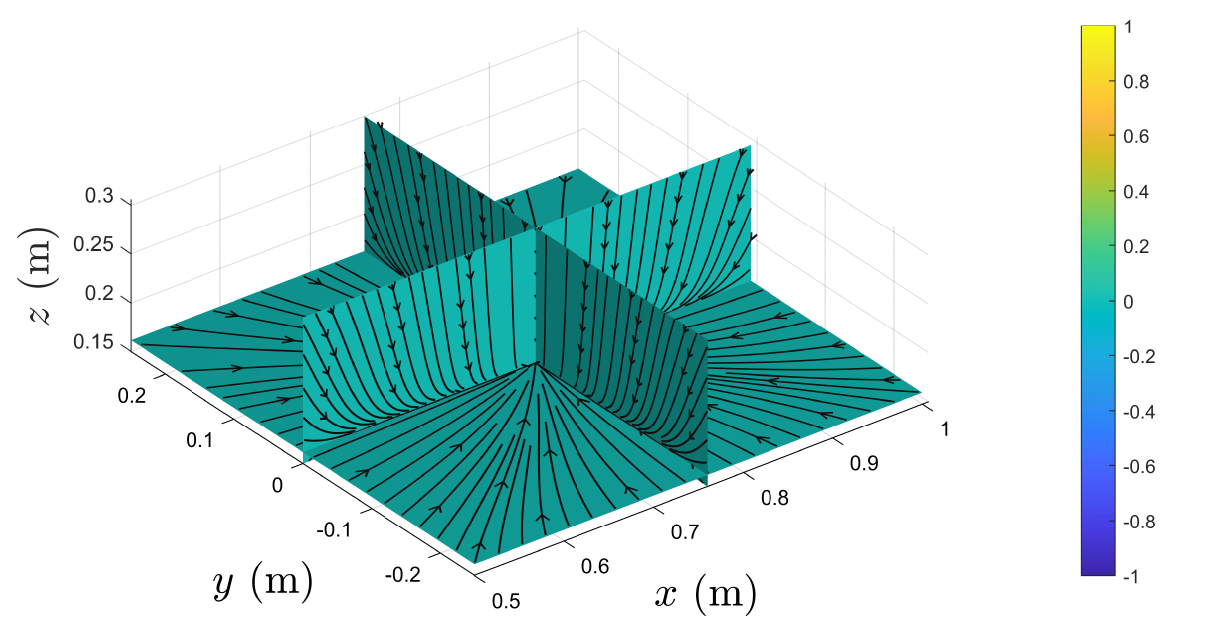}}%
		\hspace{0.00cm}
		\subfigure[\label{fig:CurSOmegaNonCon11}]
		{\includegraphics[width=0.9\columnwidth]{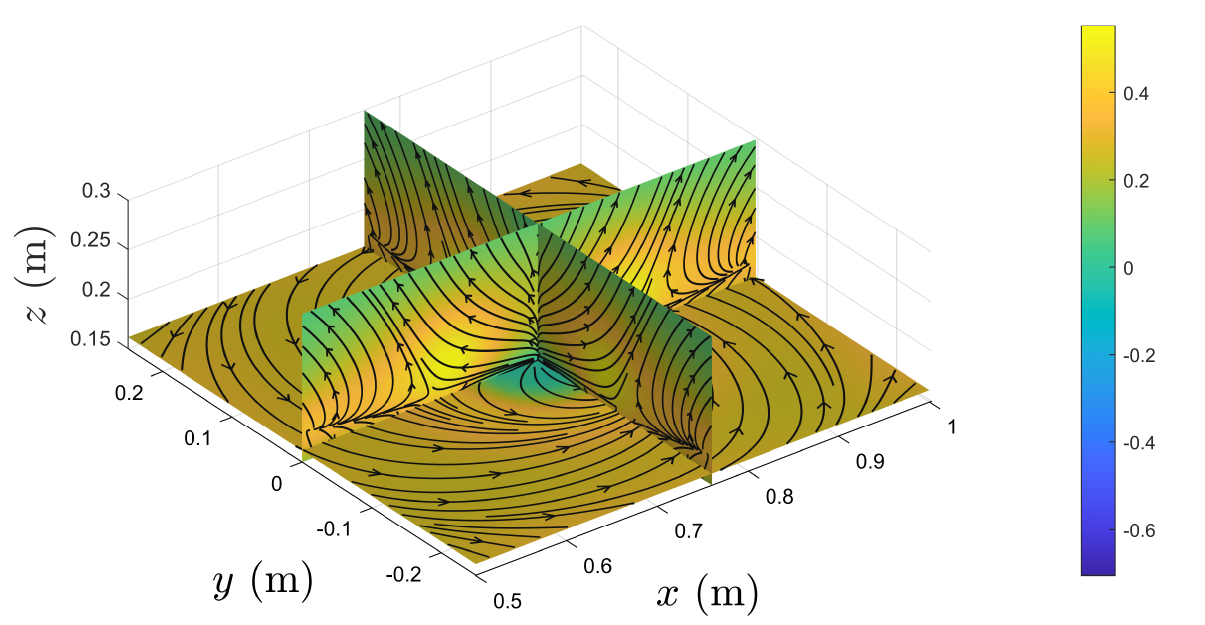}}
		\vspace{0.00cm}	
	\end{center}
	\caption{\label{fig:CurSDecomp} {Schematic of DS decomposition in Section~\ref{subsubsubsec:Exp42}. For clarity, we use three mutually perpendicular slices to show the 3D vector field and its angular velocity distribution. The color of each point in the figure indicates the value of the angular velocity of the vector field at the corresponding point. Different decomposition methods and parameters correspond to different decomposition results: conservative DS (a) and non-conservative DS (b) correspond to ${\boldsymbol{\omega }} = \left[ {1,0} \right]$, and conservative DS (c) and non-conservative DS (d) correspond to ${\boldsymbol{\omega }} = \left[ {1,1} \right]$. (a) The conservative DS correspond to ${\boldsymbol{\omega }} = \left[ {1,0} \right]$.  (b) The non-conservative DS correspond to ${\boldsymbol{\omega }} = \left[ {1,0} \right]$. (c) The conservative DS correspond to ${\boldsymbol{\omega }} = \left[ {1,1} \right]$. (d) The non-conservative DS correspond to ${\boldsymbol{\omega }} = \left[ {1,1} \right]$.}}
\end{figure*}

Since the nominal DS (\ref{eq:nominds}) is non-conservative, the depletion of energy in the energy tank is slowed down by decomposing it. Similar to Section~\ref{subsubsec:Exp3}, we use the decomposition strategies ${\boldsymbol{\omega }} = \left[ {1,0} \right]$ and ${\boldsymbol{\omega }} = \left[ {1,1} \right]$, which correspond to the decomposition index (\ref{eq:decompindex1}) and index (\ref{eq:decompindex}), respectively. The result of the decomposition is shown in Figure~\ref{fig:CurSDecomp}, where it can be seen that the angular velocity of the conservative part is constant to zero. Unlike the 2D case, different decompositions correspond to non-conservative DSs with different angular velocity distributions, as shown in Figure~\ref{fig:CurSOmegaNonCon10} and Figure~\ref{fig:CurSOmegaNonCon11}. This is due to the difference in the definition of the angular velocity of the 3D vector field from the 2D. Details can be found in Appendix~\ref{apd:AngularVelocity}.

For a comprehensive comparison, we compare four controllers in simulation and experiment: 1) Using the nominal DS (\ref{eq:nominds}) without decomposition. 2) Using the nominal DS (\ref{eq:nominds}) and the decomposition strategy ${\boldsymbol{\omega }} = \left[ {1,0} \right]$. 3) Using the nominal DS (\ref{eq:nominds}) and the decomposition strategy ${\boldsymbol{\omega }} = \left[ {1,1} \right]$. 4) Using the projection DS in Section~\ref{subsubsubsec:Exp41} as the nominal DS.

\begin{figure*}[!h]
	\begin{center}
		\subfigure[\label{fig:SimuEnergyCurC}]
		{\includegraphics[width=1\columnwidth]{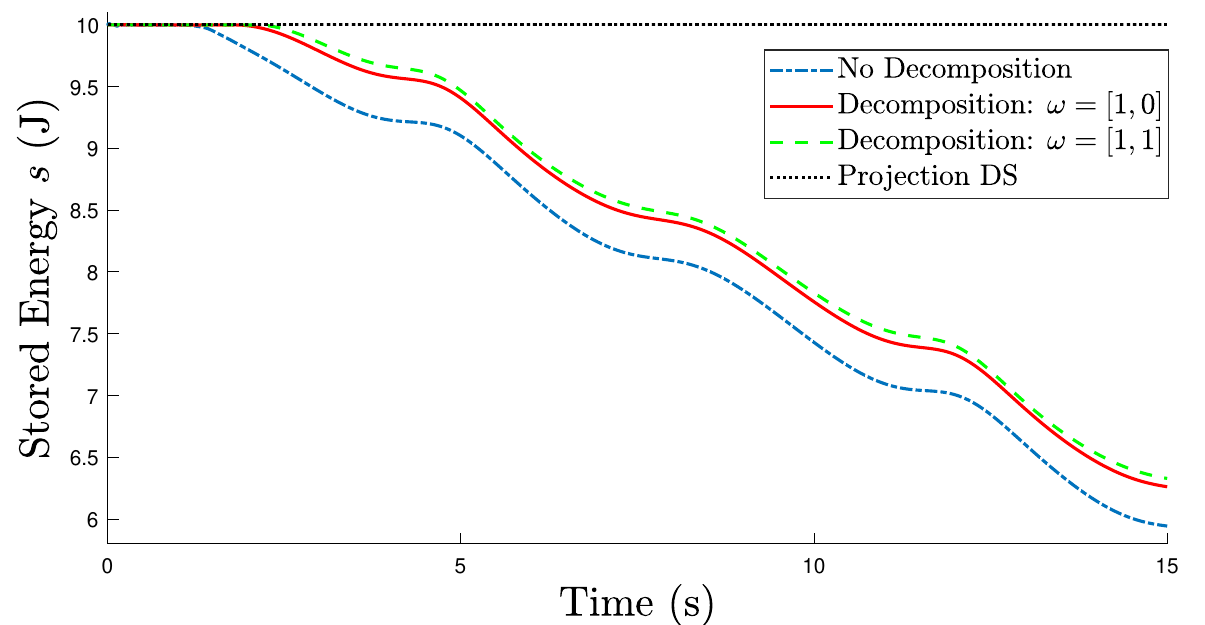}}%
		\hspace{0.01cm}
		\subfigure[\label{fig:ExpEnergyCurC}]
		{\includegraphics[width=1\columnwidth]{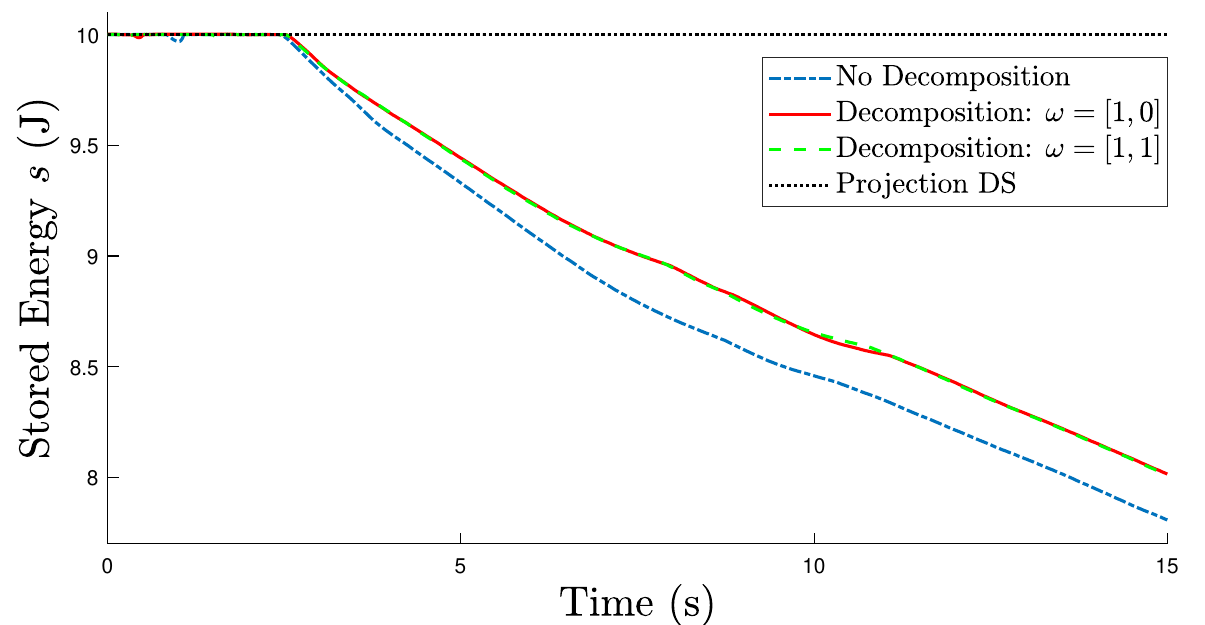}}
		\vspace{0.00cm}		
	\end{center}
	\caption{\label{fig:EnergyCurC} {Schematic illustration of energy dissipation in the energy tank under different strategies. The blue dash-dot line indicates the case where the DS does not decompose. The red solid line corresponds to the decomposition strategy ${\boldsymbol{\omega }} = \left[ {1,0} \right]$. The green dashed line corresponds to the decomposition strategy ${\boldsymbol{\omega }} = \left[ {1,1} \right]$. The black dotted line corresponds to the high-dimensional projection DS. (a) The profile of the energy change in simulation. (b) The profile of the energy change in the experiment.}}
\end{figure*}

Figure~\ref{fig:EnergyCurC} shows the change of the energy for four different strategies. The simulations and experiments show similar trends. The energy decays the fastest when no decomposition is performed. After using the decomposition strategy ${\boldsymbol{\omega }} = \left[ {1,0} \right]$, the energy decay is mitigated, but it still decays very fast. The enhancement obtained by using decomposition strategy ${\boldsymbol{\omega }} = \left[ {1,1} \right]$ is limited. Unlike the case in Section~\ref{subsubsec:Exp3}, the decomposition strategy in this section has a limited effect on suppressing the energy decay. This is because the nominal DS (\ref{eq:nominds}), which represents the circular motion, is a curl field with a small conservative component in this section. Fortunately, the projection DS constructed in Section~\ref{subsubsubsec:Exp41} can maintain the passivity without depleting the energy $s$ (Proposition~\ref{pro:PassiveDSContr1}) and thus handles the special case well.

%% file: sections/7_conclusion.tex

\begin{figure*}[!ht]
	\centering
	\includegraphics[width=1.9\columnwidth]{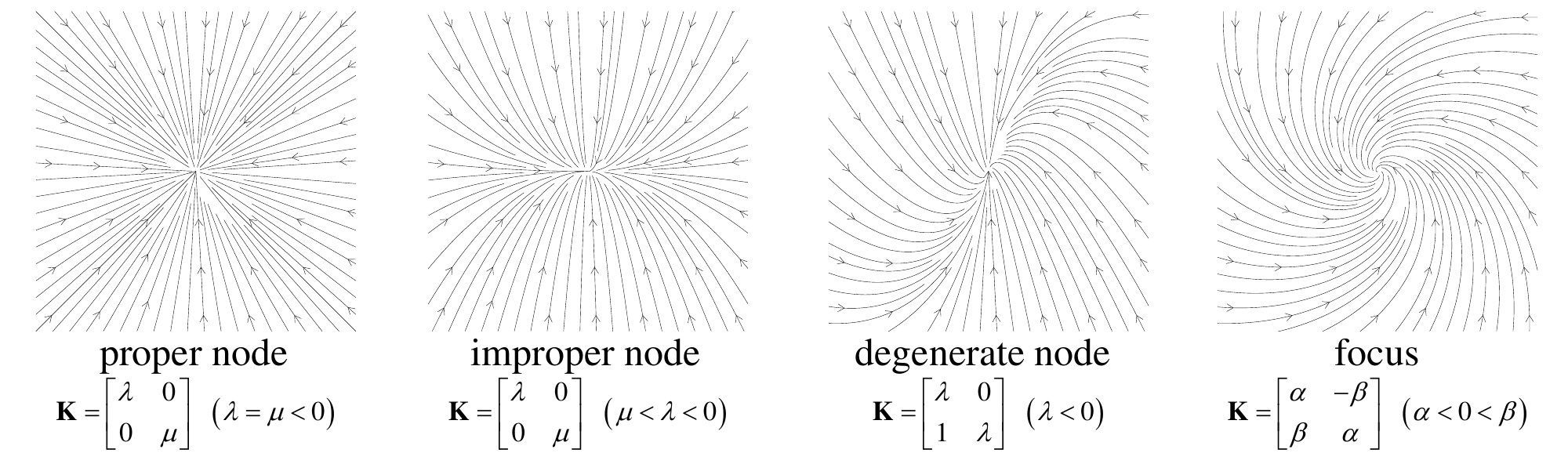}
	\caption{Schematic diagram of the local stiffness encoding. A rational construction can enable a nonlinear system ${\boldsymbol{\dot \xi }} = {\mathbf{f}}\left( {\boldsymbol{\xi }} \right)$ to exhibit the same qualitative structure as a linear system ${\boldsymbol{\dot \xi }} = {{\mathbf{K}}\boldsymbol{\xi }}$ in the local domain of the equilibrium point ${{\boldsymbol{\xi }}_0}$. Different eigenvalue cases of the stiffness matrix ${\mathbf{K}}$ correspond to different patterns: proper node, improper node, degenerate node and focus are illustrated in the figure. }
	\label{fig:LocalStiffEncode}
\end{figure*}

\section{Conclusion}\label{sec:Conclusion}
\subsection{Summary} \label{sec:summary}
In this paper, the stiffness encoding framework for modulating DS is presented. We give the quantitative effect of the combination between symmetry, exactness and negative definiteness (contraction theory) of the stiffness matrix on the DS. For the problem of guaranteeing the passivity of the control system, we obtain the conservative DS based on the conservative stiffness matrix. We use GP to construct structure-specific potential energy functions so that the learned DS has symmetric attraction behavior, and thus, the robotic system has better perturbation resistance. We further extend the conservative DS generation method to SE(3) and to special demonstration trajectories (e.g., closed and self-intersecting trajectories). We design the corresponding impedance control laws for the closed and self-intersecting trajectories to achieve the passivity of the control system without relying on the energy tank. For non-conservative DS, we construct a decomposition index considering the energy tank's structure. By optimizing the index, we decompose the vector field using a conservative stiffness matrix, which effectively slows down the energy decay in the energy tank and improves the stability margin of the system.

We fully validate the theory and methodology of this paper through a series of simulations and experiments. Thanks to the DS's symmetric attraction behavior, the robot shows significant resistance to environmental perturbations when performing point-to-point motion tasks as well as closed and self-intersecting motion tasks. Meanwhile, the system's passivity in various motion scenarios is achieved by carefully designing the DS and the control law, and the robot exhibits advantageous robustness properties in the motion and human-robot interaction experiments.

\subsection{Limitations and future developments} \label{sec:limitanddevelop}
In this study, the controller could control the system in real time when the robot executes the motion. Since the GP involves the computation of the inverse of the matrix, when the spatial dimension of the motion continues to increase and the number of training points of the GP increases, the computation of the controller increases significantly, which may hinder the subsequent more comprehensive application. Moreover, the trajectories involved in this paper cover most of the robot's motion scenarios. However, when the complexity of the demonstration trajectories is further increased, the fitting accuracy will decrease, and the selection and optimization of some hyperparameters will become difficult. Using neural networks or other advanced methods may be an effective way to solve this problem. It is worth noting that other methods can still be naturally combined with the theory and methods in the subsequent sections of this paper after improving the method in Section~\ref{sec:Conservative DS}.

The stiffness encoding framework proposed in Figure~\ref{fig:StiffEncode} focuses on analyzing the global properties (continuity, conservativeness, and contraction) of the DS. Based on stiffness encoding, we can also fine-tune the local properties of the DS, such as the system's qualitative structure in the local domain of the equilibrium point ${{\boldsymbol{\xi }}_0}$, as shown in Figure~\ref{fig:LocalStiffEncode}. When a nonlinear system ${\boldsymbol{\dot \xi }} = {\mathbf{f}}\left( {\boldsymbol{\xi }} \right)$ is dominated by a linear structure ${\boldsymbol{\dot \xi }} = {{\mathbf{K}}\boldsymbol{\xi }}$ near the equilibrium point, the qualitative structure of the system is determined by the linear part~\citep{slotine1991applied}.

\cite{mohammadineural} trained the neural network that generates the negative definite stiffness matrix and subsequently integrated the matrix to obtain the contractive DS. In this paper, our proposed stiffness encoding framework gives a series of important stiffness properties. Inspired by the above work, we can regulate the global and local properties of the DS by training the neural network to generate stiffness matrices of specific properties in the future.

Notice that the set of stiffness matrices with specific properties (symmetric, exact, negative definite, etc.) in this paper has the structure of a manifold or group. Generating the DS using matrices with specific properties can be considered an optimization process on matrix manifolds. Subsequent work can incorporate tools such as differential geometry to guide DS generation from an optimization perspective~\citep{absil2008optimization}.

%% file: appendices/appendix_A.tex

\section{DS Corresponding to the Conservative Stiffness}\label{apd:EfPf}
Next, we show the forms of the elastic force ${\mathbf{f}}\left( {{\boldsymbol{\xi }}} \right)$ and potential function $V\left( {{\boldsymbol{\xi }}} \right)$ corresponding to the conservative stiffness ${\mathbf{K}} = {\left[ {{k_{ij}}} \right]_{n \times n}}$. We derive the following second-order partial differential equation from the symmetry and exactness of the conservative stiffness matrix:
\begin{equation}
	\label{eq:constiffcond}
	\left\{ {\begin{array}{*{20}{c}}
			{\frac{{\partial {k_{ki}}}}{{\partial {\xi ^j}}} = \frac{{\partial {k_{kj}}}}{{\partial {\xi ^i}}}} \\ 
			{{k_{ij}} = {k_{ji}}} 
	\end{array}} \right. \Rightarrow \left\{ {\begin{array}{*{20}{c}}
			{\frac{{\partial {k_{ii}}}}{{\partial {\xi ^j}}} = \frac{{\partial {k_{ij}}}}{{\partial {\xi ^i}}}} \\ 
			{\frac{{\partial {k_{jj}}}}{{\partial {\xi ^i}}} = \frac{{\partial {k_{ji}}}}{{\partial {\xi ^j}}}} 
	\end{array}} \right. \Rightarrow \frac{{{\partial ^2}{k_{ii}}}}{{\partial {{\left( {{\xi ^j}} \right)}^2}}} = \frac{{{\partial ^2}{k_{jj}}}}{{\partial {{\left( {{\xi ^i}} \right)}^2}}}.
\end{equation}
Then the generalized solution for the diagonal elements of ${\mathbf{K}}$ can be expressed as ${k_{ii}}\left( {{\boldsymbol{\xi }}} \right) = \frac{{{\partial ^2}g\left( {\boldsymbol{\xi }} \right)}}{{\partial {{\left( {{\xi ^i}} \right)}^2}}}$, where $g\left( {\boldsymbol{\xi }} \right)$ is an arbitrary differentiable function. The generalization of the non-diagonal elements of ${\mathbf{K}}$ can be further obtained by substituting ${k_{ii}}\left( {{\boldsymbol{\xi }}} \right)$ into (\ref{eq:constiffcond}):
\begin{equation}
	\left\{ {\begin{array}{*{20}{c}}
			{\frac{{\partial {k_{ii}}}}{{\partial {\xi ^j}}} = \frac{{{\partial ^3}g\left( {\boldsymbol{\xi }} \right)}}{{\partial {\xi ^j}\partial {{\left( {{\xi ^i}} \right)}^2}}} = \frac{{\partial {k_{ij}}}}{{\partial {\xi ^i}}}} \\ 
			{\frac{{\partial {k_{jj}}}}{{\partial {\xi ^i}}} = \frac{{{\partial ^3}g\left( {\boldsymbol{\xi }} \right)}}{{\partial {\xi ^i}\partial {{\left( {{\xi ^j}} \right)}^2}}} = \frac{{\partial {k_{ji}}}}{{\partial {\xi ^j}}}} 
	\end{array}} \right. \Rightarrow {k_{ij}} = {k_{ji}} = \frac{{{\partial ^2}g\left( {\boldsymbol{\xi }} \right)}}{{\partial {\xi ^i}\partial {\xi ^j}}}.
\end{equation}
In summary, the generalized solution of the stiffness matrix can be expressed as ${\mathbf{K}}\left( {{\boldsymbol{\xi }}} \right) = {\left[ {\frac{{{\partial ^2}g\left( {\boldsymbol{\xi }} \right)}}{{\partial {\xi ^j}\partial {\xi ^i}}}} \right]_{n \times n}}$.

The conservative elastic force ${\mathbf{f}}\left( {{\boldsymbol{\xi }}} \right)$ can be obtained by integrating conservative stiffness matrix ${\mathbf{K}}\left( {{\boldsymbol{\xi }}} \right)$:
\begin{equation}
	\begin{aligned}
		{\mathbf{f}} 
		&= \int\limits_{{{\boldsymbol{\xi }}_0}}^{\boldsymbol{\xi }} {{{\left[ {\frac{{{\partial ^2}g\left( {\boldsymbol{\xi }} \right)}}{{\partial {\xi ^j}\partial {\xi ^i}}}} \right]}_{n \times n}} \cdot d{\boldsymbol{\xi }}}
		= \int\limits_{{{\boldsymbol{\xi }}_0}}^{\boldsymbol{\xi }} {d{{\left[ {\frac{{\partial g\left( {\boldsymbol{\xi }} \right)}}{{\partial {\xi ^i}}}} \right]}_{n \times 1}}}  \hfill \\
		&= \nabla g\left( {\boldsymbol{\xi }} \right) - \nabla g\left( {{{\boldsymbol{\xi }}_0}} \right). 
	\end{aligned}
\end{equation}

Integrating again yields an expression for the potential function $V\left( {{\boldsymbol{\xi }}} \right)$:
\begin{equation}
	\begin{aligned}
		V\left( {{\boldsymbol{\xi }}} \right) &= \int\limits_{{{\boldsymbol{\xi }}_0}}^{\boldsymbol{\xi }} {\mathbf{f}}  \cdot d{\boldsymbol{\xi }} 
		= \int\limits_{{{\boldsymbol{\xi }}_0}}^{\boldsymbol{\xi }} {\nabla g\left( {\boldsymbol{\xi }} \right) \cdot d{\boldsymbol{\xi }}}  - \nabla g\left( {{{\boldsymbol{\xi }}_0}} \right) \cdot \left( {{\boldsymbol{\xi }} - {{\boldsymbol{\xi }}_0}} \right) \hfill \\
		&= g\left( {\boldsymbol{\xi }} \right) - g\left( {{{\boldsymbol{\xi }}_0}} \right) - \nabla g\left( {{{\boldsymbol{\xi }}_0}} \right) \cdot \left( {{\boldsymbol{\xi }} - {{\boldsymbol{\xi }}_0}} \right).
	\end{aligned} 
\end{equation}

%% file: appendices/appendix_B.tex

\section{Proof of Proposition \ref{pro:PassiveDSContr1}}
\label{apd:PassiveDSContr1}
\begin{proof}
Next, we show that the system (\ref{eq:RigidDyn}) controlled by (\ref{eq:PassiveDSContr1}) is passive with the storage function (\ref{eq:StorageFun}).
	
The change rate of the energy storage function (\ref{eq:StorageFun}) can be expressed as
\begin{equation}
	\label{eq:ratestoragefun}
	\begin{aligned}
		{\dot W}
		&={{{\boldsymbol{\dot \xi }}}^T}{\mathbf{M}}({\boldsymbol{\xi }}){\boldsymbol{\ddot \xi }} + \frac{1}{2}{{{\boldsymbol{\dot \xi }}}^T}{\mathbf{\dot M}}({\boldsymbol{\xi }}){\boldsymbol{\dot \xi }} \\
		&\quad + {\lambda _1}{\nabla _{\boldsymbol{\xi }}}V{({\boldsymbol{\xi }},{\xi _{vir}})^T}{\boldsymbol{\dot \xi }}{\text{ + }}{\lambda _1}\frac{{\partial V({\boldsymbol{\xi }},{\xi _{vir}})}}{{\partial {\xi _{vir}}}}{{\dot \xi }_{vir}} \\ 
		&={{{\boldsymbol{\dot \xi }}}^T}\left( {{{\boldsymbol{\tau }}_c} + {{\boldsymbol{\tau }}_e} - {\mathbf{C}}({\boldsymbol{\xi }},{\boldsymbol{\dot \xi }}){\boldsymbol{\dot \xi }} - {\mathbf{g}}({\boldsymbol{\xi }})} \right) + \frac{1}{2}{{{\boldsymbol{\dot \xi }}}^T}{\mathbf{\dot M}}({\boldsymbol{\xi }}){\boldsymbol{\dot \xi }} \\
		&\quad + {\lambda _1}{\nabla _{\boldsymbol{\xi }}}V{({\boldsymbol{\xi }},{\xi _{vir}})^T}{\boldsymbol{\dot \xi }}{\text{ + }}{\lambda _1}\frac{{\partial V({\boldsymbol{\xi }},{\xi _{vir}})}}{{\partial {\xi _{vir}}}}{{\dot \xi }_{vir}} \\
		&=\frac{1}{2}{{{\boldsymbol{\dot \xi }}}^T}({\mathbf{\dot M}} - 2{\mathbf{C}}){\boldsymbol{\dot \xi }} - {{{\boldsymbol{\dot \xi }}}^T}{\mathbf{D }({\boldsymbol{\xi }},{\xi _{vir}})\boldsymbol{\dot \xi}} + {{{\boldsymbol{\dot \xi }}}^T}{{\boldsymbol{\tau }}_e} \\
		&\quad + {\lambda _1}{{{\boldsymbol{\dot \xi }}}^T}{{\mathbf{f}}_{pro}}({\boldsymbol{\xi }},{\xi _{vir}}) + {\lambda _1}{\nabla _{\boldsymbol{\xi }}}V{({\boldsymbol{\xi }},{\xi _{vir}})^T}{\boldsymbol{\dot \xi }}\\
		&\quad + {\lambda _1}\frac{{\partial V({\boldsymbol{\xi }},{\xi _{vir}})}}{{\partial {\xi _{vir}}}}{{\dot \xi }_{vir}}.
	\end{aligned}
\end{equation}

From the definition of the high-dimensional DS (\ref{eq:HighDimDS}), we can obtain the following relation:
\begin{subequations}
	\label{eq:HDDS}
	\begin{align}
		\left[ {\begin{array}{*{20}{c}}
					{{\boldsymbol{\dot \xi }}} \\ 
					{{{\dot \xi }_{vir}}} 
		\end{array}} \right] &= {\mathbf{f}}({\boldsymbol{\xi }},{\xi _{vir}}) =  - \left[ {\begin{array}{*{20}{c}}
					{{\nabla _{\boldsymbol{\xi }}}V({\boldsymbol{\xi }},{\xi _{vir}})} \\ 
					{\frac{{\partial V({\boldsymbol{\xi }},{\xi _{vir}})}}{{\partial {\xi _{vir}}}}} 
		\end{array}} \right],\label{eq:HDDSA}\\
			{{\mathbf{f}}_{pro}}({\boldsymbol{\xi }},{\xi _{vir}}) &= {\pi _{(1, \ldots ,n)}}\left[ {{\mathbf{f}}({\boldsymbol{\xi }},{\xi _{vir}})} \right] =  - {\nabla _{\boldsymbol{\xi }}}V({\boldsymbol{\xi }},{\xi _{vir}}).\label{eq:HDDSB}
	\end{align}
\end{subequations}

Observe (\ref{eq:HDDSA}), we have 
\begin{equation}
	\label{eq:Vneg}
	\frac{{\partial V({\boldsymbol{\xi }},{\xi _{vir}})}}{{\partial {\xi _{vir}}}}\mathop {{\xi _{vir}}}\limits^.  =  - {\left( {\frac{{\partial V({\boldsymbol{\xi }},{\xi _{vir}})}}{{\partial {\xi _{vir}}}}} \right)^2} \leqslant 0.
\end{equation}

Notice that $\left( {\mathbf{\dot M}} - 2{\mathbf{C}}\right)$ is a skew matrix while substituting (\ref{eq:HDDSB}) and (\ref{eq:Vneg}) into (\ref{eq:ratestoragefun}):
\begin{equation}
	\label{eq:passcon}
	\begin{aligned}
		{\dot W}
		&= - {{{\boldsymbol{\dot \xi }}}^T}{\mathbf{D }({\boldsymbol{\xi }},{\xi _{vir}})\boldsymbol{\dot \xi}}{\text{ + }}{\lambda _1}\frac{{\partial V({\boldsymbol{\xi }},{\xi _{vir}})}}{{\partial {\xi _{vir}}}}{{\dot \xi }_{vir}} + {{{\boldsymbol{\dot \xi }}}^T}{{\boldsymbol{\tau }}_e}< {{{\boldsymbol{\dot \xi }}}^T}{{\boldsymbol{\tau }}_e}.
	\end{aligned}
\end{equation}	
The inequality (\ref{eq:passcon}) means that the robotic system is passive under the controller (\ref{eq:PassiveDSContr1}).
\end{proof}

%% file: appendices/appendix_C.tex

\section{The Derivation of Conservative Fitting Function}
\label{apd:FittingFunction}
\begin{proof}
We give details of the derivation of the fitting function ${{\mathbf{\hat f}}_{\boldsymbol{\nu }}}\left( {\boldsymbol{\xi }} \right)$ used in this paper.

For the three-dimensional linear space (i.e., ${\boldsymbol{\xi }} = {\left[ {\begin{array}{*{20}{c}}
			{{\xi _1}}&{{\xi _2}}&{{\xi _3}} 
\end{array}} \right]^T} \in {\mathbb{R}^3}$), the simplest form of the conservative stiffness matrix is the diagonal matrix ${\mathbf{K}}\left( {{\boldsymbol{\xi }}} \right) = diag\left( {{{\left[ {\begin{array}{*{20}{c}}
		{{k_{11}}\left( {{\xi _1}} \right)}&{{k_{22}}\left( {{\xi _2}} \right)}&{{k_{33}}\left( {{\xi _3}} \right)} 
\end{array}} \right]}^T}} \right)$, which is simultaneously symmetric and exact. We choose cubic polynomials for parameterization:
\begin{equation}
	{{\mathbf{K}}_{\boldsymbol{\nu }}}\left( {{\boldsymbol{\xi }}} \right) = diag\left( {\left[ {\begin{array}{*{20}{c}}
				{{\nu _1}{{\left( {{\xi _1}} \right)}^3} + {\nu _2}{{\left( {{\xi _1}} \right)}^2} + {\nu _3}{\xi _1} + {\nu _4}} \\ 
				{{\nu _5}{{\left( {{\xi _2}} \right)}^3} + {\nu _6}{{\left( {{\xi _2}} \right)}^2} + {\nu _7}{\xi _2} + {\nu _8}} \\ 
				{{\nu _9}{{\left( {{\xi _3}} \right)}^3} + {\nu _{10}}{{\left( {{\xi _3}} \right)}^2} + {\nu _{11}}{\xi _3} + {\nu _{12}}} 
		\end{array}} \right]} \right),
\end{equation}
where ${\boldsymbol{\nu }}$ is the parameter vector.

The conservative fitting function ${{\mathbf{\hat f}}_{\boldsymbol{\nu }}}\left( {\boldsymbol{\xi }} \right)$ can be obtained by integrating the conservative stiffness ${{\mathbf{K}}_{\boldsymbol{\nu }}}\left( {{\boldsymbol{\xi }}} \right)$:
\begin{equation}
	{{\mathbf{\hat f}}_{\boldsymbol{\nu }}}\left( {\boldsymbol{\xi }} \right) = \int\limits_{{{\boldsymbol{\xi }}_0}}^{\boldsymbol{\xi }} {{{\mathbf{K}}_{\boldsymbol{\nu }}}} d{\boldsymbol{\xi }} = {{\mathbf{f}}_{\boldsymbol{\nu }}}\left( {\boldsymbol{\xi }} \right) - {{\mathbf{f}}_{\boldsymbol{\nu }}}\left( {{{\boldsymbol{\xi }}_0}} \right),
\end{equation}	
where
\begin{equation}
	{{\mathbf{f}}_{\boldsymbol{\nu }}}\left( {\boldsymbol{\xi }} \right) = \left[ {\begin{array}{*{20}{c}}
			{\begin{array}{*{20}{c}}
					{{\nu _1}{{\left( {{\xi _1}} \right)}^4}/4 + {v_2}{{\left( {{\xi _1}} \right)}^3}/3 + {\nu _3}{{\left( {{\xi _1}} \right)}^2}/2 + {\nu _4}{\xi _1}} \\ 
					{{\nu _5}{{\left( {{\xi _2}} \right)}^4}/4 + {\nu _6}{{\left( {{\xi _2}} \right)}^3}/3 + {\nu _7}{{\left( {{\xi _2}} \right)}^2}/2 + {\nu _8}{\xi _2}} 
			\end{array}} \\ 
			{{\nu _9}{{\left( {{\xi _3}} \right)}^4}/4 + {\nu _{10}}{{\left( {{\xi _3}} \right)}^3}/3 + {\nu _{11}}{{\left( {{\xi _3}} \right)}^2}/2 + {\nu _{12}}{\xi _3}} 
	\end{array}} \right].
\end{equation}	

For the two-dimensional linear space (i.e., ${\boldsymbol{\xi }} = {\left[ {\begin{array}{*{20}{c}}
			{{\xi _1}}&{{\xi _2}} 
	\end{array}} \right]^T} \in {\mathbb{R}^2}$), ${{\mathbf{f}}_{\boldsymbol{\nu }}}\left( {\boldsymbol{\xi }} \right)$ can be expressed as
\begin{equation}
	{{\mathbf{f}}_{\boldsymbol{\nu }}}\left( {\boldsymbol{\xi }} \right) = \left[ {\begin{array}{*{20}{c}}
			{{\nu _1}{{\left( {{\xi _1}} \right)}^4}/4 + {v_2}{{\left( {{\xi _1}} \right)}^3}/3 + {\nu _3}{{\left( {{\xi _1}} \right)}^2}/2 + {\nu _4}{\xi _1}} \\ 
			{{\nu _5}{{\left( {{\xi _2}} \right)}^4}/4 + {\nu _6}{{\left( {{\xi _2}} \right)}^3}/3 + {\nu _7}{{\left( {{\xi _2}} \right)}^2}/2 + {\nu _8}{\xi _2}} 
	\end{array}} \right].
\end{equation}
In this paper, the use of a diagonal stiffness matrix in the simplest form and cubic polynomials produce good decomposition. For more complex cases, the non-diagonal stiffness matrix and higher-order polynomials can be further considered.

\end{proof}

%% file: appendices/appendix_D.tex

\section{Detailed Description of Angular Velocity}
\label{apd:AngularVelocity}
\begin{proof}
	Next, we introduce the definition of the angular velocity $\Omega$ of the 2D and 3D vector fields, and further come to illustrate the effect of the vector field decomposition on the angular velocity $\Omega$ of the vector field.
	\begin{enumerate}
		\item 
		The define of angular velocity $\Omega$ for the vector fields.
		\begin{enumerate}
			\item
		The 3D vector field:
		
		The vector field can be expressed in the form of components:
		\begin{equation}
			{\mathbf{f}}({\boldsymbol{\xi }}) = {f_1}({\boldsymbol{\xi }}){ {\mathbf{e}}_1} + {f_2}({\boldsymbol{\xi }}){ {\mathbf{e}}_2} + {f_3}({\boldsymbol{\xi }}){ {\mathbf{e}}_3},
		\end{equation}		
		where ${{\mathbf{e}}_i}$ is the coordinate base corresponding to ${\xi_i}$.
		
		Then the curl of the vector field ${\mathbf{f}}({\boldsymbol{\xi }})$ can be expressed as
		\begin{equation}
			\begin{aligned}	
				\operatorname{curl} {\mathbf{f}}\left( {\boldsymbol{\xi }}\right)  & = \nabla  \times {\mathbf{f}}\left( {\boldsymbol{\xi }}\right) = \left( {\frac{{\partial {f_3}}}{{\partial {\xi _2}}} - \frac{{\partial {f_2}}}{{\partial {\xi _3}}}} \right){ {\mathbf{e}}_1} \\
				&\quad + \left( {\frac{{\partial {f_1}}}{{\partial {\xi _3}}} - \frac{{\partial {f_3}}}{{\partial {\xi _1}}}} \right){ {\mathbf{e}}_2} + \left( {\frac{{\partial {f_2}}}{{\partial {\xi _1}}} - \frac{{\partial {f_1}}}{{\partial {\xi _2}}}} \right){ {\mathbf{e}}_3}.
			\end{aligned}
		\end{equation}	
		The angular velocity of a 3D vector field is defined as the following scalar quantity:
		\begin{equation}
			\Omega = \frac{1}{2}(\nabla  \times {\mathbf{f}}\left( {\boldsymbol{\xi }}\right)) \cdot \frac{{\mathbf{f}}}{{\left\| {\mathbf{f}} \right\|}}.
		\end{equation}		
			\item
		The 2D vector field:
		
		The vector field can be expressed in the form of components:
		\begin{equation}
			{\mathbf{f}}\left( {\boldsymbol{\xi }}\right) = {f_1}\left( {\boldsymbol{\xi }}\right){ {\mathbf{e}}_1} + {f_2}\left( {\boldsymbol{\xi }}\right){ {\mathbf{e}}_2}.
		\end{equation}
		Then the curl of the 2D vector field ${\mathbf{f}}({\boldsymbol{\xi }})$ can be expressed as
		\begin{equation}
			\operatorname{curl} {\mathbf{f}}\left( {\boldsymbol{\xi }}\right) = \nabla  \times {\mathbf{f}}\left( {\boldsymbol{\xi }}\right) = \left( {\frac{{\partial {f_2}}}{{\partial {\xi _1}}} - \frac{{\partial {f_1}}}{{\partial {\xi _2}}}} \right){ {\mathbf{e}}_3}.
		\end{equation}
		The angular velocity of the vector field is half the curl:
		\begin{equation}
			\Omega  = \frac{1}{2}\operatorname{curl} {\mathbf{f}}\left( {\boldsymbol{\xi }}\right) = \frac{1}{2}\left( {\frac{{\partial {f_2}}}{{\partial {\xi _1}}} - \frac{{\partial {f_1}}}{{\partial {\xi _2}}}} \right){ {\mathbf{e}}_3}.
		\end{equation}
		\end{enumerate}
		
		\item 
		The effect of vector field decomposition on angular velocity.
		
		The Vector field ${\mathbf{f}}\left( {\boldsymbol{\xi }} \right)$ have infinitely many different decompositions (show decomposition 1 and decomposition 2 below):
		\begin{equation}
			\label{eq:diffdecomp}
			{\mathbf{f}}\left( {\boldsymbol{\xi }} \right) = {{\mathbf{f}}_{c1}}\left( {\boldsymbol{\xi }} \right) + {{\mathbf{f}}_{nc1}}\left( {\boldsymbol{\xi }} \right) = {{\mathbf{f}}_{c2}}\left( {\boldsymbol{\xi }} \right) + {{\mathbf{f}}_{nc2}}\left( {\boldsymbol{\xi }} \right).
		\end{equation}
		Notice that the curl of the conservative vector field is zero:
		\begin{equation}
			\label{eq:curleq}
			curl\left( {{{\mathbf{f}}_{c1}}\left( {\boldsymbol{\xi }} \right)} \right) = curl\left( {{{\mathbf{f}}_{c2}}\left( {\boldsymbol{\xi }} \right)} \right) = {\mathbf{0}}.
		\end{equation}
		Combining equations (\ref{eq:diffdecomp}) and (\ref{eq:curleq}), we have
		\begin{equation}
			\label{eq:curleqnc}
			curl\left( {{\mathbf{f}}\left( {\boldsymbol{\xi }} \right)} \right) = curl\left( {{{\mathbf{f}}_{nc1}}\left( {\boldsymbol{\xi }} \right)} \right) = curl\left( {{{\mathbf{f}}_{nc2}}\left( {\boldsymbol{\xi }} \right)} \right).
		\end{equation}
		(\ref{eq:curleqnc}) shows that the original vector field ${\mathbf{f}}\left( {\boldsymbol{\xi }} \right)$ has the same curl as the non-conservative component part ${{\mathbf{f}}_{nc}}\left( {\boldsymbol{\xi }} \right)$ after decomposition.
		
		\begin{enumerate}
			\item
		The 2D vector field:
		
		Combining equations (\ref{eq:curleqnc}) and $\Omega  = \frac{1}{2}\operatorname{curl} {\mathbf{f}}$, we have
		\begin{equation}
			\label{eq:omegaeq}
			\Omega \left( {{\mathbf{f}}\left( {\boldsymbol{\xi }} \right)} \right) = \Omega \left( {{{\mathbf{f}}_{nc1}}\left( {\boldsymbol{\xi }} \right)} \right) = \Omega \left( {{{\mathbf{f}}_{nc2}}\left( {\boldsymbol{\xi }} \right)} \right).
		\end{equation}
		(\ref{eq:omegaeq}) reveals the reason why Figure~\ref{fig:SnakeOmega}, \ref{fig:SnakeOmegaNonCon10}, and \ref{fig:SnakeOmegaNonCon11} have the same pseudo-color.
		
			\item
		The 3D vector field:
		
		Since angular velocity is defined as $\Omega  = \frac{1}{2}\operatorname{curl} {\mathbf{f}} \cdot \frac{{\mathbf{f}}}{{\left\| {\mathbf{f}} \right\|}}$ and the directions of ${{{\mathbf{f}}_{nc1}}\left( {\boldsymbol{\xi }} \right)}$ and ${{{\mathbf{f}}_{nc2}}\left( {\boldsymbol{\xi }} \right)}$ are generally different, (\ref{eq:omegaeq}) no longer holds.
		
		\end{enumerate}
		
	\end{enumerate}
\end{proof}